\documentclass{article} 
\usepackage{iclr2027_conference,times}

\usepackage{amsmath,amsfonts,bm}

\def\eqref#1{equation~\ref{#1}}

\def\1{\bm{1}}

\DeclareMathAlphabet{\mathsfit}{\encodingdefault}{\sfdefault}{m}{sl}
\SetMathAlphabet{\mathsfit}{bold}{\encodingdefault}{\sfdefault}{bx}{n}

\usepackage{hyperref}
\usepackage{url}
\usepackage{fontawesome5}
\usepackage{booktabs}
\usepackage{multirow}
\usepackage{graphicx}
\usepackage[table]{xcolor}
\usepackage{xspace}
\usepackage{colortbl}
\usepackage{caption}
\usepackage{subcaption}
\usepackage{wrapfig}
\usepackage{needspace}
\newcommand{\method}{UniSense\xspace}
\usepackage{titletoc}
\usepackage[most]{tcolorbox}
\usepackage{makecell}
\usepackage{booktabs}
\usepackage{tabularx}
\usepackage{array}
\usepackage{tcolorbox} 
\usepackage[most]{tcolorbox}
\usepackage{xcolor}
\usepackage{tabularx}
\usepackage{array}

\newcolumntype{M}[1]{>{\hsize=#1\hsize\linewidth=\hsize\raggedright\arraybackslash}X}
\newcolumntype{N}[1]{>{\hsize=#1\hsize\linewidth=\hsize\centering\arraybackslash}X}
\definecolor{basicblue}{HTML}{DCEAF4}
\definecolor{basicpurple}{HTML}{E8E1F2}
\definecolor{depthcolor}{HTML}{F2DCCB}
\definecolor{imucolor}{HTML}{D9E8F4}
\definecolor{motioncolor}{HTML}{E4DCF0}
\definecolor{pccolor}{HTML}{E8DDE6}
\definecolor{radarcolor}{HTML}{DCEAD9}
\definecolor{smellcolor}{HTML}{F3E9C9}
\definecolor{tactilecolor}{HTML}{D9EAE5}
\definecolor{thermalcolor}{HTML}{E6DDD2}
\definecolor{groupgray}{HTML}{F1F1F1}
\definecolor{totalgray}{HTML}{F6F6F6}

\definecolor{chatborder}{RGB}{205,211,220}
\definecolor{chatlabel}{RGB}{62,72,88}

\newtcolorbox{chatbox}{
    enhanced,
    breakable,
    colback=white,
    colframe=chatborder,
    boxrule=0.6pt,
    arc=1.5mm,
    outer arc=1.5mm,
    left=8pt,
    right=8pt,
    top=7pt,
    bottom=7pt,
    before skip=8pt,
    after skip=8pt,
    fontupper=\small
}

\title{Beyond Omni: A Unified MLLM for Embodied Sensory Understanding}

\definecolor{linkblue}{rgb}{0.1, 0.5, 0.7}

\definecolor{closedcolor}{RGB}{232,236,242}
\definecolor{opensourcecolor}{RGB}{238,238,238}
\definecolor{generalcolor}{RGB}{225,238,250}
\definecolor{omnimodalcolor}{RGB}{238,232,248}
\definecolor{specializedcolor}{RGB}{252,244,207}
\definecolor{ourscolor}{RGB}{232,245,233}

\author{%
  \textbf{Shiyun Xiong$^{1,*}$ \quad Dongming Wu$^{1,*}$ \quad Peiwen Sun$^{1}$ \quad Bokang Yang$^{2,3}$ \quad Jiaxu Wang$^{1}$}\\
  \textbf{Xiaolou Sun$^{4}$ \quad Mingxing Wen$^{5}$ \quad Xiangyu Yue$^{1,2\dagger}$}\\[4pt]
  \textnormal{$^{1}$MMLab, The Chinese University of Hong Kong}
  \textnormal{$^{2}$Shenzhen Loop Area Institute}\\
  \textnormal{$^{3}$The Chinese University of Hong Kong, Shenzhen}
  \textnormal{$^{4}$Southeast University}\\
  \textnormal{$^{5}$China Singapore International Joint Research Institute}%
}
\vspace{-3mm}

\iclrfinalcopy 
\begin{document}

\maketitle
\begingroup
\renewcommand{\thefootnote}{}
\footnotetext{%
  \textsuperscript{*}Equal contribution.
  \quad
  \textsuperscript{$\dagger$}Corresponding author.\\
  \url{https://github.com/Shiyun-x/Unisense}
}
\endgroup
\begin{figure}[!ht]
    \centering
    \includegraphics[width=1\linewidth]{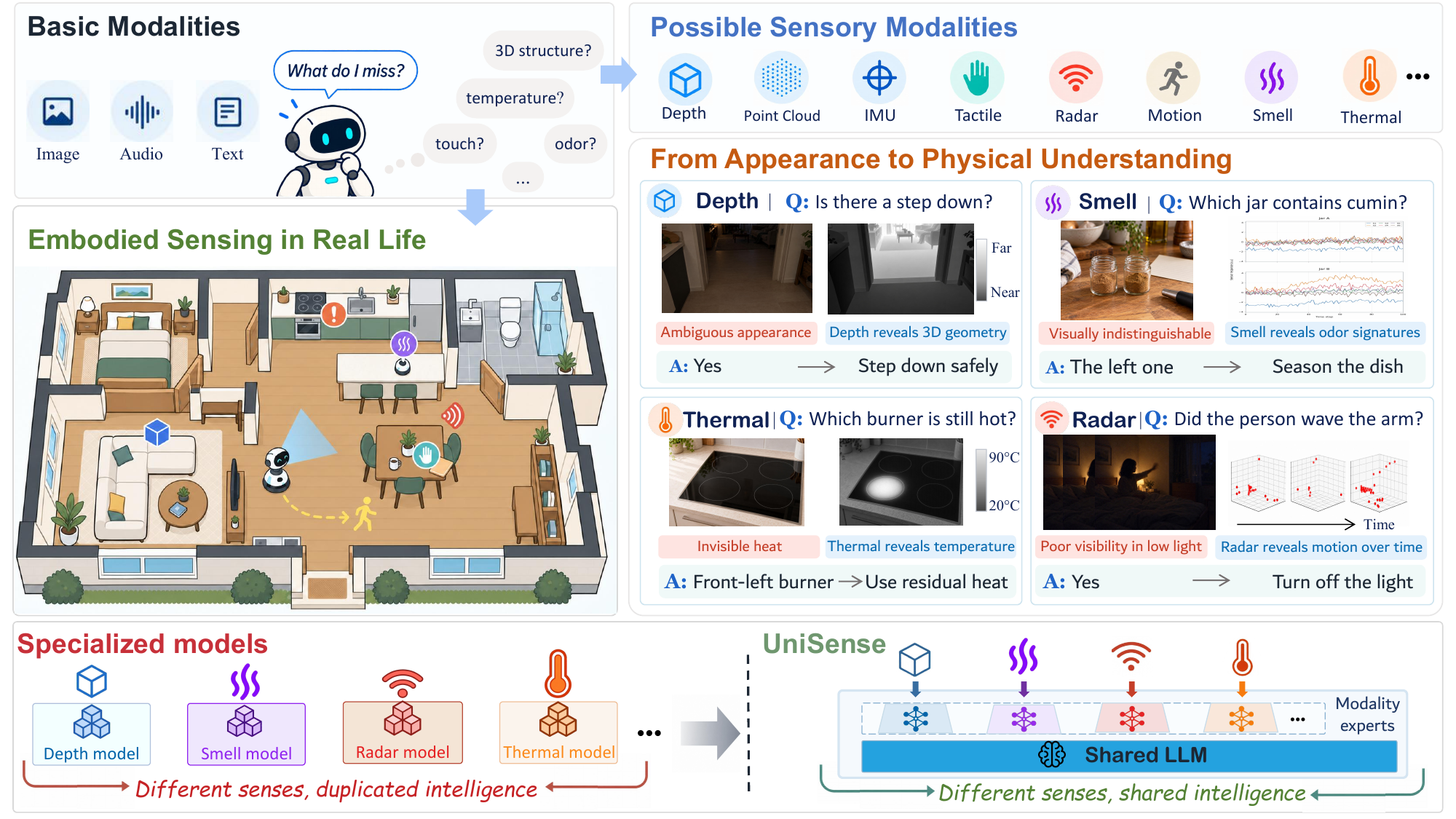}
    \vspace{-3mm}
\caption{\textbf{Different senses, shared intelligence}: broad embodied understanding requires diverse physical evidence while drawing on shared knowledge and reasoning.
Examples from depth, smell, thermal, and radar show how these signals resolve ambiguities left by basic-modality observations.
Rather than duplicating general intelligence across separate sensory models, \textbf{\method} combines modality-specific experts with a shared LLM.}
    \label{fig:main_image}
    \vspace{-2mm}
\end{figure}

\begin{abstract}
Recent embodied systems increasingly build on multimodal large language models (MLLMs) for their pretrained understanding and reasoning capabilities, but their perception of the physical world remains largely limited to basic modalities such as vision and audio.
However, broadening MLLMs to more diverse sensory modalities from the physical environment is challenging: their representations are often weakly connected to the MLLM and vary substantially in structure and statistics, while their interpretation still requires shared high-level understanding and reasoning.
We therefore propose \textbf{\method}, a unified embodied MLLM framework that supports diverse physical sensory modalities within a shared multimodal architecture.
\method follows a two-stage training process: (i) aligning sensory representations with language to establish a language-accessible interface; and (ii) tuning modality-specific experts within the shared LLM to provide specialized processing for different sensory modalities.
To support broad sensory learning, we curate  \textbf{\method--1M}, a large-scale embodied sensory training corpus comprising 1M samples across 8 sensory modalities. We further evaluate \method on 12 benchmarks spanning depth, point cloud, IMU, tactile, thermal, radar, motion, and smell understanding. Across these heterogeneous sensing tasks, \method achieves strong performance and outperforms both generalist and specialized models on most benchmarks, demonstrating the potential of pretrained MLLMs as unified models for embodied sensory understanding.
\end{abstract}

\section{Introduction}
The rapid progress of multimodal large language models (MLLMs) is reshaping Embodied AI, driving a growing line of generalist embodied models that harness MLLMs' multimodal understanding and reasoning capabilities for physical-world perception and interaction. Representative systems such as PaLM-E~\citep{driess2023palm}, RoboBrain~\citep{robobrain,robobrain2}, RynnBrain~\citep{dang2026rynnbrain} and Vesta~\citep{bjorck2026vesta} demonstrate the potential of this paradigm. However, while modern MLLMs have developed strong capabilities across modalities such as images, video, and audio, their support for diverse physical signals remains limited.

The real world contains rich information that can be captured by embodied sensory modalities, such as depth, thermal, tactile, and radar.
These modalities provide complementary physical cues about geometry, contact, temperature, and dynamics beyond basic modality inputs.
As Figure~\ref{fig:main_image} illustrates, such cues can clarify physical states that remain ambiguous in basic observations.
Accordingly, recent work has explored three main strategies for incorporating sensory modalities into language-centric models.
\textbf{(i)} \textit{Specialists} develop dedicated interfaces or adaptation mechanisms for individual sensory modalities~\citep{li2025sensorllm,lai2026radarllm,imran2024llasa}.
\textbf{(ii)} \textit{Binders}~\citep{girdhar2023imagebind,zhu2024languagebind} use image or language representations as semantic anchors to align different modalities within a shared representation space.
\textbf{(iii)} \textit{Generalists}~\citep{han2024onellm,moon2024anymal,wu2024next,zhang2025omnibind} integrate multiple modalities through modality-specific encoders, projectors, or routing modules. 
Despite these differences, most sensory modality adaptation is still concentrated at the modality interface, before entering the LLM.
A more detailed discussion of related work is provided in Appendix~\S\ref{app:related_work}. 

This emphasis largely reflects the maturity gap in modality interfaces between embodied sensory and widely studied basic modalities.
On the one hand, the latter benefit from well-established modality-specific encoders, web-scale multimodal data, and widely adopted alignment recipes~\citep{radford2021clip,chu2024qwen2audio,tschannen2025siglip}, whereas many embodied sensory modalities still lack comparable resources, making sensory--language grounding a primary challenge.
On the other hand, despite heterogeneous low-level representations, these modalities ultimately rely on shared high-level understanding and reasoning, motivating a unified LLM backbone. 
For basic modalities, mature interfaces have further enabled omni-modal models to move beyond and introduce modality-dependent computation within the LLM~\citep{li2025unimoe,mingomni2025,shukor2025scaling}.
For embodied sensory modalities, such LLM-side specialization remains limited: existing methods largely share the same parameters across representations with substantially different structures, statistics, and spatiotemporal characteristics.   Prior studies suggest that such parameter sharing across modalities or tasks can introduce conflicting learning signals~\citep{chen2024octavius,liang2025mixture,hao2026uni}.
We therefore view unified embodied sensory modeling across two levels: \emph{semantic grounding} at the modality--language interface, followed by \emph{computational specialization} within the shared LLM.

In this paper, we present \textbf{\method}, a unified MLLM for embodied sensory understanding. \method supports diverse embodied sensory modalities through lightweight modality tokenizers, a unified encoder, modality projectors, and a shared LLM. 
Building on this architecture, \method follows a two-stage training process to address the above challenges.
\textbf{(i)} Aligning different sensory representations with language through caption supervision, making physical sensory information accessible to the LLM, and then
\textbf{(ii)} tuning modality-specific experts within the LLM, allowing different sensory modalities to specialize beyond the shared backbone computation.

To support \method, we develop a data construction pipeline that applies modality-specific preprocessing to heterogeneous sensory observations and pairs the processed inputs with modality-aware captions and instruction--response pairs.
This results in a large-scale training corpus of 1M samples spanning 8 embodied sensory modalities, alongside general multimodal data. We further evaluate \method on 12 benchmarks covering depth, point cloud, IMU, tactile, thermal, radar, motion, and smell. Experiments show that \method outperforms general multimodal and specialized models on most benchmarks.
Our contributions are summarized as follows:
\vspace{-8pt}
\begin{itemize}
    \item We present \textbf{\method}, a \textbf{unified MLLM} that supports text, image, video, audio, and diverse physical sensing modalities within a shared multimodal architecture.
    \item We develop a \textbf{two-stage training strategy} that aligns and specializes sensory processing, enabling language-accessible sensory representations and modality-specific specialization within a shared LLM.
    \item We curate \textbf{\method--1M}, a multimodal training corpus spanning 8 embodied sensory modalities, and evaluate \method on 12 benchmarks for embodied sensory understanding, demonstrating its effectiveness across heterogeneous sensing tasks.
\end{itemize}

\section{Method}
\begin{figure}[t]
\vspace{-8pt}
\centering
\includegraphics[width=1\textwidth]{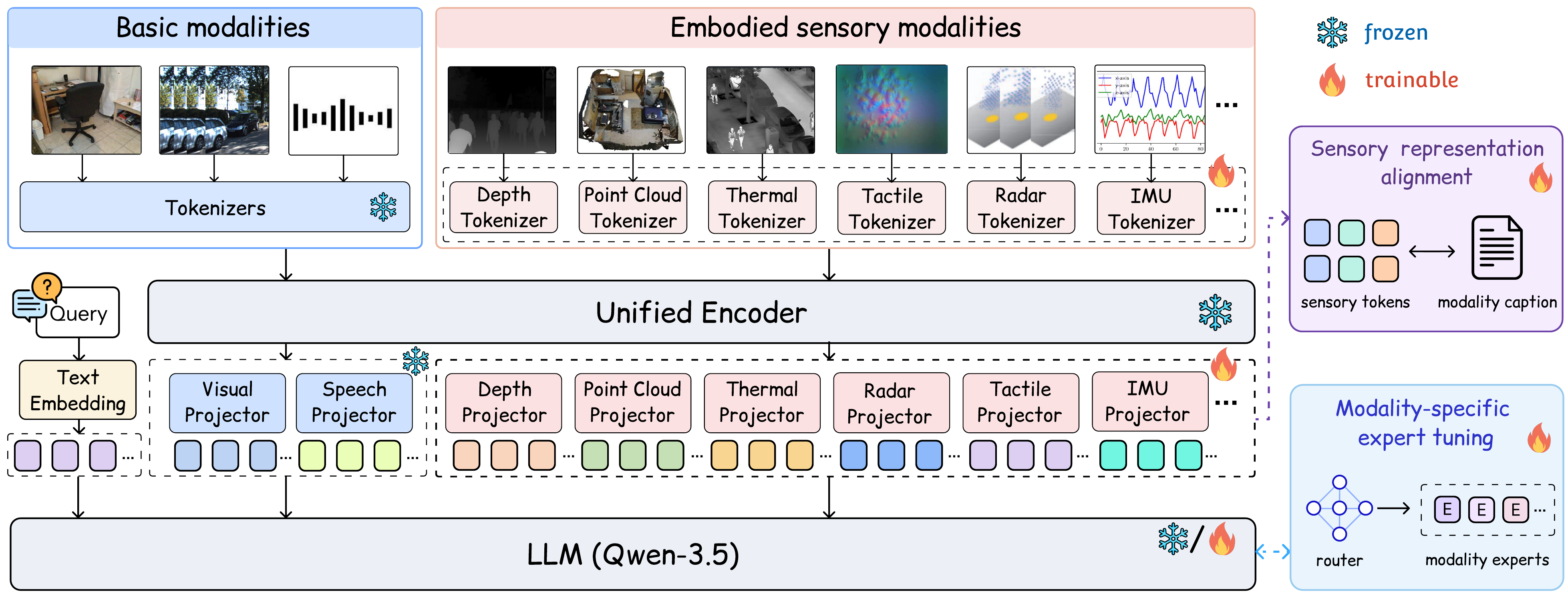}
\caption{
\textbf{The architecture of \method.}
\method unifies basic and embodied sensory modalities in a shared framework.
Basic modalities retain pretrained pathways, while sensory inputs pass through modality-specific tokenizers, a unified encoder, and modality-specific projectors before entering the LLM.
Training proceeds in two stages: (1) modality captions supervise sensory--language alignment; (2) modality-specific experts are then tuned within the LLM.
}
\label{fig:method_overview}
\vspace{-8pt}
\end{figure}

\subsection{Problem Definition}
\label{sec:problem}
We consider a unified embodied sensory understanding task, where a multimodal LLM supports heterogeneous modalities under a shared framework.
Let $\mathcal{M}$ denote the set of supported non-text modalities, including basic modalities (image, video, and audio), and embodied sensory modalities $\mathcal{M}_{\mathrm{s}}\subseteq\mathcal{M}$.
In this work, $\mathcal{M}_{\mathrm{s}}$ comprises depth, point cloud, IMU, tactile, thermal, radar, motion, and smell.
For modality $m\in\mathcal{M}$, we denote its input by $\mathbf{x}^{m}$.
The language instruction provided to the model is denoted by $\mathbf{q}$.

Our goal is to incorporate embodied sensory modalities in $\mathcal{M}_{\mathrm{s}}$ into a MLLM while considering both how their representations are grounded in the model's language space and how they are subsequently processed within the shared backbone.

\subsection{Model Architecture}
\label{sec:arch}
Figure~\ref{fig:method_overview} illustrates the architecture of \method, which consists of modality-specific tokenizers, a unified encoder, modality-specific projectors, and a shared LLM. 
We build \method on pretrained Qwen3.5 models~\citep{team2026qwen3} at 2B and 9B scales, with preliminary multimodal initialization on image and audio data but no additional video-specific training (see Appendix~\S\ref{app:basic_modalities} and \S\ref{app:training_details}).

For an embodied sensory modality $s\in\mathcal{M}_{\mathrm{s}}$, the input $\mathbf{x}^{s}$ is first transformed into encoder-compatible tokens by a tokenizer $\mathcal{T}_s$.
We adopt two tokenization pathways: \emph{pixel-aligned} modalities (e.g., depth, thermal, and tactile) reuse the standard visual patch tokenizer, whereas \emph{non-pixel modalities} (e.g., point cloud, IMU, and radar) employ dedicated modality-specific tokenizers.
The resulting tokens are then processed by a unified encoder $\mathcal{E}$ and mapped to the hidden space of the language model through a modality-specific projector $\mathcal{P}_s$:
\begin{equation}
\mathbf{Z}^{s}
=
\mathcal{P}_s\!\left(
\mathcal{E}(\mathcal{T}_s(\mathbf{x}^{s}))
\right),
\end{equation}
where $\mathbf{Z}^{s}$ denotes the resulting sensory representation. 
Image, video, and audio are processed through their corresponding input pathways. Accordingly, we use $\mathbf{Z}^{m}$ to denote the resulting representation of any non-text modality $m\in\mathcal{M}$ provided to the LLM.
Detailed tokenizer and projector designs are provided in Appendix~\S\ref{app:tokenization_projectors}. 

The resulting modality representations $\{\mathbf{Z}^{m}\}$ are combined with the language instruction $\mathbf{q}$ and processed by the LLM to generate the response.
Although these interfaces make heterogeneous inputs compatible with the pretrained backbone, they do not ensure effective language grounding or appropriate processing.
We address these two aspects through the two training stages described in \S\ref{sec:csa} and \S\ref{sec:msc}.

\subsection{Sensory Representation Alignment}
\label{sec:csa}
\setlength{\intextsep}{2pt}
\begin{wrapfigure}{r}{0.41\columnwidth}
    \centering
     \vspace{-8pt}
    \includegraphics[width=\linewidth]{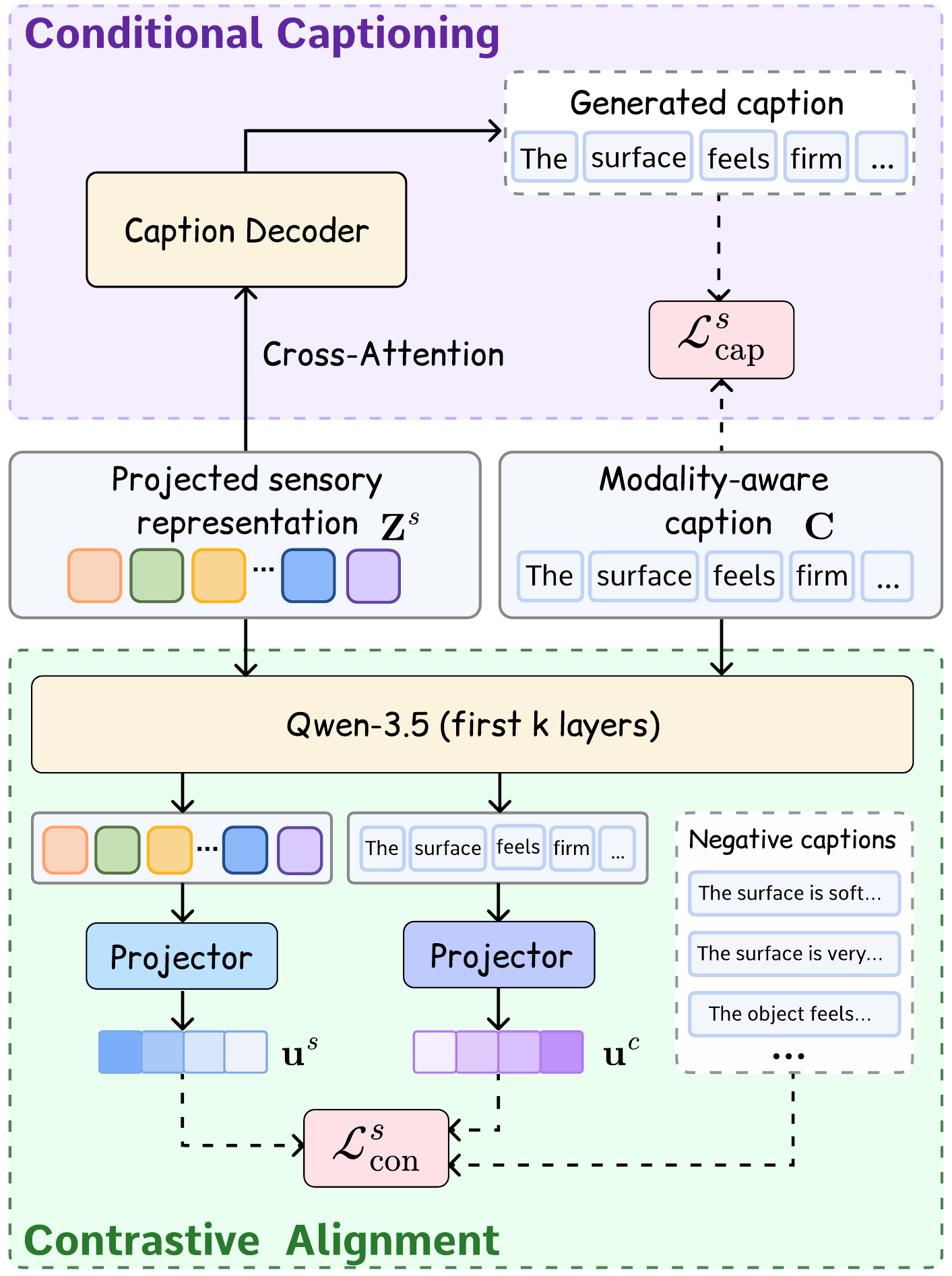}
    \vspace{-8pt}
    \caption{
Overview of the sensory representation alignment stage. Conditional captioning provides language-level supervision by decoding the paired caption from sensory representations, while contrastive alignment aligns sensory and caption representations using the first $k$ LLM layers.
}
    \label{fig:csa}
\end{wrapfigure}
Unlike basic modalities, which have been well studied, embodied sensory modalities lack comparably mature language-aligned representations, making their physical semantics difficult to ground in the target MLLM.
Prior work aligns heterogeneous modalities in shared embedding spaces~\citep{girdhar2023imagebind,zhu2024languagebind}, while generative--contrastive frameworks further show that generative supervision and contrastive learning provide complementary semantic signals~\citep{yu2022coca,lcoembed2025}.
However, such alignment is typically defined outside the representation structure of the target MLLM that ultimately consumes these modality features.
Recent MLLM studies suggest that grounding modality representations toward the pretrained linguistic structure of the target LLM can improve their integration~\citep{alignvlm2025,textspace2026}.

To align embodied sensory representations with language, we construct a modality-aware caption for each sensory input that explicitly verbalizes its physical information.
We use each caption to supervise sensory representations in two complementary ways:
conditional captioning encourages sensory representations to retain physical semantics recoverable through the target MLLM's native language interface, while contrastive alignment uses representations induced by the target MLLM itself to establish explicit sensory--language correspondence.
The overall design is illustrated in Fig.~\ref{fig:csa}, with examples of modality-aware captions provided in Appendix~\S\ref{app:training_visualization}.

For a sensory modality $s\in\mathcal{M}_{\mathrm{s}}$, let $\mathbf{Z}^{s}\in\mathbb{R}^{N_s\times d}$ denote its projected sensory representation and $\mathbf{C}=(c_1,\ldots,c_L)$ the corresponding modality-aware caption.


\noindent\textbf{Conditional Captioning.}
We impose language-level supervision on sensory representations by predicting their paired modality-aware captions.
Given the projected sensory tokens $\mathbf{Z}^{s}$, a modality-specific caption decoder $\mathcal{C}s$ predicts each token of $\mathbf{C}$ using causal self-attention over preceding caption tokens and cross-attention to $\mathbf{Z}^{s}$.
The decoder reuses the token embeddings and vocabulary projection of the target LLM, encouraging $\mathbf{Z}^{s}$ to preserve physical semantics that can be recovered through the model's native language interface.
We optimize this branch with the standard autoregressive token-level cross-entropy loss, denoted by $\mathcal{L}_{\mathrm{cap}}$.


\noindent\textbf{Contrastive Alignment.}
We further establish representation-level correspondence between sensory inputs and their captions using representations induced by the target LLM.
Specifically, $\mathbf{Z}^{s}$ and the embedded caption $\operatorname{Emb}(\mathbf{C})$ are independently processed by the same first $k$ frozen LLM layers, followed by modality-specific aggregation and separate lightweight projection heads:
\begin{equation}
\mathbf{u}^{s}
=
\operatorname{Norm}\!\left(
\phi_s\!\left(\rho_s\!\left(Q_{1:k}(\mathbf{Z}^{s})\right)\right)
\right),
\qquad
\mathbf{u}^{c}
=
\operatorname{Norm}\!\left(
\phi_c\!\left(\rho_c\!\left(Q_{1:k}(\operatorname{Emb}(\mathbf{C}))\right)\right)
\right).
\label{eq:contrastive_alignment}
\end{equation}
Here, $\rho_s$ and $\rho_c$ denote the sensory and caption aggregation operations, respectively, $\phi_s$ and $\phi_c$ are separate projection heads, and $\operatorname{Norm}(\cdot)$ denotes $\ell_2$ normalization.
For a batch of $B$ sensory--caption pairs, representations from the same sensory instance form a positive pair, while those from different sensory instances serve as in-batch negatives.  
We optimize the resulting representations with a standard symmetric InfoNCE loss, denoted by $\mathcal{L}_{\mathrm{con}}$.


\subsection{Modality-Specific Expert Tuning}
\label{sec:msc}
\begin{wrapfigure}{r}{0.42\columnwidth}
    \centering
    \vspace{-12pt}
    \includegraphics[width=\linewidth]{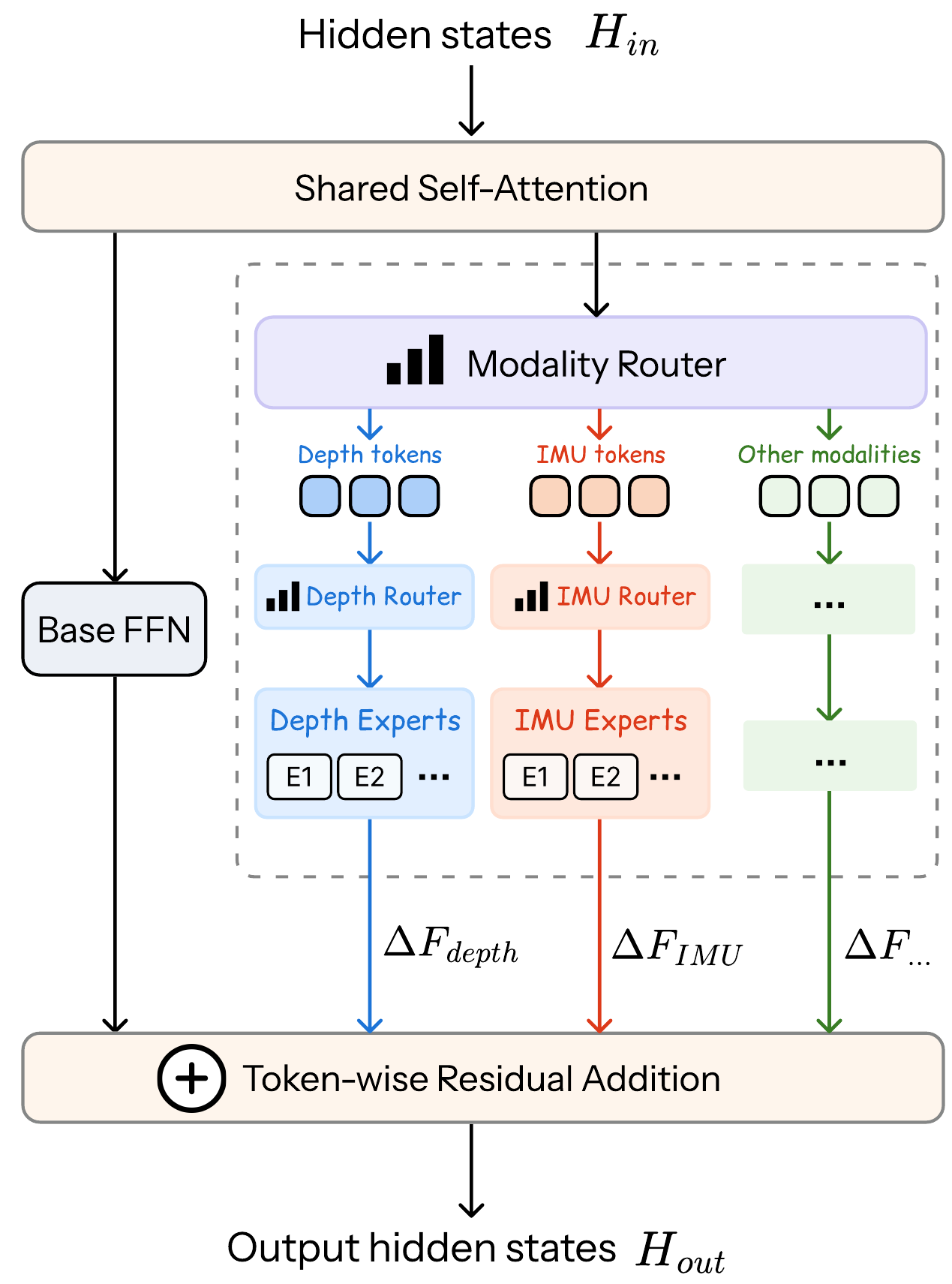}
    \caption{
Modality-specific expert computation within a Transformer layer.
After shared self-attention, a modality router directs sensory tokens to modality-specific expert pools, where token-wise routers produce residual transformations that are added to the shared base FFN output.
}
    \label{fig:msc}
        \vspace{-10pt}
\end{wrapfigure}

The first training stage grounds embodied sensory representations in the language space of the target MLLM.
However, these heterogeneous representations are still processed by the same decoder parameters, which may introduce interference across modalities or tasks~\citep{shen2024mome,liang2025mixture}.
While recent multimodal models introduce modality- or task-specific computation within the backbone~\citep{li2025unimoe,mingomni2025}, embodied sensory modeling remains largely focused on modality-side adaptation~\citep{han2024onellm,li2025sensorllm,zhou2026holollm}.

We therefore introduce modality-specific capacity within the shared LLM, while keeping self-attention shared and specializing only the feed-forward sublayers.
As shown in Fig.~\ref{fig:msc}, the base FFN remains as the shared transformation, with modality-specific residual experts providing specialized processing.

\noindent{\textbf{Modality-Aware Routing.}}
We organize sensory-specific computation through two levels of routing.
For each sensory modality $s$, we maintain a pool of $E$ lightweight residual experts
$\{\mathcal{R}_{s,i}\}_{i=1}^{E}$.
A modality router first directs each sensory token to its corresponding expert pool, after which a token-wise router $g_s$ further determines token-dependent expert contributions,
This allows tokens from the same modality to receive different residual transformations.

\noindent{\textbf{Residual Expert Adaptation.}}
For a hidden state $\mathbf{h}$ associated with sensory modality $s$, the modality-specific residual transformation is defined as:
\begin{equation}
\Delta\mathcal{F}_s(\mathbf{h})
=
\frac{\alpha}{r}
\sum_{i=1}^{E}
w_{s,i}(\mathbf{h})\,
\mathcal{R}_{s,i}(\mathbf{h}),
\qquad
\mathbf{w}_s
=
\operatorname{softmax}\!\left(g_s(\mathbf{h})\right),
\label{eq:msc_experts}
\end{equation}
where $\mathbf{w}_s=[w_{s,1},\ldots,w_{s,E}]$ denotes the normalized expert weights,
$r$ is the low-rank dimension, and $\alpha$ controls the residual scale.
Each $\mathcal{R}_{s,i}$ follows the SwiGLU-based FFN structure of the LLM backbone~\citep{team2026qwen3}, with a low-rank parameterization.

The resulting residual is then added token-wise to the output of the shared pretrained FFN:
\begin{equation}
\mathbf{h}'
=
\mathcal{F}_{\mathrm{base}}(\mathbf{h})
+
\Delta\mathcal{F}_s(\mathbf{h}),
\label{eq:msc_residual}
\end{equation}
where $\mathcal{F}_{\mathrm{base}}$ denotes the frozen FFN shared across modalities.
Thus, each token receives an input-dependent sensory-specific correction while retaining the shared pretrained transformation.

The residual expert modules  are introduced only in selected full-attention Transformer layers, while the remaining layers stay unchanged.
We further use balance and diversity regularization to stabilize expert utilization and discourage redundant residual transformations. More details are provided in Appendix~\S\ref{app:msc_impl}.

\subsection{Training}
\label{sec:training}

After the preliminary image--audio initialization described in \S\ref{sec:arch}, we train \method on a unified collection of embodied sensory datasets spanning 8 sensory modalities.
Each sensory sample is paired with a modality-aware caption used for sensory representation alignment.
Detailed data sources and sampling strategies are provided in \S\ref{app:datasets_and_benchmarks}.

Training proceeds in two stages.
The first aligns sensory representations with language, while the second tunes modality-specific experts within the shared LLM.

\noindent{\textbf{Alignment Stage.}}
We freeze the pretrained LLM backbone and optimize the newly introduced sensory tokenizers, projectors, caption decoders, and contrastive projection heads.
Given a sensory input and its instruction, $\mathcal{L}_{\mathrm{LM}}$ is the autoregressive cross-entropy loss over the target response tokens.
The paired modality-aware caption further provides the captioning objective $\mathcal{L}_{\mathrm{cap}}$ and the contrastive objective $\mathcal{L}_{\mathrm{con}}$ introduced in \S\ref{sec:csa}.
The overall objective is:
\begin{equation}
\mathcal{L}^{(1)}
=
\mathcal{L}_{\mathrm{LM}}
+
\lambda_{\mathrm{align}}
\left(
\mathcal{L}_{\mathrm{cap}}
+
\lambda_{\mathrm{con}}\mathcal{L}_{\mathrm{con}}
\right),
\end{equation}
where $\lambda_{\mathrm{align}}$ controls the overall contribution of the alignment objectives, and $\lambda_{\mathrm{con}}$ determines the relative weight of the contrastive objective.

\noindent{\textbf{Tuning Stage.}}
The pretrained LLM backbone remains frozen, while we optimize the sensory modules, modality routers, and residual experts.
We retain the language-modeling and captioning objectives, but disable the contrastive branch to permit task-driven specialization.
We further apply the load-balancing loss $\mathcal{L}_{\mathrm{bal}}$ and expert-diversity loss $\mathcal{L}_{\mathrm{div}}$ to regularize expert learning.
The second-stage objective is:
\begin{equation}
\mathcal{L}^{(2)}
=
\mathcal{L}_{\mathrm{LM}}
+
\lambda_{\mathrm{align}}\mathcal{L}_{\mathrm{cap}}
+
\lambda_{\mathrm{bal}}\mathcal{L}_{\mathrm{bal}}
+
\lambda_{\mathrm{div}}\mathcal{L}_{\mathrm{div}},
\end{equation}
where $\lambda_{\mathrm{align}}$, $\lambda_{\mathrm{bal}}$, and $\lambda_{\mathrm{div}}$ weight the captioning, load-balancing, and expert-diversity terms, respectively.
Detailed settings are provided in Appendix~\S\ref{app:training_details}.

\section{Experiments}
\begin{table*}[t]
\centering
\caption{
\textbf{Evaluation on General Multimodal benchmarks.}
(a) Image-text results on TextVQA, MME, and MMBench.
(b) Audio-image-text results on the image-audio variants of
TextVQA, DocVQA, and ChartQA.
(c) Video-text results on Video-MME and EgoSchema.
Video-MME results are reported without subtitles.
``--`` denotes unavailable results.
Due to space limitations, the citations of all benchmarks and baseline
models reported in the main-text tables are provided in
Appendix~\S\ref{app:supp_citations}.
}
\label{tab:general_capability}

\begin{subtable}[t]{0.32\textwidth}
\centering
\vspace{0pt}
\caption{Image-text benchmarks.}
\vspace{-5pt}
\label{tab:general_vlm_comparison}

\setlength{\tabcolsep}{1.5pt}
\renewcommand{\arraystretch}{0.96}

\resizebox{\linewidth}{!}{
\begin{tabular}{@{}lccc@{}}
\toprule
Model
& TextVQA$_{\mathrm{val}}$
& MME
& MMBench \\
\midrule

\rowcolor{generalcolor}
\multicolumn{4}{c}{\textit{General MLLMs}} \\

Qwen3-VL-8B-Instruct
& -- & 2243 & 84.5 \\

GLM-4.5V
& 72.0 & 2423.8 & 88.2 \\

GLM-4.1V-9B
& 79.6 & 2445.8 & 85.8 \\

Kimi-VL-A3B-2506
& 77.7 & 2352.7 & 84.4 \\

InternVL3.5-8B
& 78.2 & 2380.6 & 79.5 \\

InternVL3.5-30B-A3B
& 80.5 & 2461.9 & 84.8 \\

\midrule
\rowcolor{ourscolor}
\multicolumn{4}{c}{\textit{UniSense}} \\

UniSense-2B
& 79.5 & 2247.9 & 79.6 \\

UniSense-9B
& \textbf{83.6} & \textbf{2466.0} & \textbf{87.5} \\

\bottomrule
\end{tabular}
}
\end{subtable}
\hfill
\begin{subtable}[t]{0.38\textwidth}
\centering
\vspace{0pt}
\caption{Audio-image-text benchmarks.}
\vspace{-5pt}
\label{tab:textvqa_docvqa_s}

\setlength{\tabcolsep}{1.5pt}
\renewcommand{\arraystretch}{1.01}

\resizebox{\linewidth}{!}{
\begin{tabular}{@{}lccc@{}}
\toprule
Model
& TextVQA$_s$
& DocVQA$_s$
& ChartQA$_s$ \\
\midrule

\rowcolor{generalcolor}
\multicolumn{4}{c}{\textit{General MLLMs}} \\

Intern-Omni
& 69.1 & 79.9 & 56.0 \\

Qwen3-Omni-30B-A3B
& 70.8 & 76.4 & -- \\

Lyra-Mini-3B
& 73.4 & 74.8 & 40.7 \\

Lyra-Base-9B
& 80.0 & 85.5 & 61.0 \\

\midrule
\rowcolor{ourscolor}
\multicolumn{4}{c}{\textit{UniSense}} \\

UniSense-2B
& 73.8 & 80.0 & 53.8 \\

UniSense-9B
& \textbf{80.4} & \textbf{87.2} & \textbf{68.0} \\

\bottomrule
\end{tabular}
}
\end{subtable}
\hfill
\begin{subtable}[t]{0.26\textwidth}
\centering
\vspace{0pt}
\caption{Video-text benchmarks.}
\vspace{-5pt}
\label{tab:general_video_comparison}

\setlength{\tabcolsep}{1.8pt}
\renewcommand{\arraystretch}{0.96}

\resizebox{\linewidth}{!}{
\begin{tabular}{@{}lcc@{}}
\toprule
Model
& Video-MME
& EgoSchema \\
\midrule

\rowcolor{generalcolor}
\multicolumn{3}{c}{\textit{General MLLMs}} \\
InternVL2.5-7B
& 64.2 & 51.5 \\

InternVL3.5-2B
& 58.4 & 50.8 \\

InternVideo2.5-7B
& 65.1 & 63.9 \\

Lyra-Base-9B
& 62.8 & 63.2 \\

Qwen2-VL-2B      
& 55.6 & 54.9 \\

Qwen2.5-VL-7B
& 65.1 & 65.0 \\

\midrule
\rowcolor{ourscolor}
\multicolumn{3}{c}{\textit{UniSense}} \\

UniSense-2B
& 59.4 & 49.8 \\

UniSense-9B
& \textbf{69.8} & \textbf{66.4} \\

\bottomrule
\end{tabular}
}
\end{subtable}
\vspace{-1.5mm}
\end{table*}
\subsection{Experimental Setup}
\label{sec:setup}
Due to space limitations, we provide additional experimental details in the appendix:
\begin{itemize}
    \item \textbf{Datasets and benchmarks.}
    Details of \textbf{\method--1M}, including data sources and preparation,  together with benchmark splits, evaluation protocols, and metrics, are provided in Appendix~\S\ref{app:datasets_and_benchmarks}.
    \item \textbf{Implementation and training.}
    Model configurations, optimization settings, freezing schedules, and training hyperparameters are provided in Appendix~\S\ref{app:implementation_detatils}.
\end{itemize}

\subsection{Main Results}
\label{sec:main}
\begin{table*}[t]
\centering
\caption{
\textbf{Evaluation on Depth-Text benchmarks,}
including SpatialBench and RoboSpatial-Home.
SpatialBench has Position, Existence, Counting, Reaching, and Size subtask.
\textit{4-Task Agg.} is the sample-weighted aggregate over all subtasks except Reaching.
UniSense is evaluated zero-shot on RoboSpatial-Home.
}
\vspace{-8pt}
\label{tab:depth_spatial}

\fontsize{8.0pt}{9.0pt}\selectfont
\setlength{\tabcolsep}{3pt}
\renewcommand{\arraystretch}{0.97}

\begin{tabular*}{\textwidth}{
@{\extracolsep{\fill}}
l|ccccccc|c
@{}
}
\toprule

& \multicolumn{7}{c|}{SpatialBench}
& RoboSpatial-Home \\

Model
& Pos.
& Exist.
& Count.
& Reach.
& Size
& \makecell{4-Task Agg.}
& Agg.
&  \makecell{Config.}\\
\midrule

\rowcolor{closedcolor}
\multicolumn{9}{c}{\textit{Closed-Source Models}} \\

GPT-4o
& 61.8 & 90.0 & 85.2 & 51.7 & 40.0
& 68.1 & 65.3
& 77.2 \\


Claude-3.5
& 55.9 & 65.0 & 72.2 & -- & 26.7
& 55.1 & --
& -- \\

\rowcolor{generalcolor}
\multicolumn{9}{c}{\textit{General MLLMs}} \\

LLaVA-1.5-7B
& 44.1 & 45.0 & 82.8 & -- & 30.0
& 49.5 & --
& -- \\

LLaVA-NeXT-7B
& 47.1 & 75.0 & 84.0 & -- & 20.0
& 55.1 & --
& 68.3 \\

Qwen2-VL-7B
& -- & -- & -- & -- & --
& -- & --
& 76.4 \\

Qwen2.5-VL-3B
& 55.9 & 80.0 & 76.4 & -- & 25.0
& 58.8 & --
& -- \\

Qwen2.5-VL-7B
& 61.8 & 80.0 & 87.1 & -- & 30.0
& 64.3 & --
& 78.1 \\


\addlinespace[2pt]

\rowcolor{specializedcolor}
\multicolumn{9}{c}{\textit{Spatial-Specialized Models}} \\

SpatialVLM-3B
& 52.9 & 80.0 & 77.1 & -- & 28.3
& 58.6 & --
& -- \\

Bunny-Phi2-3B
& 50.0 & 75.0 & 89.4 & 51.7 & 26.7
& 58.7 & 57.5
& -- \\

SpatialBot-Phi2-3B
& 61.8 & 80.0 & \textbf{91.7} & 55.0 & 26.7
& 64.6 & 62.9
& -- \\

Bunny-Llama3-8B
& 50.0 & 75.0 & \textbf{91.7} & 38.3 & 23.3
& 58.5 & 55.0
& -- \\

SpatialBot-Llama3-8B
& 53.0 & 75.0 & 90.4 & 45.0 & 20.0
& 58.6 & 56.2
& -- \\

SpaceThinker-3B
& 50.0 & 82.5 & 70.0 & 57.5 & 22.5
& 55.3 & 56.5
& -- \\

SpaceOm
& 67.7 & \textbf{92.5} & 75.0 & 55.0 & \textbf{37.5}
& 68.1 & 65.8
& -- \\


SSR-7B
& 64.7 & 85.0 & 90.2 & -- & 28.3
& 66.7 & --
& -- \\


VPT
& -- & -- & -- & -- & --
& -- & --
& 77.2 \\

SpatialVTS
& -- & -- & -- & -- & --
& -- & --
& 70.7 \\


\midrule

\rowcolor{ourscolor}
\multicolumn{9}{c}{\textit{UniSense}} \\

UniSense-2B
& 67.7 & 80.0 & 89.2 & 58.3 & 31.7
& 67.2 & 65.5
& 73.2 \\

UniSense-9B
& \textbf{76.5} & 80.0 & 91.1 & \textbf{58.3} & 28.3
& \textbf{70.1} & \textbf{68.0}
& \textbf{80.5} \\

\bottomrule
\end{tabular*}
\vspace{-2mm}
\end{table*}

\begin{table*}[t]
\centering
\vspace{-2mm}
\begin{minipage}[t]{0.58\textwidth}
\centering
\captionof{table}{
\textbf{Evaluation on the Point Cloud-Text Benchmark} on ScanQA. Validation results with point-cloud (xyz), image (img), and multi-view image (MV-Img) inputs.
}
\label{tab:pointcloud_understanding}
\label{tab:scanqa_val}
\vspace{-2mm}

\fontsize{7.5pt}{8.2pt}\selectfont
\setlength{\tabcolsep}{1.4pt}
\renewcommand{\arraystretch}{0.92}

\begin{tabularx}{\linewidth}{
@{}
M{2.0}
N{0.95}
N{0.81}
N{0.81}
N{0.81}
N{0.81}
N{0.81}
@{}
}
\toprule
Model
& Input
& CIDEr
& MET.
& R-L
& B-1
& B-4 \\
\midrule

\rowcolor{generalcolor}
\multicolumn{7}{c}{\textit{General MLLMs}} \\
InternVL2-8B & img     & 62.5 & 14.5 & 34.3 & --   & 3.3 \\
Qwen2-VL-7B  & img     & 53.9 & 11.4 & 29.3 & 27.8 & 3.0 \\
LLaVA-Video  & MV-Img  & 88.7 & 17.7 & 44.6 & --   & 3.1 \\

\rowcolor{specializedcolor}
\multicolumn{7}{c}{\textit{3D-Specialized Models}} \\
ScanQA        & xyz     & 64.9 & 13.1 & 33.3 & 30.2 & 10.1 \\
LL3DA         & xyz     & 76.8 & 15.9 & 37.3 & --   & 13.5 \\
Scene-LLM     & xyz     & 80.0 & 16.6 & 40.0 & --   & 12.0 \\
ChatScene     & xyz+img & 87.7 & 18.0 & 41.6 & 43.2 & 14.3 \\
GPT4Scene     & MV-Img  & 90.9 & 17.7 & 43.6 & 43.4 & 14.6 \\
ViewMind3D
              & MV-Img & 73.4 & 15.3 & 37.6 & 35.8 & -- \\

\midrule
\rowcolor{ourscolor}
\multicolumn{7}{c}{\textit{UniSense}} \\
UniSense-2B
& xyz
& 65.1 & 21.1 & 34.0 & 36.4 & 11.0 \\

UniSense-2B
& xyz+img
& 90.0 & 27.5 & 44.6 & 44.3 & 12.7 \\

UniSense-9B
& xyz
& 74.3 & 23.4 & 36.9 & 39.2 & 12.4 \\

UniSense-9B
& xyz+img
& \textbf{99.6} & \textbf{30.2} & \textbf{47.5}
& \textbf{48.3} & \textbf{15.5} \\

\bottomrule
\end{tabularx}
\end{minipage}
\hfill
\begin{minipage}[t]{0.4\textwidth}
\centering

\captionof{table}{
\textbf{Evaluation on the Tactile-Text Benchmark} using Octopi.
Results are reported in percentage (\%) for three tasks.
}
\label{tab:octopi_tactile}

\vspace{-2mm}

\fontsize{7.5pt}{9.3pt}\selectfont
\setlength{\tabcolsep}{1.8pt}
\renewcommand{\arraystretch}{0.91}

\begin{tabularx}{\linewidth}{
@{}
M{1.9}
N{0.7}
N{0.7}
N{0.7}
@{}
}
\toprule
Model
& PC
& PSS
& POM \\
\midrule

\rowcolor{closedcolor}
\multicolumn{4}{c}{\textit{Closed-Source Models}} \\
GPT-4o          & 30.9 & 38.1 & 22.6 \\
Gemini-2.5-Pro  & 26.7 & 25.1 & 23.5 \\

\rowcolor{generalcolor}
\multicolumn{4}{c}{\textit{General MLLMs}} \\
Qwen2.5-VL-7B         & 21.5 & 23.1 & 19.9 \\
LLaVA-OneVision-7B    & 35.4 & 28.4 & 29.1 \\
TokenFlow-7B          & 38.1 & 32.6 & 29.5 \\

\rowcolor{specializedcolor}
\multicolumn{4}{c}{\textit{Tactile-Specialized Models}} \\
Octopi-7B  & 45.5 & 48.0 & 22.2 \\
UniTac-3B  & 55.0 & 58.9 & 42.1 \\
UniTac-7B  & 57.3 & 59.2 & \textbf{64.6} \\

\midrule
\rowcolor{ourscolor}
\multicolumn{4}{c}{\textit{UniSense}}\\
UniSense-2B
& 58.4 & 72.8 & 45.5 \\

UniSense-9B
& \textbf{61.9} & \textbf{73.4} & 53.3 \\

\bottomrule
\end{tabularx}
\end{minipage}
\vspace{-2mm}
\end{table*}

\begin{table*}[t]
\centering

\begin{minipage}[t]{0.56\textwidth}
\centering
\captionof{table}{
\textbf{Evaluation on IMU-Text Benchmarks}, including
HHAR, MotionSense, Shoaib, and UCI.
Each entry reports F1 / Precision (\%).
}
\vspace{-2mm}
\label{tab:imu_har_results}

\scriptsize
\setlength{\tabcolsep}{2.8pt}
\renewcommand{\arraystretch}{0.97}

\resizebox{\linewidth}{!}{
\begin{tabular}{@{}lcccc@{}}
\toprule
Model
& HHAR
& MotionSense
& Shoaib
& UCI \\
\midrule

\rowcolor{closedcolor}
\multicolumn{5}{c}{\textit{Closed-Source Models}} \\

GPT-4o
& 7.0 / 14.0
& 8.0 / 26.0
& 14.0 / 35.0
& 4.0 / 16.0 \\

\rowcolor{generalcolor}
\multicolumn{5}{c}{\textit{General LLMs}} \\

Vicuna-7B
& 5.0 / 12.0
& 9.0 / 15.0
& 8.0 / 11.0
& 4.0 / 25.0 \\

\rowcolor{specializedcolor}
\multicolumn{5}{c}{\textit{IMU-Specialized Models}} \\

GenHAR
& 63.0 / --
& 81.0 / --
& 74.0 / --
& 79.0 / -- \\

LanHAR
& 70.0 / --
& 76.0 / --
& 71.0 / --
& 75.0 / -- \\

LLaSA-7B
& 84.0 / 88.0
& 83.0 / 85.0
& \textbf{81.0} / 84.0
& 72.0 / 82.0 \\

\midrule

\rowcolor{ourscolor}
\multicolumn{5}{c}{\textit{UniSense}} \\

UniSense-2B
& 92.0 / 92.0
& 75.0 / 79.0
& 72.0 / 76.0
& 87.0 / 87.0 \\

UniSense-9B
& \textbf{97.0 / 97.0}
& \textbf{87.0 / 89.0}
& \textbf{81.0 / 85.0}
& \textbf{90.0 / 90.0} \\

\bottomrule
\end{tabular}
}

\end{minipage}
\hfill
%
\begin{minipage}[t]{0.415\textwidth}
\centering
\captionof{table}{
\textbf{Evaluation on Thermal-Text Benchmarks} using Infrared-Benchmark.
Average accuracy is reported under zero-shot and fine-tuning settings.
}
\vspace{-2mm}
\label{tab:infrared_template_benchmark}
\label{tab:thermal_results}

\scriptsize
\setlength{\tabcolsep}{5.0pt}
\renewcommand{\arraystretch}{1.09}

\resizebox{\linewidth}{!}{
\begin{tabular}{@{}lcc@{}}
\toprule
Model
& Zero-Shot
& Fine-Tuning \\
\midrule

\rowcolor{generalcolor}
\multicolumn{3}{c}{\textit{General MLLMs}} \\

ImageBindLLM-7B
& 24.2
& 48.5 \\

PandaGPT-7B
& 26.8
& 41.3 \\

PandaGPT-13B
& 34.3
& 46.0 \\

\rowcolor{specializedcolor}
\multicolumn{3}{c}{\textit{Thermal-Specialized Models}} \\

Infrared-LLaVA-7B
& 56.6
& 79.0 \\

\midrule

\rowcolor{ourscolor}
\multicolumn{3}{c}{\textit{UniSense}} \\

UniSense-2B
& --
& 86.8 \\

UniSense-9B
& \textbf{70.2}
& \textbf{89.9} \\

\bottomrule
\end{tabular}
}

\end{minipage}
\vspace{-2.5mm}
\end{table*}

\begin{table*}[t]
\centering
\caption{
\textbf{Evaluation on Radar-Text, Motion-Text, and Smell-Text Benchmarks.}
(a) Radar results on HoloLLM's language-grounded MM-Fi benchmark under the Cross-Environment (CrossEnv) split; Human Action QA and Action Caption use accuracy and METEOR, respectively.
(b) Motion results on QuAnTS; binary and multiple-choice QA use F1, while open-ended QA uses ROUGE-L and METEOR. TS, GT, and AE denote motion time series, ground-truth actions, and action-encoder representations, respectively.
(c) Smell classification on SmellNet using Acc@1 and Macro-F1 (\%).
}
\label{tab:radar_motion_smell}
\vspace{-2mm}
\begin{subtable}[t]{0.27\textwidth}
\centering
\caption{Radar.}
\vspace{-1.5mm}
\label{tab:radar_crossenv}

\scriptsize
\setlength{\tabcolsep}{2.5pt}
\renewcommand{\arraystretch}{1.0}

\resizebox{\linewidth}{!}{
\begin{tabular}{@{}lcc@{}}
\toprule
Model
& \makecell{Human\\Action QA}
& \makecell{Action\\Caption}\\
\midrule

\rowcolor{generalcolor}
\multicolumn{3}{c}{\textit{General MLLMs}} \\

Tokenpacker & 4.7  & 6.0  \\
Honeybee    & 1.5  & 10.4 \\
OneLLM      & 1.1  & 2.2  \\
ImageBind   & 24.0 & 17.5 \\

\rowcolor{specializedcolor}
\multicolumn{3}{c}{\textit{Radar-Specialized Models}} \\

HoloLLM & 61.4 & 24.5 \\

\midrule

\rowcolor{ourscolor}
\multicolumn{3}{c}{\textit{UniSense}} \\

UniSense-2B & 61.0 & 46.0 \\
UniSense-9B & \textbf{65.2} & \textbf{46.1} \\

\bottomrule
\end{tabular}}
\end{subtable}
\hfill
%
\begin{subtable}[t]{0.43\textwidth}
\centering
\caption{Motion.}
\vspace{-1.5mm}
\label{tab:quants}

\scriptsize
\setlength{\tabcolsep}{2.8pt}
\renewcommand{\arraystretch}{1.02}

\resizebox{\linewidth}{!}{
\begin{tabular}{@{}llcccc@{}}
\toprule
Model
& Input
& Bin. F1
& Mul. F1
& R-L
& MET. \\
\midrule

\rowcolor{specializedcolor}
\multicolumn{6}{c}{\textit{Motion-Specialized Models}} \\

Naive
& TS
& 35.6 & 9.6 & 6.0 & 5.8 \\

ChatTS
& TS
& 34.8 & 29.1 & 4.6 & 10.6 \\

xQA-Llama
& GT
& 78.2 & 81.5 & 9.1 & 36.8 \\

xQA-Qwen
& GT
& 78.4 & 88.0 & 17.9 & 37.0 \\

xQA-Llama
& AE
& 78.3 & 81.2 & 9.0 & 36.8 \\

xQA-Qwen
& AE
& 78.5 & 88.0 & 17.8 & 36.9 \\

\midrule

\rowcolor{ourscolor}
\multicolumn{6}{c}{\textit{UniSense}} \\

UniSense-2B
& TS
& 89.7 & 97.3 & 82.1 & 72.3 \\

UniSense-9B
& TS
& \textbf{97.4}
& \textbf{99.6}
& \textbf{85.0}
& \textbf{74.2} \\

\bottomrule
\end{tabular}
}
\end{subtable}
\hfill
%
\begin{subtable}[t]{0.25\textwidth}
\centering
\caption{Smell.}
\vspace{-1.5mm}
\label{tab:smell_results}

\scriptsize
\setlength{\tabcolsep}{2.5pt}
\renewcommand{\arraystretch}{1.15}

\resizebox{\linewidth}{!}{
\begin{tabular}{@{}lcc@{}}
\toprule
Model
& Acc@1
& F1 \\
\midrule

\rowcolor{specializedcolor}
\multicolumn{3}{c}{\textit{Smell-Specialized Models}} \\

MLP
& 26.8 & 24.7 \\

CNN
& 52.7 & 50.5 \\

LSTM
& 57.9 & 56.0 \\

ScentFormer
& 56.1 & 55.5 \\

\midrule

\rowcolor{ourscolor}
\multicolumn{3}{c}{\textit{UniSense}} \\

UniSense-2B
& 65.2 & 63.4 \\

UniSense-9B
& \textbf{74.5}
& \textbf{72.2} \\

\bottomrule
\end{tabular}
}
\end{subtable}
\vspace{-3mm}
\end{table*}
In this section, we evaluate \method on general multimodal tasks and benchmarks spanning 8 embodied sensory modalities. These evaluations examine both its general multimodal capability and its ability to handle heterogeneous physical sensing signals. Unless otherwise specified, higher scores indicate better performance across all reported metrics.

\noindent\textbf{General Multimodal Understanding.}
\method-9B achieves 83.6 on TextVQA~\citep{singh2019towards} and 2466.0 on MME~\citep{NEURIPS2025_d79a27cf}, leading the compared models on both benchmarks . It also performs strongly on speech-conditioned image and video understanding (Table~\ref{tab:general_capability}), indicating that sensory training preserves its basic-modality capabilities.

\noindent\textbf{Depth and Point Cloud Understanding.}
Tables~\ref{tab:depth_spatial} and~\ref{tab:pointcloud_understanding} evaluate spatial and 3D understanding.
\method-9B achieves 70.1 on the SpatialBench~\citep{cai2024spatialbot} 4-task aggregate and 80.5 accuracy on RoboSpatial-Home~\citep{song2025robospatial} without benchmark-specific training, surpassing both general-purpose MLLMs and spatial-specialized models.
On ScanQA~\citep{azuma2022scanqa} , \method-9B achieves 74.3 CIDEr using point clouds alone, which increases to 99.6 when combined with images, outperforming all compared baselines. This improvement shows that newly introduced sensory inputs can be naturally composed with basic modality observations to exploit complementary geometric and appearance information.

\noindent\textbf{IMU, Radar, and Motion Understanding.}
Across four IMU activity-recognition datasets, \method-9B outperforms LLaSA~\citep{imran2024llasa} on three tasks and matches its F1 on Shoaib (Table~\ref{tab:imu_har_results}).
On radar, it outperforms HoloLLM~\citep{zhou2026holollm} in both action recognition and captioning, with a 21.6-point gain in caption METEOR (Table~\ref{tab:radar_crossenv}).
On QuAnTS~\citep{divo2025quants}, it answers motion questions directly from skeleton sequences, outperforming baselines given ground-truth actions or pretrained action features (Table~\ref{tab:quants}).

\noindent\textbf{Tactile, Thermal, and Smell Understanding.}
We further test \method on tactile, smell, and thermal modalities in Tables~\ref{tab:octopi_tactile}, ~\ref{tab:smell_results}, and~\ref{tab:thermal_results}.
On Octopi~\citep{yu2024octopi}, \method-9B leads on property comparison and superlative selection, reaching 61.9 and 73.4. On Infrared-Benchmark~\citep{jiang2024infrared} , it reaches 70.2 zero-shot and 89.9 after fine-tuning, outperforming Infrared-LLaVA~\citep{jiang2024infrared} by 13.6 and 10.9 points. For smell recognition~\citep{feng2026smellnet}, \method-9B reaches 74.5 Acc@1 and 72.2 Macro F1, surpassing specialized recognition models.

\vspace{-2mm}
\subsection{Ablation Studies}
\label{sec:ablation}
\begin{table*}[t]
\centering

\caption{
\textbf{Component ablation of UniSense.}
Starting from the baseline, we progressively introduce sensory representation alignment and modality-specific expert tuning.
Results are reported on representative benchmarks spanning depth, point cloud, IMU, thermal, and radar.
}
\label{tab:component_ablation}

\vspace{-2mm}

\fontsize{8pt}{8.2pt}\selectfont
\setlength{\tabcolsep}{2.8pt}
\renewcommand{\arraystretch}{0.80}

\begin{tabular*}{\textwidth}{
@{\extracolsep{\fill}}
lccc|ccccc
@{}
}
\toprule
&
\multicolumn{3}{c|}{\textbf{Method}}
&
\multicolumn{5}{c}{\textbf{Benchmarks}}
\\
\cmidrule(lr){2-4}
\cmidrule(lr){5-9}

Setting
&
Proj.
&
Alignment
&
Experts
&
RoboSpatial-Home
&
ScanQA
&
IMU-4-Task
&
Infrared-Benchmark
&
MMFi-VQA
\\

&
&
&
&
Config.
&
CIDEr
&
Avg. F1
&
Acc.
&
Acc.
\\
\midrule

Baseline
&
\checkmark
&
&
&
74.0
&
96.7
&
86.0
&
88.2
&
60.1
\\

+ Alignment
&
\checkmark
&
\checkmark
&
&
77.2
&
97.7
&
88.0
&
89.0
&
63.5
\\

\textbf{Full}
&
\checkmark
&
\checkmark
&
\checkmark
&
\textbf{80.5}
&
\textbf{99.6}
&
\textbf{89.0}
&
\textbf{89.9}
&
\textbf{65.2}
\\

\bottomrule
\end{tabular*}
\vspace{-3mm}
\end{table*}

\begin{figure*}[t]
    \centering

    \includegraphics[width=1\textwidth]{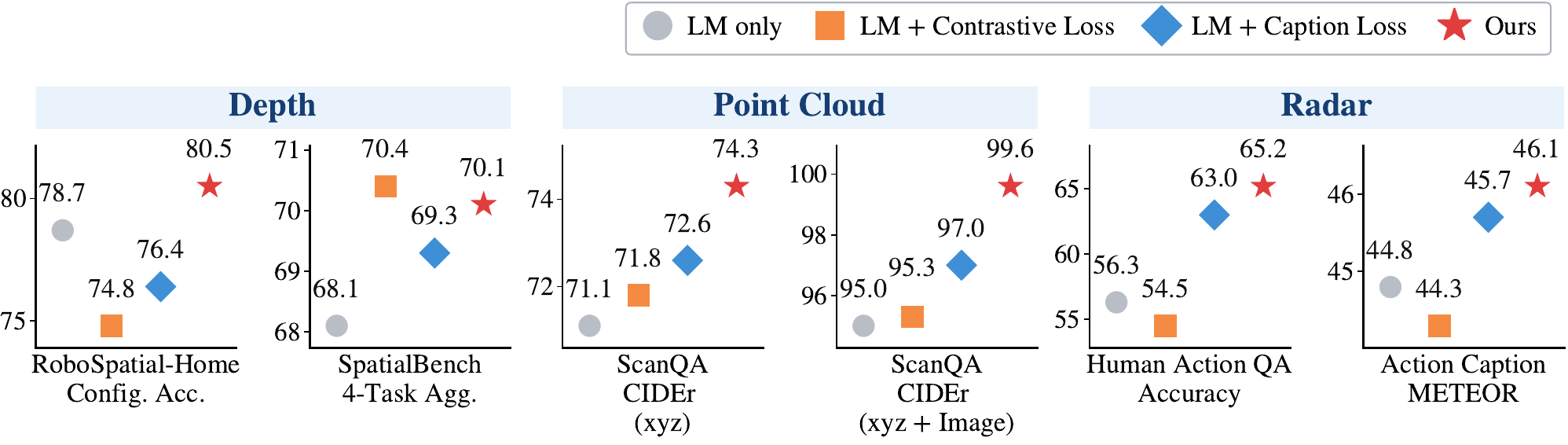}
    
    \vspace{-1.5mm}
    {\small (a)}
    \vspace{-2mm}
    \includegraphics[width=1\textwidth]{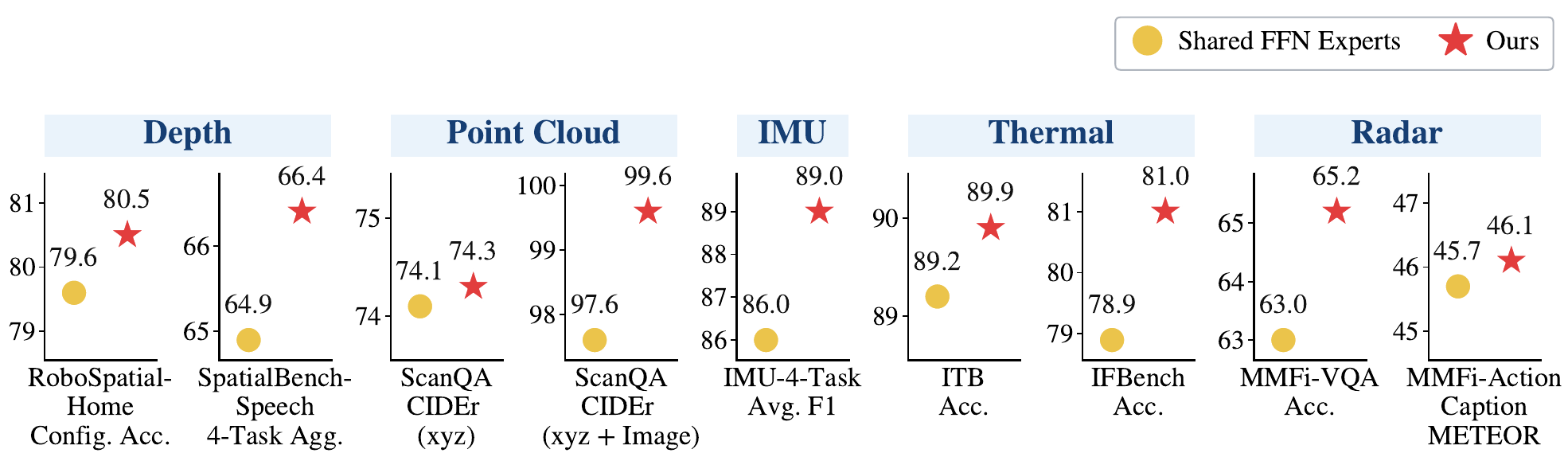}
    
    {\small (b)}
    
    \vspace{-1mm}
    \caption{
     \textbf{Ablation of alignment objectives and expert specialization.}
(a) Comparison of language-modeling supervision, conditional captioning, contrastive alignment, and their combination during sensory representation alignment.
(b) Comparison between a shared expert pool and modality-specific expert pools during expert tuning.
}
    \label{fig:csa_msc_ablation}
    \vspace{-4mm}
\end{figure*}
All ablations use the Qwen3.5-9B backbone with the same training and evaluation settings. 
Additional analyses and detailed configurations are provided in Appendix~\S\ref{app:ablation_settings} and Appendix~\S\ref{app:additional_exp}.

\noindent\textbf{Component Ablation.}
Table~\ref{tab:component_ablation} evaluates the two training stages.
Sensory representation alignment consistently improves the baseline, while modality-specific expert tuning brings further gains across all evaluated benchmarks, showing that language grounding and modality-specific specialization provide complementary benefits.

\noindent\textbf{Alignment Objectives.}
Figure~\ref{fig:csa_msc_ablation}(a) compares conditional captioning and contrastive alignment.
The two objectives show different strengths across modalities, while their combination achieves the best overall performance.
This consistent improvement indicates that the two objectives provide complementary supervision and jointly produce better-aligned sensory representations.

\noindent\textbf{Expert Specialization.}
Figure~\ref{fig:csa_msc_ablation}(b) compares shared and modality-specific expert pools.
Modality-specific experts consistently outperform the shared alternative across all evaluated modalities, supporting specialized computation for heterogeneous sensory inputs.
The gains are particularly evident on SpatialBench-Speech~\citep{cai2024spatialbot} and ScanQA~\citep{azuma2022scanqa} with xyz+Image inputs, suggesting that modality-specific computation reduces interference across heterogeneous modalities and better preserves complementary cross-modal information.

\section{Conclusion}
\vspace{-2mm}
We present \method, a unified MLLM for understanding diverse embodied sensory signals beyond basic multimodal inputs.
By grounding heterogeneous sensory representations in language and introducing modality-specific processing within the shared LLM, \method enables 8 physical sensing modalities to be modeled under a common architecture.
Extensive experiments show that this unified formulation performs strongly across heterogeneous sensory tasks and can effectively integrate sensory signals with existing multimodal inputs.
Taken together, these results suggest that pretrained MLLMs can move beyond serving as interfaces for new modalities, providing a shared foundation for understanding heterogeneous physical signals without treating each modality as an isolated system.
More broadly, our findings point toward embodied multimodal models that share a common reasoning backbone while retaining the specialization required by different sensing processes.

\bibliography{iclr2027_conference}

@inproceedings{driess2023palm,
  title={PaLM-E: An Embodied Multimodal Language Model},
  author={Driess, Danny and Xia, Fei and Sajjadi, Mehdi SM and Lynch, Corey and Chowdhery, Aakanksha and Ichter, Brian and Wahid, Ayzaan and Tompson, Jonathan and Vuong, Quan and Yu, Tianhe and others},
  booktitle={International Conference on Machine Learning},
  pages={8469--8488},
  year={2023},
  organization={PMLR}
}

@inproceedings{robobrain,
  title={Robobrain: A unified brain model for robotic manipulation from abstract to concrete},
  author={Ji, Yuheng and Tan, Huajie and Shi, Jiayu and Hao, Xiaoshuai and Zhang, Yuan and Zhang, Hengyuan and Wang, Pengwei and Zhao, Mengdi and Mu, Yao and An, Pengju and others},
  booktitle={2025 IEEE/CVF Conference on Computer Vision and Pattern Recognition (CVPR)},
  pages={1724--1734},
  year={2025},
  organization={IEEE}
}

@article{robobrain2,
  title={Robobrain 2.0 technical report},
  author={Cao, Mingyu and Tan, Huajie and Ji, Yuheng and Chen, Xiansheng and Lin, Minglan and Li, Zhiyu and Cao, Zhou and Wang, Pengwei and Zhou, Enshen and others},
  journal={arXiv preprint arXiv:2507.02029},
  year={2025}
}

@inproceedings{zhao2025embodied,
  title={Embodied-r: Collaborative framework for activating embodied spatial reasoning in foundation models via reinforcement learning},
  author={Zhao, Baining and Wang, Ziyou and Fang, Jianjie and Gao, Chen and Man, Fanhang and Cui, Jinqiang and Wang, Xin and Chen, Xinlei and Li, Yong and Zhu, Wenwu},
  booktitle={Proceedings of the 33rd ACM International Conference on Multimedia},
  pages={11071--11080},
  year={2025}
}

@article{dang2026rynnbrain,
  title={Rynnbrain: Open embodied foundation models},
  author={Dang, Ronghao and Guo, Jiayan and Hou, Bohan and Leng, Sicong and Li, Kehan and Li, Xin and Liu, Jiangpin and Mao, Yunxuan and Wang, Zhikai and Yuan, Yuqian and others},
  journal={arXiv preprint arXiv:2602.14979},
  year={2026}
}

@article{bjorck2026vesta,
  title={Vesta: A Generalist Embodied Reasoning Model},
  author={Bjorck, Johan and Li, Zhiqi and Man, Yunze and Wang, Jing and Cheng, An-Chieh and Liu, Sifei and Wang, Shihao and Yu, Zhiding and Badki, Abhishek and Birchfield, Stan and others},
  journal={arXiv preprint arXiv:2606.20905},
  year={2026}
}

@article{zhang2026hy,
  title={Hy-embodied-0.5-vla: From vision-language-action models to a real-world robot learning stack},
  author={Zhang, He and Xiang, Lingzhu and Lin, Haitao and Huang, Zeyu and Wang, Minghui and Zhong, Dingyan and Dong, Yubo and Wu, Yihao and Rao, Yongming and Zhang, Dongsheng and others},
  journal={arXiv preprint arXiv:2606.14409},
  year={2026}
}

@inproceedings{majumdar2024openeqa,
  title={Openeqa: Embodied question answering in the era of foundation models},
  author={Majumdar, Arjun and Ajay, Anurag and Zhang, Xiaohan and Putta, Pranav and Yenamandra, Sriram and Henaff, Mikael and Silwal, Sneha and Mcvay, Paul and Maksymets, Oleksandr and Arnaud, Sergio and others},
  booktitle={2024 IEEE/CVF Conference on Computer Vision and Pattern Recognition (CVPR)},
  pages={16488--16498},
  year={2024},
  organization={IEEE}
}

@article{dissanayake2025good,
  title={How Good are Foundation Models in Step-by-Step Embodied Reasoning?},
  author={Dissanayake, Dinura and Heakl, Ahmed and Thawakar, Omkar and Ahsan, Noor and Thawkar, Ritesh and More, Ketan and Lahoud, Jean and Anwer, Rao and Cholakkal, Hisham and Laptev, Ivan and others},
  journal={arXiv preprint arXiv:2509.15293},
  year={2025}
}

@inproceedings{kim2025openvla,
  title={OpenVLA: An Open-Source Vision-Language-Action Model},
  author={Kim, Moo Jin and Pertsch, Karl and Karamcheti, Siddharth and Xiao, Ted and Balakrishna, Ashwin and Nair, Suraj and Rafailov, Rafael and Foster, Ethan P and Sanketi, Pannag R and Vuong, Quan and others},
  booktitle={Conference on Robot Learning},
  pages={2679--2713},
  year={2025},
  organization={PMLR}
}

@article{black2024pi_0,
  title={$\pi\_0$: A Vision-Language-Action Flow Model for General Robot Control},
  author={Black, Kevin and Brown, Noah and Driess, Danny and Esmail, Adnan and Equi, Michael and Finn, Chelsea and Fusai, Niccolo and Groom, Lachy and Hausman, Karol and Ichter, Brian and others},
  journal={arXiv preprint arXiv:2410.24164},
  year={2024}
}

@inproceedings{chen2024octavius,
  title={Octavius: Mitigating task interference in mllms via lora-moe},
  author={Chen, Zeren and Liu, Huayang and Yin, Zhenfei and Liu, Si and Sheng, Lyu and Ouyang, Wanli and Shao, Jing and others},
  booktitle={International Conference on Learning Representations},
  volume={2024},
  pages={24652--24673},
  year={2024}
}

@inproceedings{hao2026uni,
  title={Uni-X: mitigating modality conflict with a two-end-separated architecture for unified multimodal models},
  author={Hao, Jitai and Liu, Hao and Xiao, Xinyan and Huang, Qiang and Yu, Jun},
  booktitle={International Conference on Learning Representations},
  volume={2026},
  pages={123737--123758},
  year={2026}
}

@article{li2025unimoe,
  author  = {Li, Yunxin and Jiang, Shenyuan and Hu, Baotian and Wang, Longyue
             and Zhong, Wanqi and Luo, Wenhan and Ma, Lin and Zhang, Min},
  title   = {Uni-MoE: Scaling Unified Multimodal LLMs With Mixture of Experts},
  journal = {IEEE Transactions on Pattern Analysis and Machine Intelligence},
  year    = {2025},
  volume  = {47},
  number  = {5},
  pages   = {3424--3439},
  doi     = {10.1109/TPAMI.2025.3532688}
}

@misc{mingomni2025,
  title         = {Ming-Omni: A Unified Multimodal Model for Perception and Generation},
  author        = {{Inclusion AI}},
  year          = {2025},
  eprint        = {2506.09344},
  archivePrefix = {arXiv},
  primaryClass  = {cs.AI}
}

@inproceedings{shukor2025scaling,
  author    = {Shukor, Mustafa and Fini, Enrico and
               da Costa, Victor Guilherme Turrisi and Cord, Matthieu
               and Susskind, Joshua and El-Nouby, Alaaeldin},
  title     = {Scaling Laws for Native Multimodal Models},
  booktitle = {Proceedings of the IEEE/CVF International Conference on Computer Vision (ICCV)},
  year      = {2025},
  pages     = {12--23}
}

@inproceedings{radford2021clip,
  title={Learning transferable visual models from natural language supervision},
  author={Radford, Alec and Kim, Jong Wook and Hallacy, Chris and Ramesh, Aditya and Goh, Gabriel and Agarwal, Sandhini and Sastry, Girish and Askell, Amanda and Mishkin, Pamela and Clark, Jack and others},
  booktitle={International conference on machine learning},
  pages={8748--8763},
  year={2021},
  organization={PmLR}
}

@article{tschannen2025siglip,
  title={Siglip 2: Multilingual vision-language encoders with improved semantic understanding, localization, and dense features},
  author={Tschannen, Michael and Gritsenko, Alexey and Wang, Xiao and Naeem, Muhammad Ferjad and Alabdulmohsin, Ibrahim and Parthasarathy, Nikhil and Evans, Talfan and Beyer, Lucas and Xia, Ye and Mustafa, Basil and others},
  journal={arXiv preprint arXiv:2502.14786},
  year={2025}
}

@article{chu2024qwen2audio,
  title={Qwen2-Audio Technical Report},
  author={Chu, Yunfei and Xu, Jin and Yang, Qian and Wei, Haojie and Wei, Xipin and Guo, Zhifang and Leng, Yichong and Lv, Yuanjun and He, Jinzheng and Lin, Junyang and Zhou, Chang and Zhou, Jingren},
  journal={arXiv preprint arXiv:2407.10759},
  year={2024}
}

@article{yu2022coca,
  title={Coca: Contrastive captioners are image-text foundation models},
  author={Yu, Jiahui and Wang, Zirui and Vasudevan, Vijay and Yeung, Legg and Seyedhosseini, Mojtaba and Wu, Yonghui},
  journal={arXiv preprint arXiv:2205.01917},
  year={2022}
}

@inproceedings{han2024onellm,
  title={Onellm: One framework to align all modalities with language},
  author={Han, Jiaming and Gong, Kaixiong and Zhang, Yiyuan and Wang, Jiaqi and Zhang, Kaipeng and Lin, Dahua and Qiao, Yu and Gao, Peng and Yue, Xiangyu},
  booktitle={Proceedings of the IEEE/CVF Conference on Computer Vision and Pattern Recognition},
  pages={26584--26595},
  year={2024}
}

@article{zhou2026holollm,
  title={Holollm: Multisensory foundation model for language-grounded human sensing and reasoning},
  author={Zhou, Chuhao and Yang, Jianfei},
  journal={Advances in Neural Information Processing Systems},
  volume={38},
  pages={73543--73568},
  year={2026}
}

@inproceedings{li2025sensorllm,
  title={Sensorllm: Aligning large language models with motion sensors for human activity recognition},
  author={Li, Zechen and Deldari, Shohreh and Chen, Linyao and Xue, Hao and Salim, Flora D},
  booktitle={Proceedings of the 2025 Conference on Empirical Methods in Natural Language Processing},
  pages={354--379},
  year={2025}
}

@inproceedings{liang2025mixture,
  title={Mixture-of-Transformers: A Sparse and Scalable Architecture for Multi-Modal Foundation Models},
  author={Liang, Weixin and YU, LILI and Luo, Liang and Iyer, Srini and Dong, Ning and Zhou, Chunting and Ghosh, Gargi and Lewis, Mike and Yih, Wen-tau and Zettlemoyer, Luke and others},
  booktitle={ICLR 2025 Workshop on World Models: Understanding, Modelling and Scaling},
  year={2025}
}

@inproceedings{zhong2025lyra,
  title={Lyra: An efficient and speech-centric framework for omni-cognition},
  author={Zhong, Zhisheng and Wang, Chengyao and Liu, Yuqi and Yang, Senqiao and Tang, Longxiang and Zhang, Yuechen and Li, Jingyao and Qu, Tianyuan and Li, Yanwei and Chen, Yukang and others},
  booktitle={Proceedings of the IEEE/CVF International Conference on Computer Vision},
  pages={3694--3704},
  year={2025}
}

@inproceedings{zhou2025unialign,
  title={Unialign: Scaling multimodal alignment within one unified model},
  author={Zhou, Bo and Li, Liulei and Wang, Yujia and Liu, Huafeng and Yao, Yazhou and Wang, Wenguan},
  booktitle={Proceedings of the Computer Vision and Pattern Recognition Conference},
  pages={29644--29655},
  year={2025}
}

@inproceedings{girdhar2023imagebind,
  title={Imagebind: One embedding space to bind them all},
  author={Girdhar, Rohit and El-Nouby, Alaaeldin and Liu, Zhuang and Singh, Mannat and Alwala, Kalyan Vasudev and Joulin, Armand and Misra, Ishan},
  booktitle={Proceedings of the IEEE/CVF conference on computer vision and pattern recognition},
  pages={15180--15190},
  year={2023}
}

@inproceedings{su2023pandagpt,
  title={Pandagpt: One model to instruction-follow them all},
  author={Su, Yixuan and Lan, Tian and Li, Huayang and Xu, Jialu and Wang, Yan and Cai, Deng},
  booktitle={Proceedings of the 1st Workshop on Taming Large Language Models: Controllability in the Era of Interactive Assistants!},
  pages={11--23},
  year={2023}
}

@inproceedings{zhu2024languagebind,
  title={Languagebind: Extending video-language pretraining to n-modality by language-based semantic alignment},
  author={Zhu, Bin and Lin, Bin and Ning, Munan and Yan, Yang and Cui, Jiaxi and HongFa, WANG and Pang, Yatian and Jiang, Wenhao and Zhang, Junwu and Li, Zongwei and others},
  booktitle={International Conference on Learning Representations},
  volume={2024},
  pages={9588--9608},
  year={2024}
}

@inproceedings{moon2024anymal,
  title={Anymal: An efficient and scalable any-modality augmented language model},
  author={Moon, Seungwhan and Madotto, Andrea and Lin, Zhaojiang and Nagarajan, Tushar and Smith, Matt and Jain, Shashank and Yeh, Chun-Fu and Murugesan, Prakash and Heidari, Peyman and Liu, Yue and others},
  booktitle={Proceedings of the 2024 Conference on Empirical Methods in Natural Language Processing: Industry Track},
  pages={1314--1332},
  year={2024}
}

@inproceedings{wu2024next,
  title={NExT-GPT: any-to-any multimodal LLM},
  author={Wu, Shengqiong and Fei, Hao and Qu, Leigang and Ji, Wei and Chua, Tat-Seng},
  booktitle={Proceedings of the 41st International Conference on Machine Learning},
  pages={53366--53397},
  year={2024}
}

@inproceedings{zhang2025omnibind,
  title={Omnibind: Large-scale omni multimodal representation via binding spaces},
  author={Zhang, Ziang and Hong, Minjie and Zhang, Hang and Liu, Luping and Huang, Rongjie and Cheng, Xize and Ji, Shengpeng and Jin, Tao and Zhao, Hengshuang and Zhao, Zhou and others},
  booktitle={International Conference on Learning Representations},
  volume={2025},
  pages={20831--20851},
  year={2025}
}

@inproceedings{xu2024pointllm,
  title={PointLLM: Empowering Large Language Models to Understand Point Clouds},
  author={Xu, Runsen and Wang, Xiaolong and Wang, Tai and Chen, Yilun and Pang, Jiangmiao and Lin, Dahua},
  booktitle={European Conference on Computer Vision},
  pages={131--147},
  year={2024}
}

@inproceedings{lai2026radarllm,
  title={Radarllm: empowering large language models to understand human motion from millimeter-wave point cloud sequence},
  author={Lai, Zengyuan and Yang, Jiarui and Xia, Songpengcheng and Lin, Lizhou and Sun, Lan and Wang, Renwen and Liu, Jianran and Wu, Qi and Pei, Ling},
  booktitle={Proceedings of the AAAI Conference on Artificial Intelligence},
  pages={5791--5799},
  year={2026}
}

@inproceedings{fu2024touch,
  title={A Touch, Vision, and Language Dataset for Multimodal Alignment},
  author={Fu, Letian and Datta, Gaurav and Huang, Huang and Panitch, William Chung-Ho and Drake, Jaimyn and Ortiz, Joseph and Mukadam, Mustafa and Lambeta, Mike and Calandra, Roberto and Goldberg, Ken},
  booktitle={International Conference on Machine Learning},
  pages={14080--14101},
  year={2024},
  organization={PMLR}
}

@article{liu2023visual,
  title={Visual instruction tuning},
  author={Liu, Haotian and Li, Chunyuan and Wu, Qingyang and Lee, Yong Jae},
  journal={Advances in neural information processing systems},
  volume={36},
  pages={34892--34916},
  year={2023}
}

@inproceedings{li2023blip,
  title={Blip-2: Bootstrapping language-image pre-training with frozen image encoders and large language models},
  author={Li, Junnan and Li, Dongxu and Savarese, Silvio and Hoi, Steven},
  booktitle={International conference on machine learning},
  pages={19730--19742},
  year={2023},
  organization={PMLR}
}

@article{dai2023instructblip,
  title={Instructblip: Towards general-purpose vision-language models with instruction tuning},
  author={Dai, Wenliang and Li, Junnan and Li, Dongxu and Tiong, Anthony and Zhao, Junqi and Wang, Weisheng and Li, Boyang and Fung, Pascale N and Hoi, Steven},
  journal={Advances in neural information processing systems},
  volume={36},
  pages={49250--49267},
  year={2023}
}

@article{team2024chameleon,
  title={Chameleon: Mixed-modal early-fusion foundation models},
  author={Chameleon},
  journal={arXiv preprint arXiv:2405.09818},
  year={2024}
}

@article{lin2026moe,
  title={Moe-llava: Mixture of experts for large vision-language models},
  author={Lin, Bin and Tang, Zhenyu and Ye, Yang and Huang, Jinfa and Zhang, Junwu and Pang, Yatian and Jin, Peng and Ning, Munan and Luo, Jiebo and Yuan, Li},
  journal={IEEE Transactions on Multimedia},
  year={2026},
  publisher={IEEE}
}

@article{shen2024mome,
  title={Mome: Mixture of multimodal experts for generalist multimodal large language models},
  author={Shen, Leyang and Chen, Gongwei and Shao, Rui and Guan, Weili and Nie, Liqiang},
  journal={Advances in neural information processing systems},
  volume={37},
  pages={42048--42070},
  year={2024}
}

@article{shi2026lmfusion,
  title={Lmfusion: Adapting pretrained language models for multimodal generation},
  author={Shi, Weijia and Han, Xiaochuang and Zhou, Chunting and Liang, Weixin and Lin, Xi and Zettlemoyer, Luke and Yu, Lili},
  journal={Advances in Neural Information Processing Systems},
  volume={38},
  pages={9163--9182},
  year={2026}
}

@article{deng2025emerging,
  title={Emerging properties in unified multimodal pretraining},
  author={Deng, Chaorui and Zhu, Deyao and Li, Kunchang and Gou, Chenhui and Li, Feng and Wang, Zeyu and Zhong, Shu and Yu, Weihao and Nie, Xiaonan and Song, Ziang and others},
  journal={arXiv preprint arXiv:2505.14683},
  year={2025}
}

@inproceedings{lcoembed2025,
  title     = {Scaling Language-Centric Omnimodal Representation Learning},
  author    = {Xiao, Chenghao and Chan, Hou Pong and Zhang, Hao and Xu, Weiwen and
               Aljunied, Mahani and Rong, Yu},
  booktitle = {Advances in Neural Information Processing Systems},
  volume    = {38},
  pages     = {175826--175857},
  year      = {2025},
  doi       = {10.52202/085713-5291}
}

@inproceedings{alignvlm2025,
  title     = {{ALIGNVLM}: Bridging Vision and Language Latent Spaces for
               Multimodal Document Understanding},
  author    = {Masry, Ahmed and Rodriguez, Juan A. and Zhang, Tianyu and
               Wang, Suyuchen and Wang, Chao and Feizi, Aarash and
               Suresh, Akshay Kalkunte and Puri, Abhay and Jian, Xiangru and
               No{\"e}l, Pierre-Andr{\'e} and Madhusudhan, Sathwik Tejaswi and
               Pedersoli, Marco and Liu, Bang and Chapados, Nicolas and
               Bengio, Yoshua and Hoque, Enamul and Pal, Christopher and
               Laradji, Issam H. and Vazquez, David and Taslakian, Perouz and
               Gella, Spandana and Rajeswar, Sai},
  booktitle = {Advances in Neural Information Processing Systems},
  volume    = {38},
  pages     = {161722--161748},
  year      = {2025},
  doi       = {10.52202/085713-4875}
}

@inproceedings{textspace2026,
  title     = {The Dominance of Text Space: Unveiling the Asymmetric Nature of
               Cross-Modal Alignment in Large Language Models},
  author    = {Chen, Linqing and Zhong, Hanmeng and Wu, Wentao and Zhou, Peng},
  booktitle = {Proceedings of the 64th Annual Meeting of the Association for
               Computational Linguistics (Volume 1: Long Papers)},
  pages     = {36666--36678},
  year      = {2026},
  publisher = {Association for Computational Linguistics},
  address   = {San Diego, California, United States},
  doi       = {10.18653/v1/2026.acl-long.1699},
  url       = {https://aclanthology.org/2026.acl-long.1699/}
}

@inproceedings{radford2023robust,
  title={Robust speech recognition via large-scale weak supervision},
  author={Radford, Alec and Kim, Jong Wook and Xu, Tao and Brockman, Greg and McLeavey, Christine and Sutskever, Ilya},
  booktitle={International conference on machine learning},
  pages={28492--28518},
  year={2023},
  organization={PMLR}
}

@inproceedings{singh2019towards,
  title={Towards vqa models that can read},
  author={Singh, Amanpreet and Natarajan, Vivek and Shah, Meet and Jiang, Yu and Chen, Xinlei and Batra, Dhruv and Parikh, Devi and Rohrbach, Marcus},
  booktitle={2019 IEEE/CVF Conference on Computer Vision and Pattern Recognition (CVPR)},
  pages={8309--8318},
  year={2019},
  organization={IEEE}
}

@inproceedings{ardila2020common,
  title={Common voice: A massively-multilingual speech corpus},
  author={Ardila, Rosana and Branson, Megan and Davis, Kelly and Kohler, Michael and Meyer, Josh and Henretty, Michael and Morais, Reuben and Saunders, Lindsay and Tyers, Francis and Weber, Gregor},
  booktitle={Proceedings of the twelfth language resources and evaluation conference},
  pages={4218--4222},
  year={2020}
}

@inproceedings{panayotov2015librispeech,
  title={Librispeech: an asr corpus based on public domain audio books},
  author={Panayotov, Vassil and Chen, Guoguo and Povey, Daniel and Khudanpur, Sanjeev},
  booktitle={2015 IEEE international conference on acoustics, speech and signal processing (ICASSP)},
  pages={5206--5210},
  year={2015},
  organization={IEEE}
}

@article{wiedmann2025finevision,
  title={Finevision: Open data is all you need},
  author={Wiedmann, Luis and Zohar, Orr and Mahla, Amir and Wang, Xiaohan and Li, Rui and Frere, Thibaud and von Werra, Leandro and Gosthipaty, Aritra Roy and Marafioti, Andr{\'e}s},
  journal={arXiv preprint arXiv:2510.17269},
  year={2025}
}

@inproceedings{NEURIPS2025_d79a27cf,
 author = {Fu, Chaoyou and Chen, Peixian and Shen, Yunhang and Qin, Yulei and Zhang, Mengdan and Lin, Xu and Yang, Jinrui and Zheng, Xiawu and Li, Ke and Sun, Xing and Wu, Yunsheng and Ji, Rongrong and Shan, Caifeng and He, Ran},
 booktitle = {Advances in Neural Information Processing Systems},
 doi = {10.52202/085713-4899},
 editor = {D. Belgrave and C. Zhang and H. Lin and R. Pascanu and P. Koniusz and M. Ghassemi and N. Chen},
 pages = {},
 publisher = {Curran Associates, Inc.},
 title = {MME: A Comprehensive Evaluation Benchmark for Multimodal Large Language Models},
 url = {https://proceedings.neurips.cc/paper_files/paper/2025/file/d79a27cf2772fe00be7f341efc0eb517-Paper-Datasets_and_Benchmarks_Track.pdf},
 volume = {38, Main Conference},
 year = {2025}
}

@inproceedings{mathew2021docvqa,
  title={Docvqa: A dataset for vqa on document images},
  author={Mathew, Minesh and Karatzas, Dimosthenis and Jawahar, CV},
  booktitle={2021 IEEE Winter Conference on Applications of Computer Vision (WACV)},
  pages={2199--2208},
  year={2021},
  organization={IEEE}
}

@inproceedings{masry2022chartqa,
  title={Chartqa: A benchmark for question answering about charts with visual and logical reasoning},
  author={Masry, Ahmed and Tan, Jia Qing and Joty, Shafiq and Hoque, Enamul and others},
  booktitle={Findings of the association for computational linguistics: ACL 2022},
  pages={2263--2279},
  year={2022}
}

@inproceedings{liu2024mmbench,
  title={Mmbench: Is your multi-modal model an all-around player?},
  author={Liu, Yuan and Duan, Haodong and Zhang, Yuanhan and Li, Bo and Zhang, Songyang and Zhao, Wangbo and Yuan, Yike and Wang, Jiaqi and He, Conghui and Liu, Ziwei and others},
  booktitle={European conference on computer vision},
  pages={216--233},
  year={2024},
  organization={Springer}
}

@inproceedings{fu2025video,
  title={Video-mme: The first-ever comprehensive evaluation benchmark of multi-modal llms in video analysis},
  author={Fu, Chaoyou and Dai, Yuhan and Luo, Yongdong and Li, Lei and Ren, Shuhuai and Zhang, Renrui and Wang, Zihan and Zhou, Chenyu and Shen, Yunhang and Zhang, Mengdan and others},
  booktitle={2025 IEEE/CVF Conference on Computer Vision and Pattern Recognition (CVPR)},
  pages={24108--24118},
  year={2025},
  organization={IEEE}
}

@article{mangalam2023egoschema,
  title={Egoschema: A diagnostic benchmark for very long-form video language understanding},
  author={Mangalam, Karttikeya and Akshulakov, Raiymbek and Malik, Jitendra},
  journal={Advances in Neural Information Processing Systems},
  volume={36},
  pages={46212--46244},
  year={2023}
}

@misc{openai2025gpt51,
  title        = {GPT-5.1: A Smarter, More Conversational ChatGPT},
  author       = {{OpenAI}},
  year         = {2025},
  month        = nov,
  howpublished = {\url{https://openai.com/index/gpt-5-1/}}
}

@misc{anthropic2024claude35,
  title={Claude 3.5 Sonnet},
  author={{Anthropic}},
  year={2024},
  month={June},
  howpublished={Anthropic},
  url={https://www.anthropic.com/news/claude-3-5-sonnet}
}

@article{comanici2025gemini,
  title={Gemini 2.5: Pushing the frontier with advanced reasoning, multimodality, long context, and next generation agentic capabilities},
  author={Comanici, Gheorghe and Bieber, Eric and Schaekermann, Mike and Pasupat, Ice and Sachdeva, Noveen and Dhillon, Inderjit and Blistein, Marcel and Ram, Ori and Zhang, Dan and Rosen, Evan and others},
  journal={arXiv preprint arXiv:2507.06261},
  year={2025}
}

@article{hurst2024gpt,
  title={Gpt-4o system card},
  author={Hurst, Aaron and Lerer, Adam and Goucher, Adam P and Perelman, Adam and Ramesh, Aditya and Clark, Aidan and Ostrow, AJ and Welihinda, Akila and Hayes, Alan and Radford, Alec and others},
  journal={arXiv preprint arXiv:2410.21276},
  year={2024}
}

@article{xu2025qwen3,
  title={Qwen3-omni technical report},
  author={Xu, Jin and Guo, Zhifang and Hu, Hangrui and Chu, Yunfei and Wang, Xiong and He, Jinzheng and Wang, Yuxuan and Shi, Xian and He, Ting and Zhu, Xinfa and others},
  journal={arXiv preprint arXiv:2509.17765},
  year={2025}
}

@article{wang2024qwen2,
  title={Qwen2-vl: Enhancing vision-language model's perception of the world at any resolution},
  author={Wang, Peng and Bai, Shuai and Tan, Sinan and Wang, Shijie and Fan, Zhihao and Bai, Jinze and Chen, Keqin and Liu, Xuejing and Wang, Jialin and Ge, Wenbin and others},
  journal={arXiv preprint arXiv:2409.12191},
  year={2024}
}

@article{qwen2vl,
  title={Qwen2.5-VL Technical Report},
  author={Bai, Shuai and Chen, Keqin and Liu, Xuejing and Wang, Jialin and Ge, Wenbin and Song, Sibo and Dang, Kai and Wang, Peng and Wang, Shijie and Tang, Jun and Zhong, Humen and Zhu, Yuanzhi and Yang, Mingkun and Li, Zhaohai and Wan, Jianqiang and Wang, Pengfei and Ding, Wei and Fu, Zheren and Xu, Yiheng and Ye, Jiabo and Zhang, Xi and Xie, Tianbao and Cheng, Zesen and Zhang, Hang and Yang, Zhibo and Xu, Haiyang and Lin, Junyang},
  journal={arXiv preprint arXiv:2502.13923},
  year={2025}
}

@article{team2026qwen3,
  title={Qwen3.5-Omni Technical Report},
  author={Qwen},
  journal={arXiv preprint arXiv:2604.15804},
  year={2026}
}

@article{bai2025qwen3,
  title={Qwen3-vl technical report},
  author={Bai, Shuai and Cai, Yuxuan and Chen, Ruizhe and Chen, Keqin and Chen, Xionghui and Cheng, Zesen and Deng, Lianghao and Ding, Wei and Gao, Chang and Ge, Chunjiang and others},
  journal={arXiv preprint arXiv:2511.21631},
  year={2025}
}

@inproceedings{cha2024honeybee,
  title={Honeybee: Locality-enhanced projector for multimodal llm},
  author={Cha, Junbum and Kang, Wooyoung and Mun, Jonghwan and Roh, Byungseok},
  booktitle={2024 IEEE/CVF Conference on Computer Vision and Pattern Recognition (CVPR)},
  pages={13817--13827},
  year={2024},
  organization={IEEE}
}

@inproceedings{qu2025tokenflow,
  title={Tokenflow: Unified image tokenizer for multimodal understanding and generation},
  author={Qu, Liao and Zhang, Huichao and Liu, Yiheng and Wang, Xu and Jiang, Yi and Gao, Yiming and Ye, Hu and Du, Daniel K and Yuan, Zehuan and Wu, Xinglong},
  booktitle={2025 IEEE/CVF Conference on Computer Vision and Pattern Recognition (CVPR)},
  pages={2545--2555},
  year={2025},
  organization={IEEE}
}

@article{zhu2025internvl3,
  title={Internvl3: Exploring advanced training and test-time recipes for open-source multimodal models},
  author={Zhu, Jinguo and Wang, Weiyun and Chen, Zhe and Liu, Zhaoyang and Ye, Shenglong and Gu, Lixin and Tian, Hao and Duan, Yuchen and Su, Weijie and Shao, Jie and others},
  journal={arXiv preprint arXiv:2504.10479},
  year={2025}
}

@article{wang2025internvl3,
  title={Internvl3. 5: Advancing open-source multimodal models in versatility, reasoning, and efficiency},
  author={Wang, Weiyun and Gao, Zhangwei and Gu, Lixin and Pu, Hengjun and Cui, Long and Wei, Xingguang and Liu, Zhaoyang and Jing, Linglin and Ye, Shenglong and Shao, Jie and others},
  journal={arXiv preprint arXiv:2508.18265},
  year={2025}
}

@article{team2025kimi,
  title={Kimi-vl technical report},
  author={Team, Kimi and Du, Angang and Yin, Bohong and Xing, Bowei and Qu, Bowen and Wang, Bowen and Chen, Cheng and Zhang, Chenlin and Du, Chenzhuang and Wei, Chu and others},
  journal={arXiv preprint arXiv:2504.07491},
  year={2025}
}

@article{li2025tokenpacker,
  title={Tokenpacker: Efficient visual projector for multimodal llm},
  author={Li, Wentong and Yuan, Yuqian and Liu, Jian and Tang, Dongqi and Wang, Song and Qin, Jie and Zhu, Jianke and Zhang, Lei},
  journal={International Journal of Computer Vision},
  volume={133},
  number={10},
  pages={6794--6812},
  year={2025},
  publisher={Springer}
}

@article{chen2024expanding,
  title={Expanding performance boundaries of open-source multimodal models with model, data, and test-time scaling},
  author={Chen, Zhe and Wang, Weiyun and Cao, Yue and Liu, Yangzhou and Gao, Zhangwei and Cui, Erfei and Zhu, Jinguo and Ye, Shenglong and Tian, Hao and Liu, Zhaoyang and others},
  journal={arXiv preprint arXiv:2412.05271},
  year={2024}
}

@article{wang2025internvideo2,
  title={Internvideo2. 5: Empowering video mllms with long and rich context modeling},
  author={Wang, Yi and Li, Xinhao and Yan, Ziang and He, Yinan and Yu, Jiashuo and Zeng, Xiangyu and Wang, Chenting and Ma, Changlian and Huang, Haian and Gao, Jianfei and others},
  journal={arXiv preprint arXiv:2501.12386},
  year={2025}
}

@misc{opengvlab2024internomni,
  title={InternOmni: Extending InternVL with Audio Modality},
  author={{OpenGVLab}},
  year={2024},
  month={July},
  url={https://internvl.github.io/blog/2024-07-27-InternOmni/}
}

@misc{vicuna2023,
    title = {Vicuna: An Open-Source Chatbot Impressing GPT-4 with 90\%* ChatGPT Quality},
    url = {https://lmsys.org/blog/2023-03-30-vicuna/},
    author = {Chiang, Wei-Lin and Li, Zhuohan and Lin, Zi and Sheng, Ying and Wu, Zhanghao and Zhang, Hao and Zheng, Lianmin and Zhuang, Siyuan and Zhuang, Yonghao and Gonzalez, Joseph E. and Stoica, Ion and Xing, Eric P.},
    month = {March},
    year = {2023}
}

@article{hong2025glm,
  title={Glm-4.5 v and glm-4.1 v-thinking: Towards versatile multimodal reasoning with scalable reinforcement learning},
  author={Hong, Wenyi and Yu, Wenmeng and Gu, Xiaotao and Wang, Guo and Gan, Guobing and Tang, Haomiao and Cheng, Jiale and Qi, Ji and Ji, Junhui and Pan, Lihang and others},
  journal={arXiv preprint arXiv:2507.01006},
  year={2025}
}

@article{han2023imagebind,
  title={Imagebind-llm: Multi-modality instruction tuning},
  author={Han, Jiaming and Zhang, Renrui and Shao, Wenqi and Gao, Peng and Xu, Peng and Xiao, Han and Zhang, Kaipeng and Liu, Chris and Wen, Song and Guo, Ziyu and others},
  journal={arXiv preprint arXiv:2309.03905},
  year={2023}
}

@InProceedings{Liu_2024_CVPR,
    author    = {Liu, Haotian and Li, Chunyuan and Li, Yuheng and Lee, Yong Jae},
    title     = {Improved Baselines with Visual Instruction Tuning},
    booktitle = {Proceedings of the IEEE/CVF Conference on Computer Vision and Pattern Recognition (CVPR)},
    month     = {June},
    year      = {2024},
    pages     = {26296-26306}
}

@article{
li2025llavaonevision,
title={{LL}a{VA}-OneVision: Easy Visual Task Transfer},
author={Bo Li and Yuanhan Zhang and Dong Guo and Renrui Zhang and Feng Li and Hao Zhang and Kaichen Zhang and Peiyuan Zhang and Yanwei Li and Ziwei Liu and Chunyuan Li},
journal={Transactions on Machine Learning Research},
issn={2835-8856},
year={2025},
url={https://openreview.net/forum?id=zKv8qULV6n},
note={}
}

@misc{liu2024llavanext,
    title={LLaVA-NeXT: Improved reasoning, OCR, and world knowledge},
    url={https://llava-vl.github.io/blog/2024-01-30-llava-next/},
    author={Liu, Haotian and Li, Chunyuan and Li, Yuheng and Li, Bo and Zhang, Yuanhan and Shen, Sheng and Lee, Yong Jae},
    month={January},
    year={2024}
}

@article{
zhang2025llavavideo,
title={{LL}a{VA}-Video: Video Instruction Tuning With Synthetic Data},
author={Yuanhan Zhang and Jinming Wu and Wei Li and Bo Li and Zejun MA and Ziwei Liu and Chunyuan Li},
journal={Transactions on Machine Learning Research},
issn={2835-8856},
year={2025},
url={https://openreview.net/forum?id=EElFGvt39K},
note={}
}

@article{he2024efficient,
  title={Efficient multimodal learning from data-centric perspective},
  author={He, Muyang and Liu, Yexin and Wu, Boya and Yuan, Jianhao and Wang, Yueze and Huang, Tiejun and Zhao, Bo},
  journal={arXiv preprint arXiv:2402.11530},
  year={2024}
}

@inproceedings{liang2026enhancing,
  title={Enhancing spatial reasoning through visual and textual thinking},
  author={Liang, Xun and Guo, Xin and Jin, Zhongming and Pan, Weihang and Shang, Penghui and Cai, Deng and Lin, Binbin and Ye, Jieping},
  booktitle={Proceedings of the AAAI Conference on Artificial Intelligence},
  volume={40},
  number={28},
  pages={23433--23441},
  year={2026}
}

@inproceedings{yu2025auto,
  title={Auto-controlled image perception in mllms via visual perception tokens},
  author={Yu, Runpeng and Ma, Xinyin and Wang, Xinchao},
  booktitle={2025 IEEE/CVF International Conference on Computer Vision (ICCV)},
  pages={21822--21831},
  year={2025},
  organization={IEEE}
}

@article{zhang2025spar,
  title={From flatland to space: Teaching vision-language models to perceive and reason in 3d},
  author={Zhang, Jiahui and Chen, Yurui and Xu, Yueming and Huang, Ze and Mei, Jilin and Chen, Chunhui and Zhou, Yanpeng and Yuan, Yu-Jie and Cai, Xinyue and Huang, Guowei and others},
  journal={Advances in Neural Information Processing Systems},
  volume={38},
  year={2026}
}

@article{krishna2017visual,
  title={Visual genome: Connecting language and vision using crowdsourced dense image annotations},
  author={Krishna, Ranjay and Zhu, Yuke and Groth, Oliver and Johnson, Justin and Hata, Kenji and Kravitz, Joshua and Chen, Stephanie and Kalantidis, Yannis and Li, Li-Jia and Shamma, David A and others},
  journal={International journal of computer vision},
  volume={123},
  number={1},
  pages={32--73},
  year={2017},
  publisher={Springer}
}

@inproceedings{kirillov2023segment,
  title={Segment anything},
  author={Kirillov, Alexander and Mintun, Eric and Ravi, Nikhila and Mao, Hanzi and Rolland, Chloe and Gustafson, Laura and Xiao, Tete and Whitehead, Spencer and Berg, Alexander C and Lo, Wan-Yen and others},
  booktitle={2023 IEEE/CVF international conference on computer vision (ICCV)},
  pages={3992--4003},
  year={2023},
  organization={IEEE}
}

@inproceedings{mishra2019ocr,
  title={Ocr-vqa: Visual question answering by reading text in images},
  author={Mishra, Anand and Shekhar, Shashank and Singh, Ajeet Kumar and Chakraborty, Anirban},
  booktitle={2019 international conference on document analysis and recognition (ICDAR)},
  pages={947--952},
  year={2019},
  organization={IEEE}
}

@inproceedings{daxberger2025mm,
  title={Mm-spatial: Exploring 3d spatial understanding in multimodal llms},
  author={Daxberger, Erik and Wenzel, Nina and Griffiths, David and Gang, Haiming and Lazarow, Justin and Kohavi, Gefen and Kang, Kai and Eichner, Marcin and Yang, Yinfei and Dehghan, Afshin and others},
  booktitle={2025 IEEE/CVF International Conference on Computer Vision (ICCV)},
  pages={7395--7408},
  year={2025},
  organization={IEEE}
}

@inproceedings{cai2024spatialbot,
  title={Spatialbot: Precise spatial understanding with vision language models},
  author={Cai, Wenxiao and Ponomarenko, Iaroslav and Yuan, Jianhao and Li, Xiaoqi and Yang, Wankou and Dong, Hao and Zhao, Bo},
  booktitle={2025 IEEE International Conference on Robotics and Automation (ICRA)},
  pages={9490--9498},
  year={2025},
  organization={IEEE}
}

@inproceedings{song2025robospatial,
  title={Robospatial: Teaching spatial understanding to 2d and 3d vision-language models for robotics},
  author={Song, Chan Hee and Blukis, Valts and Tremblay, Jonathan and Tyree, Stephen and Su, Yu and Birchfield, Stan},
  booktitle={Proceedings of the Computer Vision and Pattern Recognition Conference},
  pages={15768--15780},
  year={2025}
}

@article{du2024cosyvoice,
  title={Cosyvoice 2: Scalable streaming speech synthesis with large language models},
  author={Du, Zhihao and Wang, Yuxuan and Chen, Qian and Shi, Xian and Lv, Xiang and Zhao, Tianyu and Gao, Zhifu and Yang, Yexin and Gao, Changfeng and Wang, Hui and others},
  journal={arXiv preprint arXiv:2412.10117},
  year={2024}
}

@article{yang2024depth,
  title={Depth anything v2},
  author={Yang, Lihe and Kang, Bingyi and Huang, Zilong and Zhao, Zhen and Xu, Xiaogang and Feng, Jiashi and Zhao, Hengshuang},
  journal={Advances in neural information processing systems},
  volume={37},
  pages={21875--21911},
  year={2024}
}

@article{liu2026ssr,
  title={Ssr: Enhancing depth perception in vision-language models via rationale-guided spatial reasoning},
  author={Liu, Yang and Ma, Ming and Yu, Xiaomin and Ding, Pengxiang and Zhao, Han and Sun, Mingyang and Huang, Siteng and Wang, Donglin},
  journal={Advances in Neural Information Processing Systems},
  volume={38},
  pages={123926--123958},
  year={2026}
}

@inproceedings{chen2024spatialvlm,
  title={Spatialvlm: Endowing vision-language models with spatial reasoning capabilities},
  author={Chen, Boyuan and Xu, Zhuo and Kirmani, Sean and Ichter, Brain and Sadigh, Dorsa and Guibas, Leonidas and Xia, Fei},
  booktitle={Proceedings of the IEEE/CVF conference on computer vision and pattern recognition},
  pages={14455--14465},
  year={2024}
}

@article{batra2025spatialthinker,
  title={Spatialthinker: Reinforcing 3d reasoning in multimodal llms via spatial rewards},
  author={Batra, Hunar and Tu, Haoqin and Chen, Hardy and Lin, Yuanze and Xie, Cihang and Clark, Ronald},
  journal={arXiv preprint arXiv:2511.07403},
  year={2025}
}

@misc{remyx2025spaceom,
  title        = {SpaceOm},
  author       = {{Remyx AI}},
  year         = {2025},
  howpublished = {\url{https://huggingface.co/remyxai/SpaceOm}},
  note         = {Hugging Face model card}
}

@article{qorsham2025end,
  title={End-to-End 3D point cloud question answering via state space models},
  author={Qorsham, Hatem and El-Deeb, Reem and Essa, Ehab},
  journal={Applied Soft Computing},
  volume={181},
  pages={113404},
  year={2025},
  publisher={Elsevier}
}

@inproceedings{khan2025pointqa,
  title={PointQA: Multi-Modality Guided Cross-Attention for 3D Visual Question Answering on Point Clouds},
  author={Khan, Muhammad Zeeshan and Gaddam, Anuroop},
  booktitle={2025 International Conference on Digital Image Computing: Techniques and Applications (DICTA)},
  pages={1--7},
  year={2025},
  organization={IEEE}
}

@inproceedings{achlioptas2020referit_3d,
    title={{ReferIt3D}: Neural Listeners for Fine-Grained 3D Object Identification in Real-World Scenes},
    author={Achlioptas, Panos and Abdelreheem, Ahmed and Xia, Fei and Elhoseiny, Mohamed and Guibas, Leonidas J.},
    booktitle={16th European Conference on Computer Vision (ECCV)},
    year={2020}
}

@InProceedings{Fu_2025_WACV,
    author    = {Fu, Rao and Liu, Jingyu and Chen, Xilun and Nie, Yixin and Xiong, Wenhan},
    title     = {Scene-LLM: Extending Language Model for 3D Visual Reasoning},
    booktitle = {Proceedings of the Winter Conference on Applications of Computer Vision (WACV)},
    month     = {February},
    year      = {2025},
    pages     = {2195-2206}
}

@inproceedings{ICLR2026_fef12656,
 author = {Qi, Zhangyang and Zhang, Zhixiong and Fang, Ye and Wang, Jiaqi and Zhao, Hengshuang},
 booktitle = {International Conference on Learning Representations},
 editor = {C. Vondrick and B. Hariharan and C. Raffel and L. Pinto and D. Yang and A. Faust},
 pages = {157277--157302},
 title = {GPT4Scene: Understand 3D Scenes from Videos with Vision-Language Models},
 url = {https://proceedings.iclr.cc/paper_files/paper/2026/file/fef126561bbf9d4467dbb8d27334b8fe-Paper-Conference.pdf},
 volume = {2026},
 year = {2026}
}

@InProceedings{Chen_2024_CVPR,
    author    = {Chen, Sijin and Chen, Xin and Zhang, Chi and Li, Mingsheng and Yu, Gang and Fei, Hao and Zhu, Hongyuan and Fan, Jiayuan and Chen, Tao},
    title     = {LL3DA: Visual Interactive Instruction Tuning for Omni-3D Understanding Reasoning and Planning},
    booktitle = {Proceedings of the IEEE/CVF Conference on Computer Vision and Pattern Recognition (CVPR)},
    month     = {June},
    year      = {2024},
    pages     = {26428-26438}
}

@inproceedings{NEURIPS2024_cebbd24f,
 author = {Huang, Haifeng and Chen, Yilun and Wang, Zehan and Huang, Rongjie and Xu, Runsen and Wang, Tai and Liu, Luping and Cheng, Xize and Zhao, Yang and Pang, Jiangmiao and Zhao, Zhou},
 booktitle = {Advances in Neural Information Processing Systems},
 doi = {10.52202/079017-3620},
 editor = {A. Globerson and L. Mackey and D. Belgrave and A. Fan and U. Paquet and J. Tomczak and C. Zhang},
 pages = {113991--114017},
 publisher = {Curran Associates, Inc.},
 title = {Chat-Scene: Bridging 3D Scene and Large Language Models with Object Identifiers},
 url = {https://proceedings.neurips.cc/paper_files/paper/2024/file/cebbd24f1e50bcb63d015611fe0fe767-Paper-Conference.pdf},
 volume = {37},
 year = {2024}
}

@article{chiang2026viewmind3d,
  title={ViewMind3D: Modular View-Aware Inference for Training-Free 3D-QA},
  author={Chiang, Ping-Kun and Wu, Kun-Ru and Li, Po-han and Chinchali, Sandeep and Topcu, Ufuk and Tseng, Yu-Chee},
  journal={arXiv preprint arXiv:2607.28442},
  year={2026}
}

@inproceedings{jia2024sceneverse,
  title={SceneVerse: Scaling 3D Vision-Language Learning for Grounded Scene Understanding},
  author={Jia, Baoxiong and Chen, Yixin and Yu, Huangyue and Wang, Yan and Niu, Xuesong and Liu, Tengyu and Li, Qing and Huang, Siyuan},
  booktitle={European Conference on Computer Vision},
  pages={289--310},
  year={2024}
}

@inproceedings{mo2025advancing,
  title={Advancing 3D scene understanding with MV-ScanQA multi-view reasoning evaluation and TripAlign pre-training dataset},
  author={Mo, Wentao and Chen, Qingchao and Peng, Yuxin and Huang, Siyuan and Liu, Yang},
  booktitle={Proceedings of the 33rd ACM International Conference on Multimedia},
  pages={12973--12980},
  year={2025}
}

@inproceedings{azuma2022scanqa,
  title={Scanqa: 3d question answering for spatial scene understanding},
  author={Azuma, Daichi and Miyanishi, Taiki and Kurita, Shuhei and Kawanabe, Motoaki},
  booktitle={proceedings of the IEEE/CVF conference on computer vision and pattern recognition},
  pages={19129--19139},
  year={2022}
}

@article{yan2023clevr3d,
  title={Comprehensive visual question answering on point clouds through compositional scene manipulation},
  author={Yan, Xu and Yuan, Zhihao and Du, Yuhao and Liao, Yinghong and Guo, Yao and Cui, Shuguang and Li, Zhen},
  journal={IEEE Transactions on Visualization and Computer Graphics},
  volume={30},
  number={12},
  pages={7473--7485},
  year={2023},
  publisher={IEEE}
}

@inproceedings{cao2025irgpt,
  title={Irgpt: Understanding real-world infrared image with bi-cross-modal curriculum on large-scale benchmark},
  author={Cao, Zhe and Zhang, Jin and Zhang, Ruiheng},
  booktitle={Proceedings of the IEEE/CVF International Conference on Computer Vision},
  pages={166--176},
  year={2025}
}

@inproceedings{jiang2024infrared,
  title={Infrared-llava: Enhancing understanding of infrared images in multi-modal large language models},
  author={Jiang, Shixin and Chen, Zerui and Liang, Jiafeng and Zhao, Yanyan and Liu, Ming and Qin, Bing},
  booktitle={Findings of the Association for Computational Linguistics: EMNLP 2024},
  pages={8573--8591},
  year={2024}
}

@inproceedings{zhang2026ifbench,
  title={IF-Bench: Benchmarking and enhancing mllms for infrared images with generative visual prompting},
  author={Zhang, Tao and Hong, Yuyang and Xia, Yang and Ding, Kun and Zhang, Zeyu and Wang, Ying and Xiang, Shiming and Pan, Chunhong},
  booktitle={Proceedings of the IEEE/CVF Conference on Computer Vision and Pattern Recognition},
  pages={8205--8215},
  year={2026}
}

@article{chung2026enhanced,
  title={Enhanced vision-language models for diverse sensor understanding: Cost-efficient optimization and benchmarking},
  author={Chung, Sangyun and Yu, Youngjoon and Kim, Seyeon and Chee, Youngchae and Ro, Yong Man},
  journal={IEEE Transactions on Image Processing},
  year={2026},
  publisher={IEEE}
}

@article{yu2024octopi,
  title={Octopi: Object property reasoning with large tactile-language models},
  author={Yu, Samson and Lin, Kelvin and Xiao, Anxing and Duan, Jiafei and Soh, Harold},
  journal={arXiv preprint arXiv:2405.02794},
  year={2024}
}

@inproceedings{
xie2025universal,
title={Universal Visuo-Tactile Video Understanding for Embodied Interaction},
author={Yifan Xie and Mingyang Li and Shoujie Li and Xingting Li and Guangyu Chen and Fei Ma and Fei Yu and Wenbo Ding},
booktitle={The Thirty-ninth Annual Conference on Neural Information Processing Systems},
year={2025},
url={https://openreview.net/forum?id=aPPnmuuNhx}
}

@article{yuan2017gelsight,
  title={Gelsight: High-resolution robot tactile sensors for estimating geometry and force},
  author={Yuan, Wenzhen and Dong, Siyuan and Adelson, Edward H},
  journal={Sensors},
  volume={17},
  number={12},
  pages={2762},
  year={2017},
  publisher={MDPI}
}

@article{tu2026unitac,
  title={UniTac: A Unified Multimodal Model for Cross-Sensor Tactile Understanding and Generation},
  author={Tu, Jiahang and Yang, Fengyu and Ma, Chenyang and Yu, Xihang and Zeng, Ziyao and Wu, Shaokai and Zhao, Hanbin and Tao, Zhi and Zhang, Chao and Qian, Hui and others},
  journal={arXiv preprint arXiv:2606.31451},
  year={2026}
}

@inproceedings{hong2026genhar,
  title={GenHAR: Generalizing Cross-domain Human Activity Recognition for Last-mile Delivery},
  author={Hong, Zhiqing and Li, Zelong and Fan, Xiubin and Yang, Guang and Guo, Baoshen and Wang, Haotian and He, Tian and Zhang, Desheng},
  booktitle={Proceedings of the 32nd ACM SIGKDD Conference on Knowledge Discovery and Data Mining V. 1},
  pages={2208--2219},
  year={2026}
}

@article{yan2025large,
  title={Large language model-guided semantic alignment for human activity recognition},
  author={Yan, Hua and Tan, Heng and Ding, Yi and Zhou, Pengfei and Namboodiri, Vinod and Yang, Yu},
  journal={Proceedings of the ACM on Interactive, Mobile, Wearable and Ubiquitous Technologies},
  volume={9},
  number={4},
  pages={1--25},
  year={2025},
  publisher={ACM New York, NY, USA}
}

@inproceedings{imran2024llasa,
  title={LLaSA: A Sensor-Aware LLM for Natural Language Reasoning of Human Activity from IMU Data},
  author={Asif Imran Shouborno, Sheikh and Khan, Mohammad Nur Hossain and Biswas, Subrata and Islam, Bashima},
  booktitle={Companion of the 2025 ACM International Joint Conference on Pervasive and Ubiquitous Computing},
  pages={893--899},
  year={2025}
}

@inproceedings{stisen2015hhar,
  title={Smart devices are different: Assessing and mitigatingmobile sensing heterogeneities for activity recognition},
  author={Stisen, Allan and Blunck, Henrik and Bhattacharya, Sourav and Prentow, Thor Siiger and Kj{\ae}rgaard, Mikkel Baun and Dey, Anind and Sonne, Tobias and Jensen, Mads M{\o}ller},
  booktitle={Proceedings of the 13th ACM conference on embedded networked sensor systems},
  pages={127--140},
  year={2015}
}

@inproceedings{anguita2013uci,
  title={A public domain dataset for human activity recognition using smartphones.},
  author={Anguita, Davide and Ghio, Alessandro and Oneto, Luca and Parra, Xavier and Reyes-Ortiz, Jorge Luis and others},
  booktitle={Esann},
  volume={3},
  number={1},
  pages={3--4},
  year={2013}
}

@inproceedings{malekzadeh2019motionsense,
  title={Mobile sensor data anonymization},
  author={Malekzadeh, Mohammad and Clegg, Richard G and Cavallaro, Andrea and Haddadi, Hamed},
  booktitle={Proceedings of the international conference on internet of things design and implementation},
  pages={49--58},
  year={2019}
}

@article{shoaib2014fusion,
  title={Fusion of smartphone motion sensors for physical activity recognition},
  author={Shoaib, Muhammad and Bosch, Stephan and Incel, Ozlem Durmaz and Scholten, Hans and Havinga, Paul JM},
  journal={Sensors},
  volume={14},
  number={6},
  pages={10146--10176},
  year={2014},
  publisher={Multidisciplinary Digital Publishing Institute}
}

@article{yang2023mmfi,
  title={Mm-fi: Multi-modal non-intrusive 4d human dataset for versatile wireless sensing},
  author={Yang, Jianfei and Huang, He and Zhou, Yunjiao and Chen, Xinyan and Xu, Yuecong and Yuan, Shenghai and Zou, Han and Lu, Chris Xiaoxuan and Xie, Lihua},
  journal={Advances in Neural Information Processing Systems},
  volume={36},
  pages={18756--18768},
  year={2023}
}

@article{xie2024chatts,
  title={Chatts: Aligning time series with llms via synthetic data for enhanced understanding and reasoning},
  author={Xie, Zhe and Li, Zeyan and He, Xiao and Xu, Longlong and Wen, Xidao and Zhang, Tieying and Chen, Jianjun and Shi, Rui and Pei, Dan},
  journal={arXiv preprint arXiv:2412.03104},
  year={2024}
}

@article{divo2025quants,
  title={Quants: Question answering on time series},
  author={Divo, Felix and Kraus, Maurice and Nguyen, Anh Q and Xue, Hao and Razzak, Imran and Salim, Flora D and Kersting, Kristian and Dhami, Devendra Singh},
  journal={arXiv preprint arXiv:2511.05124},
  year={2025}
}

@incollection{loper2023smpl,
  title={SMPL: A skinned multi-person linear model},
  author={Loper, Matthew and Mahmood, Naureen and Romero, Javier and Pons-Moll, Gerard and Black, Michael J},
  booktitle={Seminal Graphics Papers: Pushing the Boundaries, Volume 2},
  pages={851--866},
  year={2023}
}

@inproceedings{feng2026smellnet,
  title={SmellNet: A Dataset for Sensor-Based Smell Recognition and Mixture Prediction},
  author={Feng, Dewei and Dai, Wei and Li, Carol and Pernigo, Alistair and Liang, Paul},
  booktitle={International Conference on Learning Representations},
  volume={2026},
  pages={103061--103092},
  year={2026}
}

@inproceedings{hu2024wavllm,
  title={Wavllm: Towards robust and adaptive speech large language model},
  author={Hu, Shujie and Zhou, Long and Liu, Shujie and Chen, Sanyuan and Meng, Lingwei and Hao, Hongkun and Pan, Jing and Liu, Xunying and Li, Jinyu and Sivasankaran, Sunit and others},
  booktitle={Findings of the Association for Computational Linguistics: EMNLP 2024},
  pages={4552--4572},
  year={2024}
}

@article{thushara2026thermo,
  title={Thermo-VL: Extending Vision-Language Models to Thermal Infrared Perception},
  author={Thushara, Rusiru and Ranasinghe, Yasiru and Paranjape, Jay and Patel, Vishal M},
  journal={arXiv preprint arXiv:2605.21882},
  year={2026}
}

@article{ghosh2026audio,
  title={Audio flamingo 3: Advancing audio intelligence with fully open large audio language models},
  author={Ghosh, Sreyan and Goel, Arushi and Kim, Jaehyeon and Kumar, Sonal and Kong, Zhifeng and Lee, Sang-gil and Yang, Chao-Han and Duraiswami, Ramani and Manocha, Dinesh and Valle, Rafael and others},
  journal={Advances in Neural Information Processing Systems},
  volume={38},
  pages={41819--41886},
  year={2026}
}

@inproceedings{zhou2025if,
  title={What If LLMs Can Smell: A Prototype.},
  author={Zhou, Xueyi and Lu, Qi and Chae, Dong-Kyu},
  booktitle={IJCAI},
  pages={11141--11144},
  year={2025}
}

@inproceedings{zhou2025does,
  title={" Where Does This Strange Smell Come from?": Enabling Conversational Interfaces for Artificial Olfaction.},
  author={Zhou, Xueyi and Lu, Qi and Chae, Dong-Kyu},
  booktitle={EMNLP (Findings)},
  pages={10719--10745},
  year={2025}
}

@article{jiang2023motiongpt,
  title={Motiongpt: Human motion as a foreign language},
  author={Jiang, Biao and Chen, Xin and Liu, Wen and Yu, Jingyi and Yu, Gang and Chen, Tao},
  journal={Advances in Neural Information Processing Systems},
  volume={36},
  pages={20067--20079},
  year={2023}
}

@article{chen2025motionllm,
  title={Motionllm: Understanding human behaviors from human motions and videos},
  author={Chen, Ling-Hao and Lu, Shunlin and Zeng, Ailing and Zhang, Hao and Wang, Benyou and Zhang, Ruimao and Zhang, Lei},
  journal={IEEE Transactions on Pattern Analysis and Machine Intelligence},
  year={2025},
  publisher={IEEE}
}

@article{wang2026universal,
  title={Universal Skeleton Understanding via Differentiable Rendering and MLLMs},
  author={Wang, Ziyi and Li, Peiming and Wang, Xinshun and Tang, Yang and Ma, Kai-Kuang and Liu, Mengyuan},
  journal={arXiv preprint arXiv:2603.18003},
  year={2026}
}

@article{team2024octo,
  title={Octo: An open-source generalist robot policy},
  author={Ghosh, Dibya and Walke, Homer and Pertsch, Karl and Black, Kevin and Mees, Oier and Dasari, Sudeep and Hejna, Joey and Kreiman, Tobias and Xu, Charles and others},
  journal={arXiv preprint arXiv:2405.12213},
  year={2024}
}

@article{cheng2024spatialrgpt,
  title={Spatialrgpt: Grounded spatial reasoning in vision-language models},
  author={Cheng, An-Chieh and Yin, Hongxu and Fu, Yang and Guo, Qiushan and Yang, Ruihan and Kautz, Jan and Wang, Xiaolong and Liu, Sifei},
  journal={Advances in Neural Information Processing Systems},
  volume={37},
  pages={135062--135093},
  year={2024}
}

@article{hong20233d,
  title={3d-llm: Injecting the 3d world into large language models},
  author={Hong, Yining and Zhen, Haoyu and Chen, Peihao and Zheng, Shuhong and Du, Yilun and Chen, Zhenfang and Gan, Chuang},
  journal={Advances in Neural Information Processing Systems},
  volume={36},
  pages={20482--20494},
  year={2023}
}

@article{guo2023point,
  title={Point-bind \& point-llm: Aligning point cloud with multi-modality for 3d understanding, generation, and instruction following},
  author={Guo, Ziyu and Zhang, Renrui and Zhu, Xiangyang and Tang, Yiwen and Ma, Xianzheng and Han, Jiaming and Chen, Kexin and Gao, Peng and Li, Xianzhi and Li, Hongsheng and others},
  journal={arXiv preprint arXiv:2309.00615},
  year={2023}
}

@misc{zhou2024transfusionpredicttokendiffuse,
      title={Transfusion: Predict the Next Token and Diffuse Images with One Multi-Modal Model}, 
      author={Chunting Zhou and Lili Yu and Arun Babu and Kushal Tirumala and Michihiro Yasunaga and Leonid Shamis and Jacob Kahn and Xuezhe Ma and Luke Zettlemoyer and Omer Levy},
      journal={arXiv preprint arXiv:2408.11039},
      year={2024},
}

@article{wang2024emu3,
  title={Emu3: Next-token prediction is all you need},
  author={Wang, Xinlong and Zhang, Xiaosong and Luo, Zhengxiong and Sun, Quan and Cui, Yufeng and Wang, Jinsheng and Zhang, Fan and Wang, Yueze and Li, Zhen and Yu, Qiying and others},
  journal={arXiv preprint arXiv:2409.18869},
  year={2024}
}

@article{wang2026motiongpt,
  title={Motiongpt-2: A general-purpose motion-language model for motion generation and understanding},
  author={Wang, Yuan and Huang, Di and Zhang, Yaqi and Ouyang, Wanli and Jiao, Jile and Feng, Xuetao and Xu, Dan and Tang, Shixiang},
  journal={IEEE Transactions on Circuits and Systems for Video Technology},
  year={2026},
  publisher={IEEE}
}

@misc{yuan2026embodiedr15,
      title={Embodied-R1.5: Evolving Physical Intelligence via Embodied Foundation Models}, 
      author={Yifu Yuan and Yaoting Huang and Xianze Yao and Yutong Li and Shuoheng Zhang and Linqi Han and Pengyi Li and Jiangeng Sun and Wenting Jia and Zhao Zhang and Yuhao Liu and Ruihao Liao and Yucheng Hu and Qiyu Wu and Yuxiao Li and Zibin Dong and Fei Ni and Yan Zheng and Shuyang Gu and Yi Ma and Hongyao Tang and Han Hu and Jianye Hao},
      year={2026},
      eprint={2606.11324},
      archivePrefix={arXiv},
}

@misc{geminirobotics,
      title={Gemini Robotics: Bringing AI into the Physical World}, 
      author={Gemini Robotics Team and Saminda Abeyruwan and Joshua Ainslie and Jean-Baptiste Alayrac and Montserrat Gonzalez Arenas and Travis Armstrong and Ashwin Balakrishna and Robert Baruch and Maria Bauza and Michiel Blokzijl and Steven Bohez and Konstantinos Bousmalis and Anthony Brohan and Thomas Buschmann and Arunkumar Byravan and Serkan Cabi and Ken Caluwaerts and Federico Casarini and Oscar Chang and Jose Enrique Chen and Xi Chen and Hao-Tien Lewis Chiang and Krzysztof Choromanski and David D'Ambrosio and Sudeep Dasari and Todor Davchev and Coline Devin and Norman Di Palo and Tianli Ding and Adil Dostmohamed and Danny Driess and Yilun Du and Debidatta Dwibedi and Michael Elabd and Claudio Fantacci and Cody Fong and Erik Frey and Chuyuan Fu and Marissa Giustina and Keerthana Gopalakrishnan and Laura Graesser and Leonard Hasenclever and Nicolas Heess and Brandon Hernaez and Alexander Herzog and R. Alex Hofer and Jan Humplik and Atil Iscen and Mithun George Jacob and Deepali Jain and Ryan Julian and Dmitry Kalashnikov and M. Emre Karagozler and Stefani Karp and Chase Kew and Jerad Kirkland and Sean Kirmani and Yuheng Kuang and Thomas Lampe and Antoine Laurens and Isabel Leal and Alex X. Lee and Tsang-Wei Edward Lee and Jacky Liang and Yixin Lin and Sharath Maddineni and Anirudha Majumdar and Assaf Hurwitz Michaely and Robert Moreno and Michael Neunert and Francesco Nori and Carolina Parada and Emilio Parisotto and Peter Pastor and Acorn Pooley and Kanishka Rao and Krista Reymann and Dorsa Sadigh and Stefano Saliceti and Pannag Sanketi and Pierre Sermanet and Dhruv Shah and Mohit Sharma and Kathryn Shea and Charles Shu and Vikas Sindhwani and Sumeet Singh and Radu Soricut and Jost Tobias Springenberg and Rachel Sterneck and Razvan Surdulescu and Jie Tan and Jonathan Tompson and Vincent Vanhoucke and Jake Varley and Grace Vesom and Giulia Vezzani and Oriol Vinyals and Ayzaan Wahid and Stefan Welker and Paul Wohlhart and Fei Xia and Ted Xiao and Annie Xie and Jinyu Xie and Peng Xu and Sichun Xu and Ying Xu and Zhuo Xu and Yuxiang Yang and Rui Yao and Sergey Yaroshenko and Wenhao Yu and Wentao Yuan and Jingwei Zhang and Tingnan Zhang and Allan Zhou and Yuxiang Zhou},
      year={2025},
      eprint={2503.20020},
      archivePrefix={arXiv},
}
\bibliographystyle{iclr2027_conference}

\newpage
\appendix

\renewcommand\thefigure{\Alph{section}\arabic{figure}}
\renewcommand\thetable{\Alph{section}\arabic{table}} 

\section*{Appendix Contents}

\startcontents[appendices] 
\setcounter{tocdepth}{2} 
\printcontents[appendices]{}{1}{} 

\newtcolorbox{ratingbox}[1]{
    colback=white,          
    colframe=black!70,      
    boxrule=0.5mm,          
    arc=1.5mm,               
    top=2mm, bottom=2mm, left=3mm, right=3mm,
    title={#1},             
    coltitle=black,         
    fonttitle=\bfseries,    
    colbacktitle=black!20,  
    breakable,              
}

\newpage
\section{Related Work}
\label{app:related_work}
\noindent \textbf{Generalist Embodied Models.}
Recent advances in MLLMs have motivated a growing line of generalist embodied models that leverage pretrained multimodal backbones for unified perception and reasoning in physical environments.
Early systems such as PaLM-E~\citep{driess2023palm} demonstrated the potential of integrating embodied observations into large language models, while subsequent models such as RoboBrain~\citep{robobrain,robobrain2}, RynnBrain~\citep{dang2026rynnbrain}, and Vesta~\citep{bjorck2026vesta} further broaden this paradigm across diverse physical-world tasks and environments.
More recent efforts further emphasize spatial-temporal perception and high-level reasoning~\citep{zhao2025embodied,mo2025advancing,geminirobotics,zhang2026hy,yuan2026embodiedr15}, while benchmarks such as OpenEQA~\citep{majumdar2024openeqa} and FoMER~\citep{dissanayake2025good} reflect the growing focus on embodied reasoning.
In parallel, vision-language-action models map visual observations directly to low-level robot control~\citep{team2024octo,black2024pi_0,kim2025openvla}, targeting action generation rather than the high-level sensory understanding considered in this work.
Despite this progress, these generalist models still rely largely on basic modalities such as vision and audio, and their extensions beyond these modalities are typically restricted to a single additional modality.
Heterogeneous physical signals such as depth, contact, temperature, inertial dynamics, and fine-grained geometry are therefore rarely available to a single embodied reasoner, even though dedicated models exist for many of them individually.

\noindent \textbf{Heterogeneous Multimodal LLMs.}
A parallel line of work extends LLMs to increasingly diverse input modalities. 
Image-text pretraining first provides a language-anchored semantic space through large-scale cross-modal alignment~\citep{yu2022coca,tschannen2025siglip}, which subsequent binding-based methods~\citep{girdhar2023imagebind,zhu2024languagebind} reuse as a pivot to align heterogeneous modalities.
Unified multimodal frameworks~\citep{moon2024anymal,zhang2025omnibind} instead attach multiple modality encoders to a common LLM, enabling a broader set of input types and tasks within one framework.
Audio and speech models likewise benefit from large-scale paired supervision, providing mature language-aligned interfaces for acoustic inputs and spoken instructions~\citep{chu2024qwen2audio,hu2024wavllm,ghosh2026audio}.
Other works develop dedicated interfaces for individual physical signals, including depth~\citep{cai2024spatialbot,cheng2024spatialrgpt,daxberger2025mm}, point clouds~\citep{hong20233d,guo2023point,xu2024pointllm}, inertial and time-series data~\citep{imran2024llasa,li2025sensorllm}, touch~\citep{fu2024touch}, thermal~\citep{jiang2024infrared,cao2025irgpt,thushara2026thermo}, radar~\citep{lai2026radarllm,zhou2026holollm}, smell~\citep{zhou2025if,zhou2025does}, and motion~\citep{jiang2023motiongpt,chen2025motionllm,wang2026motiongpt,wang2026universal}.
However, sensory data generally receive far less linguistic supervision than web-scale image-text data, making their physical content difficult to expose fully to language-based reasoning, and each of these interfaces is designed for one modality in isolation.
In contrast, \method uses caption-guided supervision to make sensory semantics language-accessible and covers 8 embodied sensory modalities within a single model.

\noindent \textbf{Multimodal Fusion and Specialization.}
Beyond modality alignment, multimodal models also differ in how heterogeneous representations are integrated into language reasoning.  
Encoder-projector architectures~\citep{liu2023visual,li2023blip,dai2023instructblip} typically perform \emph{late fusion}, offering a simple interface but limiting deeper and fine-grained cross-modal interaction.
A related line of work unifies modalities on the encoder side, reusing a \emph{shared encoder} or a pre-aligned binding space as a single interface to the language model~\citep{han2024onellm,wu2024next}, which makes modality extension straightforward but leaves modality-specific structure largely unmodeled.
Native \emph{early-fusion} models~\citep{team2024chameleon,zhou2024transfusionpredicttokendiffuse,wang2024emu3} process multiple modalities within a shared representation space, enabling richer interaction but potentially increasing cross-modal interference.
Recent sparse architectures seek a better balance between sharing and specialization: mixture-of-experts methods~\citep{shen2024mome,li2025unimoe,lin2026moe,zhou2025unialign} introduce specialized expert pathways, with low-rank expert variants~\citep{chen2024octavius} reducing the cost of such specialization, while Mixture-of-Transformers and related architectures~\citep{liang2025mixture,deng2025emerging,shi2026lmfusion} more explicitly decouple modality-specific computation while retaining shared attention.
These mechanisms, however, have been developed primarily for semantically well-aligned modalities such as image, video, and audio.
\method instead targets weakly grounded embodied sensory signals, keeping self-attention shared while adding lightweight modality-specific residual experts to selected feed-forward layers.

\section{Datasets and Evaluation Protocols}
\subsection{Supplementary Citations}
\label{app:supp_citations}
Due to space limitations in the main text, the citations of the evaluation benchmarks and compared baselines reported in Tables~\ref{tab:general_capability}--\ref{tab:component_ablation} are collected in this section instead of in the corresponding captions.
Table~\ref{tab:cite_benchmarks} lists the evaluation benchmarks, grouped by modality following the organization of \S\ref{app:datasets_and_benchmarks}.
Tables~\ref{tab:cite_general_models} and~\ref{tab:cite_specialized_models} list the closed-source and general multimodal baselines and the modality-specialized baselines, respectively, using the same model groups as the main-text tables.
Model variants that differ only in parameter scale share a single citation and are therefore listed once.

\begin{table*}[t]
\centering
\caption{
\textbf{Supplementary citations of the evaluation benchmarks.}
Benchmarks are grouped by modality following the organization of \S\ref{app:datasets_and_benchmarks}.
}
\vspace{-8pt}
\label{tab:cite_benchmarks}

\fontsize{8.0pt}{9.0pt}\selectfont
\setlength{\tabcolsep}{3pt}
\renewcommand{\arraystretch}{0.97}

\begin{tabular*}{\textwidth}{
@{\extracolsep{\fill}}
lllll
@{}
}
\toprule
Modality & Benchmark & Citation & Benchmark & Citation \\
\midrule

\rowcolor{generalcolor}
\multicolumn{5}{c}{\textit{General Multimodal Benchmarks}} \\

Image-Text
& TextVQA & \citet{singh2019towards}
& MME & \citet{NEURIPS2025_d79a27cf} \\

& MMBench & \citet{liu2024mmbench}
& & \\

Audio-Image-Text
& DocVQA & \citet{mathew2021docvqa}
& ChartQA & \citet{masry2022chartqa} \\

Video-Text
& Video-MME & \citet{fu2025video}
 & EgoSchema & \citet{mangalam2023egoschema} \\
\rowcolor{specializedcolor}
\multicolumn{5}{c}{\textit{Embodied Sensory Benchmarks}} \\

Depth
& SpatialBench & \citet{cai2024spatialbot}
& RoboSpatial-Home & \citet{song2025robospatial} \\

Point Cloud
& ScanQA & \citet{azuma2022scanqa}
& CLEVR3D-REAL & \citet{yan2023clevr3d} \\

IMU
& HHAR & \citet{stisen2015hhar}
& MotionSense & \citet{malekzadeh2019motionsense} \\

& Shoaib & \citet{shoaib2014fusion}
& UCI-HAR & \citet{anguita2013uci} \\

Thermal
& Infrared-Benchmark & \citet{jiang2024infrared}
& IF-Bench & \citet{zhang2026ifbench} \\

Tactile
& Octopi & \citet{yu2024octopi}
& VTBS & \citet{xie2025universal} \\

Radar
& MM-Fi & \citet{yang2023mmfi}
& HoloLLM CrossEnv & \citet{zhou2026holollm} \\

Motion
& QuAnTS & \citet{divo2025quants}
& & \\

Smell
& SmellNet-Base & \citet{feng2026smellnet}
& & \\

\bottomrule
\end{tabular*}
\vspace{-2mm}
\end{table*}

\begin{table*}[t]
\centering
\caption{
\textbf{Supplementary citations of the closed-source and general multimodal baselines.
Models follow the same groups as the main-text tables.
}}
\vspace{-8pt}
\label{tab:cite_general_models}

\fontsize{8.0pt}{9.0pt}\selectfont
\setlength{\tabcolsep}{3pt}
\renewcommand{\arraystretch}{0.97}

\begin{tabular*}{\textwidth}{
@{\extracolsep{\fill}}
llll
@{}
}
\toprule
Model & Citation & Model & Citation \\
\midrule

\rowcolor{closedcolor}
\multicolumn{4}{c}{\textit{Closed-Source Models}} \\

GPT-4o & \citet{hurst2024gpt}
& Claude-3.5 & \citet{anthropic2024claude35} \\

Gemini-2.5-Pro & \citet{comanici2025gemini}
& & \\

\rowcolor{generalcolor}
\multicolumn{4}{c}{\textit{General MLLMs}} \\

LLaVA-1.5 & \citet{Liu_2024_CVPR}
& LLaVA-NeXT & \citet{liu2024llavanext} \\

LLaVA-OneVision & \citet{li2025llavaonevision}
& LLaVA-Video & \citet{zhang2025llavavideo} \\

Qwen2-VL & \citet{wang2024qwen2}
& Qwen2.5-VL & \citet{qwen2vl} \\

Qwen3-VL & \citet{bai2025qwen3}
& InternVL2 & \citet{chen2024expanding} \\

InternVL2.5& \citet{chen2024expanding}
& InternVL3.5 & \citet{wang2025internvl3} \\

InternVL3& \citet{zhu2025internvl3}
& InternVideo2.5& \citet{wang2025internvideo2} \\

Kimi-VL & \citet{team2025kimi} 
& GLM-4.1V & \citet{hong2025glm} \\

GLM-4.5V & \citet{hong2025glm}
&  Vicuna & \citet{vicuna2023}\\

TokenFlow & \citet{qu2025tokenflow}
& Tokenpacker & \citet{li2025tokenpacker} \\

Honeybee & \citet{cha2024honeybee}
&  \\

\rowcolor{generalcolor}
\multicolumn{4}{c}{\textit{General Omnimodal MLLMs}} \\

ImageBind & \citet{girdhar2023imagebind}
& ImageBind-LLM & \citet{han2023imagebind} \\

PandaGPT & \citet{su2023pandagpt}
& OneLLM & \citet{han2024onellm} \\

Intern-Omni & \citet{opengvlab2024internomni}
& Qwen3-Omni & \citet{xu2025qwen3} \\

Lyra & \citet{zhong2025lyra}
& & \\

\bottomrule
\end{tabular*}
\vspace{-2mm}
\end{table*}

\begin{table*}[t]
\centering
\caption{
\textbf{Supplementary citations of the modality-specialized baselines.}
Models follow the same groups as the main-text tables.
}
\vspace{-8pt}
\label{tab:cite_specialized_models}

\fontsize{8.2pt}{9.0pt}\selectfont
\setlength{\tabcolsep}{3pt}
\renewcommand{\arraystretch}{0.94}

\begin{tabular*}{\textwidth}{
@{\extracolsep{\fill}}
lllll
@{}
}
\toprule
\textbf{Group}
& \textbf{Model}
& \textbf{Citation}
& \textbf{Model}
& \textbf{Citation} \\
\midrule

\textbf{Spatial}
& SpatialVLM
& \citet{chen2024spatialvlm}
& Bunny
& \citet{he2024efficient} \\

&
SpatialBot
& \citet{cai2024spatialbot}
& SpaceThinker
& \citet{batra2025spatialthinker} \\

&
SpaceOm
& \citet{remyx2025spaceom}
& SSR
& \citet{liu2026ssr} \\

&
VPT
& \citet{yu2025auto}
& SpatialVTS
& \citet{liang2026enhancing} \\[1.2pt]

\midrule

\textbf{3D}
& ScanQA
& \citet{azuma2022scanqa}
& Scene-LLM
& \citet{Fu_2025_WACV} \\

&
LL3DA
& \citet{Chen_2024_CVPR}
& ChatScene
& \citet{NEURIPS2024_cebbd24f} \\

&
ViewMind3D
& \citet{chiang2026viewmind3d}
& GPT4Scene
& \citet{ICLR2026_fef12656} \\

&
ReferIt3D
& \citet{achlioptas2020referit_3d}
& BEV+MCAN
& \citet{yan2023clevr3d} \\

&
PointQA
& \citet{khan2025pointqa}
& TransVQA3D
& \citet{yan2023clevr3d} \\

&
E2E-MPCQA
& \citet{qorsham2025end}
& \\[1.2pt]

\midrule

\textbf{Tactile}
& Octopi
& \citet{yu2024octopi}
& UniTac
& \citet{tu2026unitac} \\[1.2pt]

\midrule

\textbf{IMU}
& GenHAR
& \citet{hong2026genhar}
& LanHAR
& \citet{yan2025large} \\

&
LLaSA
& \citet{imran2024llasa}
& &
\\[1.2pt]

\midrule

\textbf{Thermal}
& Infrared-LLaVA
& \citet{jiang2024infrared}
& &
\\[1.2pt]

\midrule

\textbf{Radar}
& HoloLLM
& \citet{zhou2026holollm}
& &
\\[1.2pt]

\midrule

\textbf{Motion}
& Naive
& \citet{divo2025quants}
& ChatTS
& \citet{xie2024chatts} \\

&
xQA
& \citet{divo2025quants}
& &
\\[1.2pt]

\midrule

\textbf{Smell}
& MLP, CNN, LSTM and ScentFormer
& \citet{feng2026smellnet}
 \\

\bottomrule
\end{tabular*}

\vspace{-2mm}
\end{table*}

\subsection{Datasets and Benchmarks}
\label{app:datasets_and_benchmarks}
\begin{figure}[t]
    \centering
    \includegraphics[width=1\textwidth]{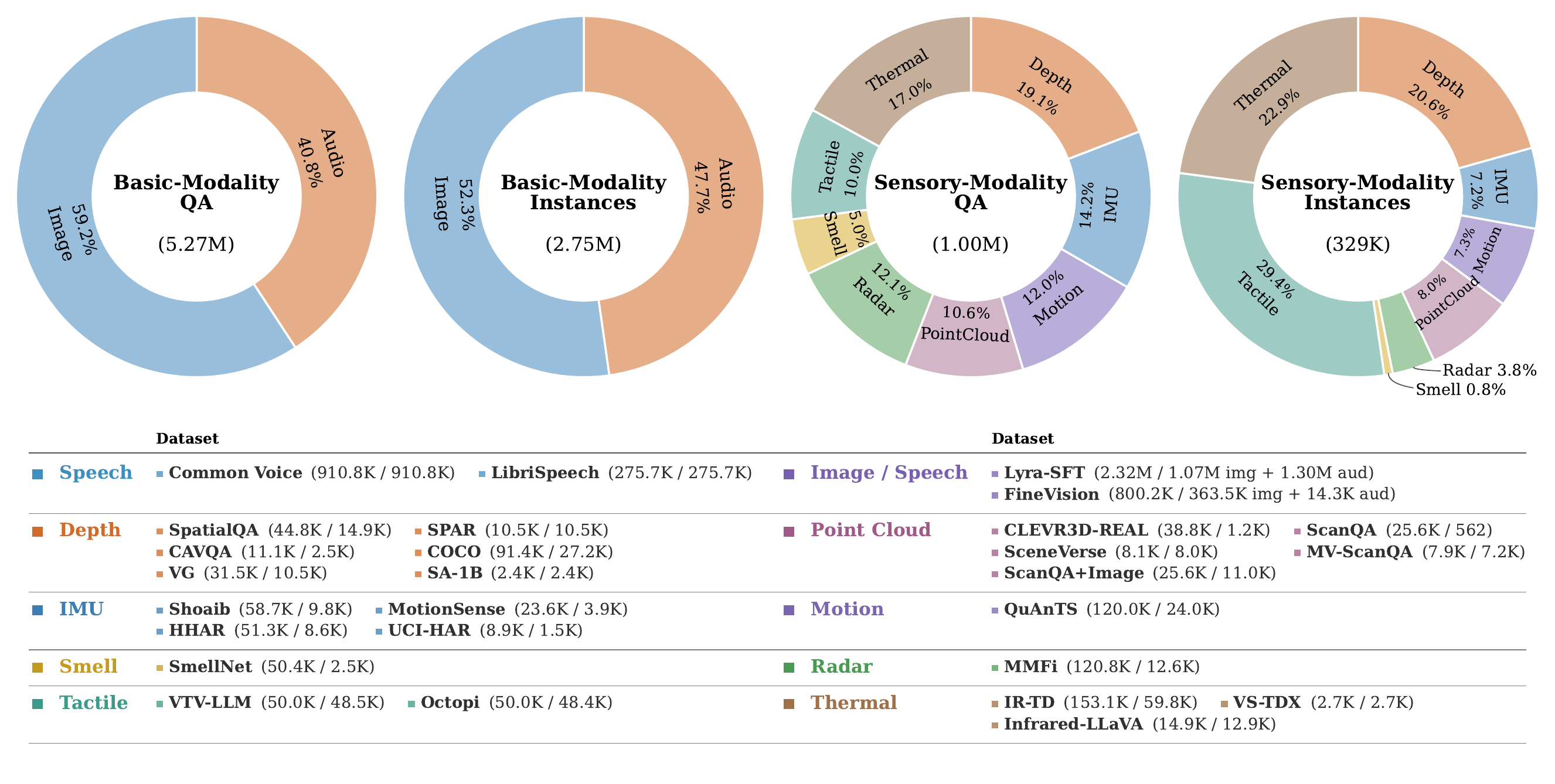}
    \caption{
    \textbf{Overview of the training data composition.}
    The donut charts illustrate the distribution of QA and unique data instances across basic and sensory modalities.
    The bottom panel provides a dataset-level breakdown for each modality, with the number of training QA and unique instances reported for every data source.
    }
    \vspace{-8pt}
    \label{fig:training_data_overview}
\end{figure}

This section describes the datasets and evaluation protocols used for each modality. We first detail the training data sources and preparation, with the overall data composition summarized in Fig.~\ref{fig:training_data_overview}, and then describe the evaluation benchmarks, data splits, input settings, and metrics. Unless otherwise specified, we follow the official benchmark splits and scoring protocols. During evaluation, \method receives only the benchmark-specified inputs, without access to training-only captions or auxiliary semantic annotations.

\subsubsection{General Multimodal}
Following the multimodal initialization pipeline of \citet{zhong2025lyra}, we first establish the image--speech--text capabilities of \method before sensory training. Speech alignment uses Common Voice~\citep{ardila2020common} and LibriSpeech~\citep{panayotov2015librispeech}, comprising approximately 1.19M speech--text pairs. We then perform image--speech--text instruction tuning on a mixture of Lyra-SFT~\citep{zhong2025lyra} and selected subsets of FineVision~\citep{wiedmann2025finevision}, containing approximately 3.12M conversations. The mixture consists primarily of Lyra-SFT together with additional FineVision image--text data. 

We evaluate the general multimodal capabilities on image--text, image--speech, and video benchmarks. Specifically, we use TextVQA~\citep{singh2019towards}, MME~\citep{NEURIPS2025_d79a27cf}, and MMBench~\citep{liu2024mmbench} for image--text understanding; the speech-conditioned variants of TextVQA, DocVQA~\citep{mathew2021docvqa}, and ChartQA~\citep{masry2022chartqa} for image--speech understanding; and Video-MME~\citep{fu2025video} and EgoSchema~\citep{mangalam2023egoschema} for video understanding. We follow the standard evaluation protocols for all benchmarks, with Video-MME evaluated without subtitles.

\subsubsection{Depth}
For depth training, we use SpatialQA~\citep{cai2024spatialbot}, SPAR~\citep{zhang2025spar}, and CA-VQA~\citep{daxberger2025mm}. We additionally construct depth-oriented training samples from COCO~\citep{mishra2019ocr}, Visual Genome (VG)~\citep{krishna2017visual}, and SA-1B~\citep{kirillov2023segment}. These datasets cover diverse depth-centric scenes and task formats, including spatial question answering, 3D reasoning, captioning, and general multimodal instruction tuning. We synthesize speech with CosyVoice~\citep{du2024cosyvoice} for SpatialQA, SPAR, CA-VQA, and part of the constructed depth subset. For samples without depth maps, we generate them using Depth Anything V2~\citep{yang2024depth}. 

For sensory representation alignment, each RGB--D sample is paired with a depth-aware description generated offline by Qwen3-Omni-30B-A3B-Instruct~\citep{xu2025qwen3}: the model first extracts objects from RGB and then generates a depth-focused caption conditioned on the RGB image, depth map, and extracted objects. 

We evaluate depth understanding on SpatialBench and RoboSpatial-Home~\citep{song2025robospatial} using paired RGB--D observations. SpatialBench comprises five subtasks: positional reasoning, object existence, counting, reaching, and size comparison. We report each subtask score, together with \textit{4-Task Agg.}, the sample-weighted aggregate over Position, Existence, Counting, and Size, and \textit{Agg.}, the aggregate over all five subtasks. All scores follow the official SpatialBench evaluation protocol. We further construct \textit{SpatialBench-Speech} by converting the question text into speech to assess the robustness of depth reasoning to spoken queries. For RoboSpatial-Home, we report zero-shot accuracy on the \textit{Configuration} split without benchmark-specific fine-tuning.

\subsubsection{Point Cloud}
For point-cloud training, we use CLEVR3D-REAL~\citep{yan2023clevr3d}, ScanQA~\citep{azuma2022scanqa}, and SceneVerse~\citep{jia2024sceneverse}. Together, these datasets cover scene-level descriptions, question answering, and compositional reasoning over indoor 3D scenes. We use the official training splits of ScanQA and CLEVR3D-REAL. 
For ScanQA, we further construct RGB-augmented examples using single-view and four-view images from MV-ScanQA~\citep{mo2025advancing}, yielding both point-cloud-only and point-cloud--image training inputs. 
For sensory representation alignment, scene-level descriptions from SceneVerse are matched to ScanQA and CLEVR3D-REAL samples by scene identity.

We evaluate point-cloud understanding on ScanQA and CLEVR3D-REAL (results reported in Table~\ref{tab:clevr3d_real}), using their official validation splits. 
On ScanQA, we evaluate both point-cloud-only (\texttt{xyz}) and point-cloud--image (\texttt{xyz+img}) inputs. Following the standard ScanQA protocol, we report CIDEr, METEOR (MET.), ROUGE-L (R-L), BLEU-1 (B-1), and BLEU-4 (B-4) for open-ended 3D question answering. 
CLEVR3D-REAL evaluates compositional reasoning over point-cloud scenes across existence, counting, comparison, query-attribute, query-object, and compare-attribute questions. We report accuracy for each question category, together with \textit{Overall}, denoting accuracy over all test questions, and \textit{Cls. Avg.}, denoting the unweighted average accuracy across question categories.

\subsubsection{IMU}
For IMU training, we use the SensorCaps and OpenSQA annotations~\citep{imran2024llasa} associated with four human activity recognition datasets: HHAR~\citep{stisen2015hhar}, MotionSense~\citep{malekzadeh2019motionsense}, Shoaib~\citep{shoaib2014fusion}, and UCI-HAR~\citep{anguita2013uci}. 
SensorCaps provides a natural-language description for each IMU window and is used for sensory representation alignment, while OpenSQA provides question--answer pairs covering activity-related understanding and reasoning. 
We construct the training and evaluation sets using stratified window-level hold-out within each source dataset, reserving approximately 100 windows per activity class for evaluation and using the remaining windows for training, with no overlap between the two splits. 
This results in approximately 23.8k training windows and 2.5k held-out evaluation windows, with multiple OpenSQA questions associated with each training window.

We evaluate IMU understanding as open-ended activity classification on the held-out windows from HHAR, MotionSense, Shoaib, and UCI-HAR, following the LLaSA~\citep{imran2024llasa} classification protocol. 
For each dataset, the model is provided with the IMU sequence and generates the predicted class name directly from the input. We report results separately for the four datasets using accuracy and macro-averaged precision, recall, and F1. 

\subsubsection{Tactile}
For tactile training, we use Octopi~\citep{yu2024octopi} and the Universal Visuo-Tactile Video Understanding corpus (VTBS)~\citep{xie2025universal}. Both datasets follow object-level splits, such that objects used for evaluation are excluded from training. 
Following the official data generation protocols, we construct 50k instruction--response pairs from each corpus, covering tactile property description, comparison, selection, and object--property reasoning, resulting in approximately 100k training examples in total. 
The original GelSight~\citep{yuan2017gelsight} videos are converted into a unified image-based representation by first selecting informative motion frames and then uniformly sampling three RGB frames from each video, which are treated as tactile observations during training. 
For sensory representation alignment, we extract the descriptive property text from the generated responses and use it as the corresponding tactile caption.

We evaluate on the official held-out-object splits of Octopi using the same three-frame tactile representation as in training. 
Octopi contains 498 questions across property comparison (PC), property superlative selection (PSS), and property--object matching (POM). Following the official protocol, we report accuracy separately for the three tasks.

\subsubsection{Thermal}
For thermal training, we use a partial subset of IR-TD from IRGPT~\citep{cao2025irgpt}, the thermal subset of VS-TDX~\citep{chung2026enhanced}, and Infrared-LLaVA~\citep{jiang2024infrared}. 
Together, these datasets cover thermal captioning, question answering, recognition, counting, and localization. 
For IR-TD, we use its caption, instruction-QA, and task-oriented training subsets, and pair thermal images with the corresponding RGB observations when available. 
For sensory representation alignment, we use the paired RGB--thermal captions provided by IR-TD. 

We evaluate thermal understanding on IF-Bench~\citep{zhang2026ifbench}
(results reported in Table~\ref{tab:ifbench}), and the Benchmark from Infrared-LLaVA, using thermal inputs for both benchmarks. 
IF-Bench is evaluated in a zero-shot setting and groups its tasks into three categories: coarse-grained perception (\textit{Coarse}), fine-grained perception (\textit{Fine}), and image reasoning (\textit{Reason}). 
We report accuracy for each of the ten tasks and the overall average. 
Infrared-LLaVA Benchmark contains six infrared question-answering tasks covering localization, aerial counting, pedestrian counting, recognition, scene understanding, and safety assessment. 
We report the average accuracy across the six tasks under both zero-shot and fine-tuning settings.

\subsubsection{Radar}
For radar training, we use mmWave point-cloud sequences from MM-Fi~\citep{yang2023mmfi} together with the language annotations introduced by HoloLLM~\citep{zhou2026holollm}. We follow the HoloLLM Cross-Environment protocol, using environments E01--E03 for training and holding out E04 for evaluation, such that the training and test sets contain disjoint environments and subjects. 
The training split contains approximately 12.6k temporal radar segments. Based on the HoloLLM action annotations, we construct a mixture of action captions and question--answering examples, and further augment the instruction data with laterality questions and multiple-choice questions targeting confusable action categories, resulting in approximately 121k training examples. 
The original HoloLLM captions are further refined offline using GPT-5.1~\citep{openai2025gpt51} with action labels as guidance. The refined captions are used for sensory representation alignment.

We evaluate radar understanding following the HoloLLM Cross-Environment protocol on the held-out E04 environment. We report results on two tasks using radar-only input: Human Action QA and Action Caption. 
For Human Action QA, the model predicts one of the predefined human action categories from the radar sequence, and performance is measured using accuracy following the HoloLLM matching protocol.
For Action Caption, the model generates a free-form description of the observed action, and we report METEOR as the primary metric.

\subsubsection{Motion}
For skeleton-motion training, we use QuAnTS~\citep{divo2025quants}, which contains synthetic SMPL~\citep{loper2023smpl} motion sequences paired with binary, multiple-choice, and open-ended question answering. 
We follow the official 80/10/10 trajectory-level split and use only the training partition, corresponding to 24k motion sequences and 120k QA pairs. 
For sensory representation alignment, we use the scene-level textual descriptions provided by QuAnTS as auxiliary captions.

We evaluate on the official QuAnTS test split, which contains 3k held-out motion trajectories and 15k questions spanning binary, multiple-choice, and open-ended QA. At inference time, the model receives only the skeleton-motion sequence and the corresponding question, without access to scene descriptions or ground-truth action sequences. 
We report binary and multiple-choice performance using accuracy and macro-averaged precision, recall, and F1, while open-ended responses are evaluated using ROUGE-L and METEOR.

\subsubsection{Smell}
For smell training, we use SmellNet-Base~\citep{feng2026smellnet}, which contains 50 single-substance categories recorded with a six-channel metal-oxide (MOX) sensor array. We follow the official day-wise split, using recordings from the first five days for training and the sixth day for evaluation, yielding approximately 2.5k training windows and 0.5k held-out test windows after temporal segmentation. 
We segment the aligned sensor readings into fixed-length windows and normalize each channel using training-set statistics. From these windows, we construct approximately 50.4k instruction samples spanning classification, reasoning, and captioning, and generate temporal descriptions that exclude the substance identity for the alignment stage.

We evaluate on the held-out day-6 split of SmellNet-Base (502 windows across 50 classes), using training-set normalization statistics. Given a sensor sequence and the full candidate set, the model predicts the substance name; we report Acc@1 and macro F1.

\section{Implementation Details}
\label{app:implementation_detatils}
\subsection{Baselines}
\label{app:baselines_and_metrics}
We compare UniSense with three groups of baselines: (i) \textbf{closed-source models}, including models from the GPT, Claude and Gemini families; (ii) \textbf{general multimodal models}, such as the Qwen-VL and InternVL families; and (iii) \textbf{modality-specialized models} designed for specific sensory modalities or tasks. For each benchmark, we follow its standard evaluation protocol and report the corresponding metrics.

\subsection{Tokenizers and Projectors}
\label{app:tokenization_projectors}
We adopt two tokenization pathways according to the structure of each sensory input.

\noindent\textbf{Pixel-aligned modalities.}
Depth, tactile, and thermal signals are represented as image-like inputs and processed using the default Qwen visual preprocessing. They are divided into patches by the pretrained visual patch embedding and then encoded by the shared vision encoder and merger. 

\noindent\textbf{Non-pixel modalities.}
Point cloud, IMU, radar, motion, and smell inputs have distinct spatial or temporal structures that cannot be directly processed by visual patch embedding. We therefore introduce a lightweight tokenizer for each modality to map its native input into a sequence of embeddings with the same hidden dimension as the visual patch tokens. These embeddings are directly fed into the shared vision encoder and merger. The tokenizer design for each modality is described below.

\noindent\textbf{Point cloud.}
Each point cloud $\mathbf{x}^{\mathrm{pc}}$ is represented by normalized XYZ coordinates and RGB features. We sample $32{,}768$ points from each scene and organize them into $1{,}024$ local groups, each containing $32$ neighboring points. Within each group, the point features are augmented with coordinates relative to the group center. A point-wise $1\times1$ convolution projects the resulting features to the vision hidden dimension, followed by max pooling over the local neighborhood:

$$
\mathcal{T}_{\mathrm{pc}}(\mathbf{x}^{\mathrm{pc}})
=
\operatorname{MaxPool}\!\left(
f_{\mathrm{pc}}
\left(
[\mathbf{F}^{\mathrm{pc}},\Delta\mathbf{P}^{\mathrm{pc}}]
\right)
\right).
$$

\noindent\textbf{IMU.}Each IMU input $\mathbf{x}^{\mathrm{imu}}$ consists of six channels corresponding to tri-axial accelerometer and gyroscope measurements. Each sequence contains $120$ temporal steps. Inputs of a different length are linearly resampled to this length before encoding. We employ a length-preserving $1$D convolution with kernel size $10$ to map the input sequence to the vision hidden dimension:
\[
\mathcal{T}_{\mathrm{imu}}(\mathbf{x}^{\mathrm{imu}})
=
\operatorname{Conv1D}_{6\rightarrow d_{\mathrm{vit}}}^{(k=10)}
\left(
\mathbf{x}^{\mathrm{imu}}
\right).
\]
The tokenizer produces $120$ IMU tokens, which are subsequently processed by the shared vision encoder and reduced to $30$ tokens by the vision merger in $\mathcal{E}$.

\noindent\textbf{Radar.}
Each radar input $\mathbf{x}^{\mathrm{radar}}$ is represented as a temporal sequence of point sets, where each point contains five attributes: 3D coordinates, Doppler, and intensity. Each segment is uniformly sampled to at most $32$ frames, with each frame containing at most $64$ points. For each frame, the original point features are augmented with coordinates relative to the masked centroid. A point-wise projection followed by masked max pooling produces a frame-level representation:
\[
\mathcal{T}_{\mathrm{radar}}(\mathbf{x}^{\mathrm{radar}})
=
\operatorname{MaskedMaxPool}\!\left(
f_{\mathrm{radar}}
\left(
[\mathbf{X}^{\mathrm{radar}},\Delta\mathbf{P}^{\mathrm{radar}}]
\right)
\right).
\]
The resulting frame-level representations are temporally interpolated to $64$ radar tokens, which are subsequently processed by the shared vision encoder and reduced to $16$ tokens by the vision merger in $\mathcal{E}$. Padded points are excluded from both centroid computation and feature pooling.

\noindent\textbf{Motion.}
Each skeleton--motion input $\mathbf{x}^{\mathrm{motion}}$ is represented as a sequence of $24$ skeleton joints with 3D coordinates. Each frame is flattened into a $72$-dimensional feature vector, and each sequence is linearly resampled to $320$ temporal steps. We employ a length-preserving $1$D convolution with kernel size $10$ to map the input sequence to the vision hidden dimension:
\[
\mathcal{T}_{\mathrm{motion}}(\mathbf{x}^{\mathrm{motion}})
=
\operatorname{Conv1D}_{72\rightarrow d_{\mathrm{vit}}}^{(k=10)}
\left(
\mathbf{x}^{\mathrm{motion}}
\right).
\]
The tokenizer produces $320$ motion tokens, which are subsequently processed by the shared vision encoder and reduced to $80$ tokens by the vision merger in $\mathcal{E}$.

\noindent\textbf{Smell.}
Each smell input $\mathbf{x}^{\mathrm{smell}}$ consists of temporal measurements from six metal-oxide gas sensors.  Each sequence contains $100$ temporal steps, and each channel is standardized using its training-set mean and standard deviation. We employ a length-preserving $1$D convolution with kernel size $10$ to project the signal sequence to the vision hidden dimension:
\[
\mathcal{T}_{\mathrm{smell}}(\mathbf{x}^{\mathrm{smell}})
=
\operatorname{Conv1D}_{6\rightarrow d_{\mathrm{vit}}}^{(k=10)}
\left(
\mathbf{x}^{\mathrm{smell}}
\right).
\]
The tokenizer produces $100$ smell tokens, which are subsequently processed by the shared vision encoder and reduced to $25$ tokens by the vision merger in $\mathcal{E}$.

\noindent\textbf{Modality-specific projectors.}
After the shared vision encoder and merger, each sensory modality
$s\in\mathcal{M}_{\mathrm{s}}$ is equipped with an independent residual projector $\mathcal{P}_s$.
Let
$\mathbf{H}^{s}=[\mathbf{h}^{s}_1,\ldots,\mathbf{h}^{s}_{N_s}]$
denote the merged representation of modality $s$, where
$\mathbf{h}^{s}_i\in\mathbb{R}^{d_{\mathrm{llm}}}$.
The projector is applied independently to each token:
\[
\mathbf{z}^{s}_i
=
\mathbf{h}^{s}_i
+
\mathbf{W}_{2}^{s}
\operatorname{ReLU}
\left(
\mathbf{W}_{1}^{s}\mathbf{h}^{s}_i
\right),
\]
and the projected sensory representation is
$\mathbf{Z}^{s}=[\mathbf{z}^{s}_1,\ldots,\mathbf{z}^{s}_{N_s}]
=\mathcal{P}_s(\mathbf{H}^{s})$.
Here,
$\mathbf{W}_{1}^{s}\in\mathbb{R}^{d_p\times d_{\mathrm{llm}}}$
and
$\mathbf{W}_{2}^{s}\in\mathbb{R}^{d_{\mathrm{llm}}\times d_p}$,
with $d_p=768$.
The output layer is zero-initialized, such that the projector starts as an identity mapping.

\subsection{Basic Modality Processing}
\label{app:basic_modalities}
Unlike embodied sensory inputs, basic modalities retain the processing pathways inherited from the pretrained multimodal backbone. 

\noindent\textbf{Image pathway.}
RGB images are processed by the native Qwen3.5 visual encoder and merger. The resulting image tokens are directly inserted into the language sequence without an additional learnable projector, and the entire pathway remains unchanged during sensory training.

\noindent\textbf{Speech pathway.}
Speech signals are encoded using Whisper-large-v3-turbo~\citep{radford2023robust}. The resulting acoustic features are mapped to the language hidden dimension through a speech projector. The Whisper encoder remains frozen, while the speech projector is pretrained to align acoustic representations with the language model.

\noindent\textbf{Multimodal initialization.}
Before sensory training, UniSense is initialized from an image--speech--text instruction-tuned Qwen3.5 checkpoint, with its original image and speech pathways retained.

\subsection{Sensory Representation Alignment}
\label{app:csa_impl}
We provide additional implementation details for the sensory representation alignment stage introduced in Sec.~\ref{sec:csa}. Unless otherwise specified, the following settings correspond to the Qwen3.5-9B configuration.

\noindent\textbf{Conditional captioning.}
We use a lightweight 6-layer caption decoder with a hidden dimension of $1{,}024$ and $16$ attention heads. The decoder reuses the frozen token embeddings and language-modeling head of Qwen, while its remaining parameters are trainable. Cross-attention uses only the projected tokens of the target sensory modality as memory, and the decoder output is used solely for auxiliary caption reconstruction rather than the main response-generation pathway. The caption decoder remains trainable in both sensory-training stages.


\noindent\textbf{Contrastive alignment.}
The projected sensory tokens and caption tokens are independently processed by the first $k$ frozen layers of the language backbone. In the default configuration, we use $k=8$. We apply mean pooling over valid sensory tokens and use the final valid caption token as the textual representation. The two representations are then passed through separate LayerNorm and linear projection heads, mapped to $512$ dimensions, and $\ell_2$-normalized before computing the symmetric InfoNCE loss.

Representations are gathered across all distributed workers to construct global in-batch negatives. Pairs corresponding to the same underlying sensory instance or sharing an identical normalized caption are excluded from the negative set to reduce false-negative supervision. The contrastive branch is used only during the first sensory-training stage and is disabled after the transition to modality-specific expert tuning.

\subsection{Modality-Specific Expert Tuning}
\label{app:msc_impl}
During the second sensory-training stage, modality-specific residual experts are activated in the selected full-attention decoder layers, while the pretrained self-attention and base FFN remain frozen. For the 9B model, expert modules are inserted into the eight full-attention layers at indices 3, 7, 11, 15, 19, 23, 27, and 31.

\noindent\textbf{Expert configuration.}
Each embodied sensory modality uses an independent router and a pool of $E=3$ residual experts. The router produces softmax weights over all experts in the corresponding modality-specific pool. Text, image, and speech tokens bypass these expert pools and retain the original pretrained FFN computation.

\noindent\textbf{Residual parameterization.}
Each expert applies low-rank residual adaptations to the gate, up, and down projections of the backbone SwiGLU FFN. We use rank $r=32$ and scaling factor $\alpha=64$, corresponding to a residual scaling factor of $\alpha/r=2$.

\noindent\textbf{Regularization.}
We use router load-balancing together with z-loss regularization, with a combined weight of $10^{-3}$. An additional expert-diversity objective is applied with a target weight of $0.07$, which is linearly increased from zero during the early expert-tuning stage to stabilize optimization.

\subsection{Training Details}
\label{app:training_details}
Unless otherwise specified, the following settings correspond to the Qwen3.5-9B configuration; the 2B model follows the same two-stage training procedure with scale-specific configurations.

\noindent\textbf{Multimodal initialization.}
We construct a common multimodal initialization checkpoint before sensory training. We first connect a frozen Whisper encoder to the language model by pretraining the speech projector. We then perform image--speech--text instruction tuning, where the language model is adapted with rank-$64$ LoRA while the speech projector remains trainable. No additional video-specific training is introduced during these preliminary stages. These stages establish the basic-modality pathways and are not counted as part of the two-stage sensory training of \method.

\noindent\textbf{Two-stage training.}
Starting from the merged multimodal checkpoint, the pretrained language backbone and modality encoders remain frozen throughout sensory training. In the alignment Stage, we train the sensory tokenizers, modality projectors, caption decoders, and contrastive projection heads for 1 epoch using language-modeling, caption-reconstruction, and contrastive objectives.

The tuning Stage continues optimizing the sensory tokenizers, projectors, and caption decoders, while activating the modality-specific routers and residual experts. The contrastive branch is disabled after the stage transition, whereas caption supervision is retained together with the expert regularization objectives. Stage~2 is trained for 2 additional epochs.

\begin{table}[t]
\centering
\small
\setlength{\tabcolsep}{4.0pt}
\renewcommand{\arraystretch}{1.08}

\caption{
Stage-wise training configuration of UniSense.
The reported learning rate is the base learning rate;
modality projectors use $1.5\times10^{-4}$.
}
\label{tab:training_stages}

\begin{tabular}{@{}l p{0.46\columnwidth} cc@{}}
\toprule
Phase & Trainable modules & Base LR & Epochs \\
\midrule

Speech alignment
& Speech projector
& $1\times10^{-3}$
& 1 \\

Multimodal initialization
& LLM LoRA, speech projector
& $2\times10^{-4}$
& 1 \\

Alignment Stage
& Tokenizers, projectors, caption decoder, contrastive heads
& $1\times10^{-4}$
& 1 \\

Tuning Stage
& Tokenizers, projectors, caption decoder, routers, experts
& $1\times10^{-4}$
& 2 \\

\bottomrule
\end{tabular}
\end{table}

\noindent\textbf{Loss configuration.}
During the alignment Stage, the caption and contrastive objectives are jointly applied with the language-modeling objective. We set the contrastive weight to $\lambda_{\mathrm{con}}=0.5$ and use an overall alignment weight of $\lambda_{\mathrm{align}}=0.08$ for all modalities. In the tuning Stage, the contrastive objective is removed while caption reconstruction is retained. The router regularization is weighted by $10^{-3}$, and the expert-diversity weight is linearly increased from $0$ to $0.07$ during the early expert-tuning stage.

\noindent\textbf{Optimization.}
We use AdamW with cosine learning-rate decay and a $5\%$ warm-up ratio. Modality projectors use a learning rate of $1.5\times10^{-4}$, while the remaining trainable sensory modules use a base learning rate of $1\times10^{-4}$. The maximum sequence length is $2{,}048$ for speech alignment and $4{,}096$ for both multimodal initialization and sensory training. All models are trained in bfloat16 on 4 NVIDIA H200 GPUs.

\subsection{Ablation Settings}
\label{app:ablation_settings}

All ablation experiments are conducted with the Qwen3.5-9B backbone. Unless otherwise specified, we keep the training data, modality tokenizers and projectors, optimization schedule, and evaluation protocol unchanged across variants.

\begin{figure}[t]
    \centering
    \begin{subfigure}[t]{0.30\textwidth}
        \vspace{0pt}
        \centering
        \includegraphics[width=0.95\linewidth]
        {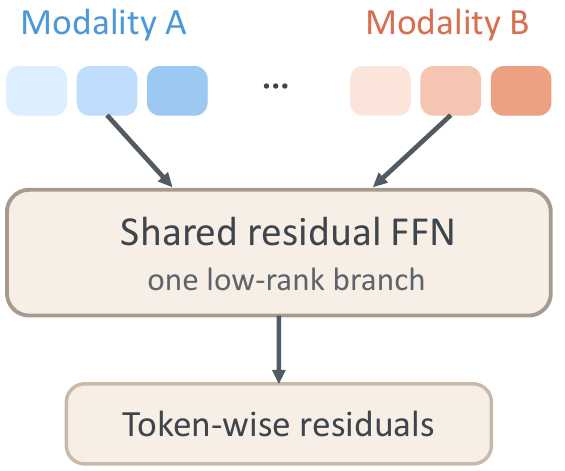}
        \caption{Shared Residual FFN}
        \label{fig:expert_shared_ffn}
    \end{subfigure}
    \hfill
    \begin{subfigure}[t]{0.28\textwidth}
        \vspace{0pt}
        \centering
        \includegraphics[width=0.95\linewidth]
        {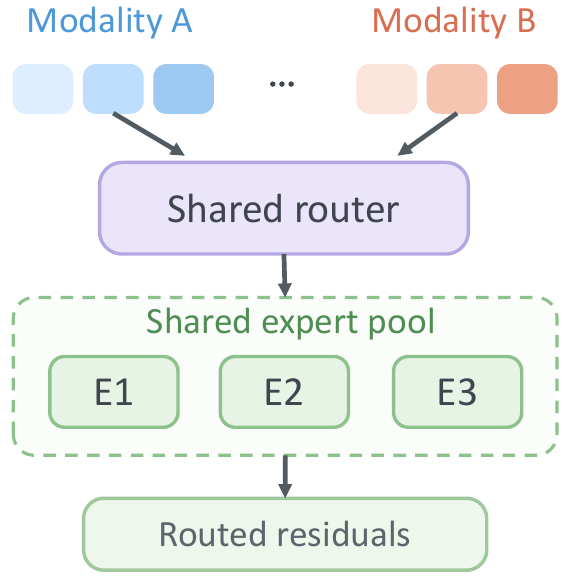}
        \caption{Shared Expert Pool}
        \label{fig:expert_shared_pool}
    \end{subfigure}
    \hfill
    \begin{subfigure}[t]{0.36\textwidth}
        \vspace{0pt}
        \centering
        \includegraphics[width=0.95\linewidth]
        {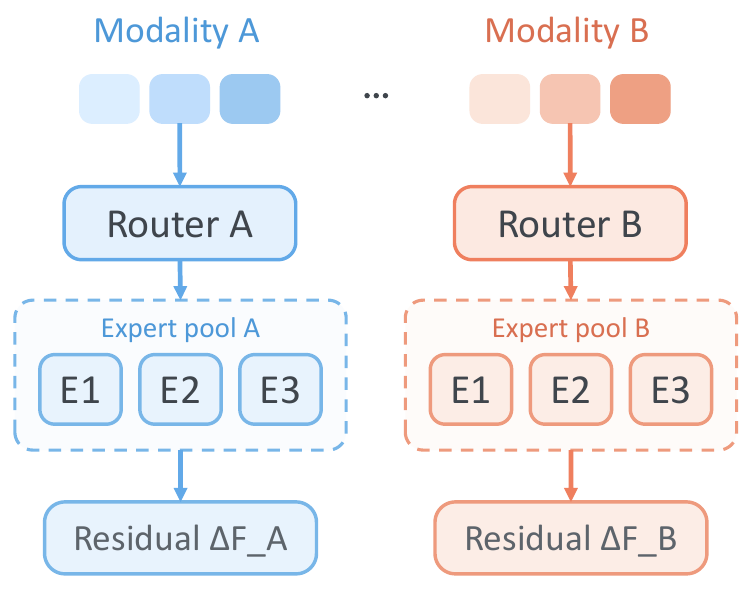}
        \caption{Modality-Specific Expert Pools}
        \label{fig:expert_modality_specific}
    \end{subfigure}

    \caption{
        \textbf{Illustration of the tuning Stage adaptation architectures.}
        The shared residual FFN in (a) is used in the component ablation,
        while (b) and (c) compare shared and modality-specific expert pools.
    }
    \label{fig:expert_architecture_comparison}
\end{figure}

\noindent\textbf{Component ablation.}
We evaluate the contributions of sensory representation alignment and modality-specific expert tuning using the three configurations summarized in Table~\ref{tab:component_ablation_settings}. All variants follow the same two-stage training schedule. The baseline uses only the language-modeling loss on downstream instruction--response pairs during the alignment stage, followed by a single low-rank residual FFN shared across all sensory modalities without routing during the tuning stage(Fig.~\ref{fig:expert_shared_ffn}). We then introduce conditional captioning and contrastive alignment while retaining the same shared adaptation. The full model further replaces the shared residual FFN with modality-specific routers and expert pools.

\begin{table}[t]
\centering
\small
\setlength{\tabcolsep}{5pt}
\renewcommand{\arraystretch}{1.08}
\caption{Configurations used for the main component ablation.}
\label{tab:component_ablation_settings}
\begin{tabular}{lcc}
\toprule
Setting & Alignment Stage & Tuning Stage \\
\midrule
Baseline
& $\mathcal L_{\mathrm{LM}}$ & Shared residual FFN \\

Representation alignment
& $\mathcal L_{\mathrm{LM}}+\mathcal L_{\mathrm{cap}}+\mathcal L_{\mathrm{con}}$ & Shared residual FFN \\

Full
& $\mathcal L_{\mathrm{LM}}+\mathcal L_{\mathrm{cap}}+\mathcal L_{\mathrm{con}}$ & Modality-specific experts \\
\bottomrule
\end{tabular}
\end{table}

The shared residual FFN is activated only in the tuning Stage, following the same training schedule as the modality-specific experts. This keeps the amount of training data and the number of optimization steps consistent across configurations. We report results on representative benchmarks spanning depth, point cloud, IMU, thermal and radar.


\noindent\textbf{Alignment objectives.}
All variants retain $\mathcal{L}_{\mathrm{LM}}$, the language-modeling loss on downstream instruction--response pairs. We independently enable the two auxiliary objectives during the alignment stage, yielding four settings: (1) task LM only; (2) task LM with conditional captioning ($\mathcal{L}_{\mathrm{cap}}$); (3) task LM with contrastive alignment ($\mathcal{L}_{\mathrm{con}}$); and (4) task LM with both objectives. All variants use the same sensory front ends, training data, and optimization settings.

\noindent\textbf{Expert architectures.}
Starting from the sensory representation-aligned setting, we further examine how expert sharing affects the tuning Stage adaptation. We compare a \emph{Shared Expert Pool} (Fig.~\ref{fig:expert_shared_pool}), where all sensory modalities route over a common pool of 3 residual experts, with \emph{Modality-Specific Expert Pools} (Fig.~\ref{fig:expert_modality_specific}), where each modality uses an independent router and expert pool. Both variants start from the same representation-aligned initialization and retain the frozen pretrained base FFN.

\section{Additional Experiments}
\label{app:additional_exp}
\noindent\textbf{Additional Benchmark Results.}
We further provide additional results on thermal and point-cloud understanding in Tables~\ref{tab:ifbench} and~\ref{tab:clevr3d_real}.
\method-9B reaches 81.03 average accuracy on IF-Bench and 50.1 class-average accuracy on CLEVR3D-REAL, demonstrating strong performance across both benchmarks.

\noindent\textbf{Expert placement.}
We study expert placement on two modalities, depth and radar, comparing four strategies: all Transformer layers, the early or late half of the full-attention layers, and all full-attention layers.
Table~\ref{tab:expert_placement} shows that expert placement matters more than simply increasing its layer coverage.
While depth exhibits some benchmark-specific preference, restricting experts to full-attention layers yields the most robust performance overall, particularly for radar where it consistently outperforms all alternative placements.

\begin{table*}[t]
\centering
\caption{
\textbf{Evaluation on IF-Bench.}
We report the overall average (\textit{Avg.}) and performance across ten infrared understanding and reasoning categories.
}
\label{tab:ifbench}

\setlength{\tabcolsep}{3.2pt}
\renewcommand{\arraystretch}{1.05}

\resizebox{\textwidth}{!}{
\begin{tabular}{lccccccccccc}
\toprule
\textbf{Model}
& \textbf{Avg.}
& \textbf{SU}
& \textbf{IT}
& \textbf{VC}
& \textbf{TL}
& \textbf{SRU}
& \textbf{OC}
& \textbf{TFU}
& \textbf{AR}
& \textbf{TFR}
& \textbf{CR} \\
\midrule

\rowcolor{generalcolor}
\multicolumn{12}{c}{\textit{General MLLMs}} \\

InternVL3-2B
& 65.6
& 91.7 & 82.1 & 61.9 & 56.3 & 44.7
& 42.1 & 55.7 & 70.0 & 68.2 & 83.1 \\

InternVL3.5-2B
& 62.3
& 90.5 & 86.3 & 61.9 & 55.7 & 44.0
& 38.5 & 47.7 & 67.5 & 57.6 & 73.0 \\

InternVL3.5-2B-Thinking
& 65.5
& 91.0 & 89.5 & 62.5 & 50.2 & 46.5
& 44.1 & 58.9 & 68.5 & 69.9 & 74.0 \\

Qwen2.5-VL-7B
& 71.1
& 93.3 & 86.5 & 74.4 & 60.3 & 49.5
& 42.7 & 58.9 & 74.0 & 88.6 & 83.1 \\

InternVL3-8B
& 71.2
& 95.0 & 87.8 & 71.6 & 66.4 & 51.9
& 45.0 & 57.7 & 79.0 & 70.8 & 87.1 \\

InternVL3.5-8B
& 69.1
& 92.6 & 79.7 & 66.8 & 66.7 & 52.3
& 45.0 & 61.8 & 67.7 & 82.8 & 75.6 \\

InternVL3.5-8B-Thinking
& 72.3
& 92.8 & 85.5 & 65.3 & 67.3 & 58.6
& 47.3 & 65.0 & 71.0 & 91.5 & 78.8 \\

LLaVA-OneVision-1.5-8B-Instruct
& 75.9
& 95.0 & 89.7 & 70.9 & 73.1 & 54.2
& 54.6 & 62.0 & 79.2 & 91.9 & 88.3 \\

Qwen3-VL-8B-Instruct
& 78.8
& 93.9 & 93.1 & 79.7 & 74.1 & \textbf{65.5}
& \textbf{60.2} & 68.6 & 75.8 & 88.3 & 88.9 \\

Qwen3-VL-8B-Thinking
& 76.5
& 94.6 & 91.2 & 74.8 & 62.0 & 65.3
& 58.0 & 69.3 & 76.7 & 89.8 & 83.1 \\

GLM-4.1V-9B-Thinking
& 77.2
& 94.4 & 92.1 & 77.8 & 67.8 & 59.7
& 55.0 & 69.3 & 77.7 & 89.4 & 88.9 \\

\midrule

\rowcolor{ourscolor}
\multicolumn{12}{c}{\textit{\textbf{\method}}} \\

\method-2B
& 74.26
& 96.15 & 86.49 & 70.69 & 61.64 & 59.26
& 51.72 & \textbf{70.00} & 72.31 & 93.22 & 83.87 \\

\method-9B
& \textbf{81.03}
& \textbf{97.44} & \textbf{94.59} & \textbf{81.03} & \textbf{80.82} & 62.96
& 59.77 & \textbf{70.00} & \textbf{80.00} & \textbf{94.92} & \textbf{90.32} \\

\bottomrule
\end{tabular}
}

\vspace{-2pt}
\end{table*}

\begin{table*}[t]
\centering
\caption{
\textbf{Evaluation on CLEVR3D-REAL.}
We report accuracy (\%) across six point-cloud compositional reasoning categories:
existence (\textit{Exist.}), counting (\textit{Count.}), integer comparison (\textit{Comp. Int.}),
attribute querying (\textit{Query Attr.}), object querying (\textit{Query Obj.}), and
attribute comparison (\textit{Comp. Attr.}).
\textit{Overall} denotes accuracy over all test questions, while
\textit{Cls. Avg.} denotes the unweighted average across question categories.
}
\label{tab:clevr3d_real}

\setlength{\tabcolsep}{3.2pt}
\renewcommand{\arraystretch}{1.05}

\resizebox{\textwidth}{!}{
\begin{tabular}{lcccccccc}
\toprule
\textbf{Model}
& \textbf{Exist.}
& \textbf{Count.}
& \textbf{Comp. Int.}
& \textbf{Query Attr.}
& \textbf{Query Obj.}
& \textbf{Comp. Attr.}
& \textbf{Overall}
& \textbf{Cls. Avg.} \\
\midrule

\rowcolor{specializedcolor}
\multicolumn{9}{c}{\textit{3D-Specialized Models}} \\

ReferIt3D
& 85.4 & 26.6 & 57.5 & 38.3 & 11.3 & 58.5 & 41.8 & 46.2 \\

BEV+MCAN
& 85.7 & 27.3 & 60.0 & 39.5 & 14.4 & 53.7 & 42.5 & 46.8 \\

ScanQA
& 84.8 & 26.3 & 55.6 & 41.6 & 13.2 & 54.7 & 42.6 & 46.0 \\

TransVQA3D
& \textbf{88.2}
& \textbf{30.7}
& 63.1
& 37.3
& 21.7
& 57.1
& 43.7
& 49.6 \\

PointQA
& 85.9
& 29.0
& 60.3
& 39.9
& \textbf{23.2}
& 57.5
& 43.9
& 49.3 \\

E2E-MPCQA
& 86.8
& 30.6
& \textbf{64.3}
& \textbf{46.1}
& 9.2
& 59.1
& \textbf{46.7}
& 49.3 \\

\midrule

\rowcolor{ourscolor}
\multicolumn{9}{c}{\textit{\textbf{\method}}} \\

\method-2B
& 86.5
& 18.9
& 60.1
& 38.7
& 18.6
& 57.3
& 43.2
& 46.7 \\

\method-9B
& 86.8
& 30.4
& \textbf{64.3}
& 41.3
& 18.9
& \textbf{59.3}
& 45.3
& \textbf{50.1} \\

\bottomrule
\end{tabular}
}

\vspace{-2pt}
\end{table*}

\begin{table*}[t]
\centering
\caption{
\textbf{Effect of expert placement.}
We vary the Transformer layers equipped with modality-specific experts while keeping all other training settings unchanged.
\textit{Early-half} and \textit{Late-half} denote placing experts in the first and second halves of the full-attention layers, respectively.
\textit{Full-attn.} denotes our default setting, where experts are attached only to full-attention layers.
}
\label{tab:expert_placement}

\vspace{-2pt}

\begin{subtable}[t]{0.47\textwidth}
\centering
\caption{Depth understanding.}
\label{tab:expert_placement_depth}

\setlength{\tabcolsep}{4pt}
\renewcommand{\arraystretch}{1.08}

\resizebox{\linewidth}{!}{
\begin{tabular}{lcc}
\toprule
\textbf{Placement}
& \makecell{\textbf{SpatialBench}\\\textbf{Agg.}}
& \makecell{\textbf{RoboSpatial-Home}\\\textbf{Config.}} \\
\midrule

All layers
& 67.1 & 78.0 \\

Early-half
& \textbf{69.0} & 79.6 \\

Late-half
& 67.6 & 77.2 \\

Full-attn.
& 68.0  & \textbf{80.5} \\

\bottomrule
\end{tabular}
}
\end{subtable}
\hfill
\begin{subtable}[t]{0.41\textwidth}
\centering
\caption{Radar understanding.}
\label{tab:expert_placement_radar}

\setlength{\tabcolsep}{4pt}
\renewcommand{\arraystretch}{1.08}

\resizebox{\linewidth}{!}{
\begin{tabular}{lcc}
\toprule
\textbf{Placement}
& \makecell{\textbf{Action QA}\\\textbf{Acc.}}
& \makecell{\textbf{Action Caption}\\\textbf{METEOR}} \\
\midrule

All layers
& 63.8 & 45.4 \\

Early-half
& 61.2 & 45.3 \\

Late-half
& 58.4 & 45.1 \\

Full-attn.
& \textbf{65.2} & \textbf{46.1} \\

\bottomrule
\end{tabular}
}
\end{subtable}

\vspace{-2pt}
\end{table*}

\clearpage

\section{Visualization of Training Samples}
\label{app:training_visualization}
This section presents representative training examples across sensory modalities:
depth (Figs.~\ref{fig:train_depth_rel} and~\ref{fig:train_depth_close}),
point cloud (Figs.~\ref{fig:train_pc_loc} and~\ref{fig:train_pc_layout}),
IMU (Figs.~\ref{fig:train_imu_cls} and~\ref{fig:train_imu_walk}),
thermal (Figs.~\ref{fig:train_thermal_hot} and~\ref{fig:train_thermal_src}),
tactile (Figs.~\ref{fig:train_tactile_surf} and~\ref{fig:train_tactile_cmp}),
radar (Figs.~\ref{fig:train_radar_act} and~\ref{fig:train_radar_twist}),
motion (Figs.~\ref{fig:train_motion_cmp} and~\ref{fig:train_motion_seq}),
and smell (Figs.~\ref{fig:train_smell_id} and~\ref{fig:train_smell_mc}).
Each example shows the sensory input, its instruction--response pair, and the modality-aware caption used for sensory representation alignment.

\newsavebox{\casespagebox}
\newlength{\casesfitht}
\newlength{\casesboxht}
\gdef\trainingcasecaption{}
\gdef\trainingcaselabel{}

\newenvironment{casespage}{%
  \par
  \setlength{\casesfitht}{\dimexpr\pagegoal-\pagetotal-12pt\relax}%
  \ifdim\casesfitht>0.98\textheight
    \setlength{\casesfitht}{0.98\textheight}%
  \fi
  \begin{lrbox}{\casespagebox}%
  \begin{minipage}[t]{\linewidth}%
  \setlength{\parindent}{0pt}%
  \captionsetup{skip=3pt,font=small,labelfont=bf}%
  \tcbset{unbreakable, before skip=2pt, after skip=2pt}%
}{%
  \end{minipage}%
  \end{lrbox}%
  \setlength{\casesboxht}{\dimexpr\ht\casespagebox+\dp\casespagebox\relax}%
  \ifdim\casesboxht>\casesfitht
    \noindent\makebox[\linewidth][c]{%
      \resizebox*{!}{\casesfitht}{\usebox{\casespagebox}}}%
  \else
    \noindent\usebox{\casespagebox}%
  \fi
  \par\vfill\clearpage
}

\newenvironment{trainingcase}[2]{%
  \gdef\trainingcasecaption{#1}%
  \gdef\trainingcaselabel{#2}%
  \begin{chatbox}%
}{%
  \end{chatbox}%
  \captionof{figure}{\trainingcasecaption}%
  \expandafter\label\expandafter{\trainingcaselabel}%
  \par\vspace{5pt}%
}

\newcommand{\depthinput}[2]{%
\begin{center}
    \begin{minipage}[t]{0.48\linewidth}
        \centering
        \includegraphics[width=\linewidth]{#1}\\[-2pt]
        {\footnotesize\textbf{RGB}}
    \end{minipage}
    \hfill
    \begin{minipage}[t]{0.48\linewidth}
        \centering
        \includegraphics[width=\linewidth]{#2}\\[-2pt]
        {\footnotesize\textbf{Depth}}
    \end{minipage}
\end{center}
}

\newcommand{\pointcloudinput}[2]{%
\begin{center}
    \begin{minipage}[t]{0.48\linewidth}
        \centering
        \includegraphics[width=\linewidth]{#1}\\[-2pt]
        {\footnotesize\textbf{Point-Cloud View 1}}
    \end{minipage}
    \hfill
    \begin{minipage}[t]{0.48\linewidth}
        \centering
        \includegraphics[width=\linewidth]{#2}\\[-2pt]
        {\footnotesize\textbf{Point-Cloud View 2}}
    \end{minipage}
\end{center}
}

\newcommand{\imuinput}[1]{%
\begin{center}
    \includegraphics[width=0.96\linewidth]{#1}\\[-2pt]
    {\footnotesize\textbf{Visualization of the IMU sequence}}
\end{center}
}

\newcommand{\thermalinput}[1]{%
\begin{center}
    \includegraphics[
        width=0.72\linewidth,
        height=0.28\textheight,
        keepaspectratio
    ]{#1}\\[-2pt]
    {\footnotesize\textbf{Thermal Image}}
\end{center}
}

\newcommand{\radarinput}[1]{%
\begin{center}
    \includegraphics[
        width=0.96\linewidth,
        height=0.32\textheight,
        keepaspectratio
    ]{#1}\\[-2pt]
    {\footnotesize\textbf{Visualization of the radar sequence}}
\end{center}
}

\newcommand{\tactiletriplet}[4]{%
\begin{center}
    \begin{minipage}[t]{0.315\linewidth}
        \centering
        \includegraphics[
            width=\linewidth,
            height=0.16\textheight,
            keepaspectratio
        ]{#2}
    \end{minipage}
    \hfill
    \begin{minipage}[t]{0.315\linewidth}
        \centering
        \includegraphics[
            width=\linewidth,
            height=0.16\textheight,
            keepaspectratio
        ]{#3}
    \end{minipage}
    \hfill
    \begin{minipage}[t]{0.315\linewidth}
        \centering
        \includegraphics[
            width=\linewidth,
            height=0.16\textheight,
            keepaspectratio
        ]{#4}
    \end{minipage}

    \vspace{2pt}
    {\footnotesize\textbf{#1}}
\end{center}
}

\newcommand{\smellinput}[1]{%
\begin{center}
    \includegraphics[
        width=0.96\linewidth,
        height=0.28\textheight,
        keepaspectratio
    ]{#1}\\[-2pt]
    {\footnotesize\textbf{Visualization of the smell sensor sequence}}
\end{center}
}

\newcommand{\motioninput}[1]{%
\begin{center}
    \includegraphics[
        width=0.96\linewidth,
        height=0.30\textheight,
        keepaspectratio
    ]{#1}\\[-2pt]
    {\footnotesize\textbf{Visualization of the skeleton-motion sequence}}
\end{center}
}

\begin{casespage}
\noindent\textbf{Depth.}
The following examples show paired RGB--depth inputs, instruction--response supervision, and depth-aware captions used for sensory representation alignment.

\begin{trainingcase}{Depth relative-depth QA.}{fig:train_depth_rel}
\depthinput
    {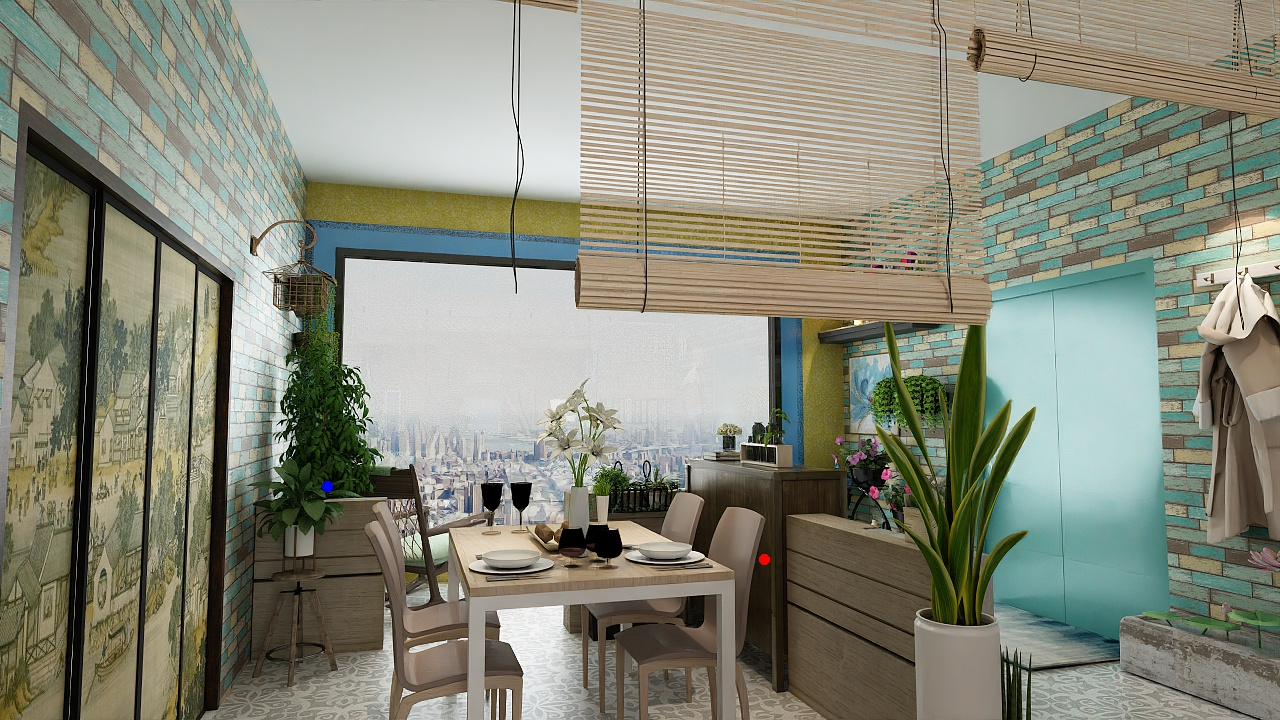}
    {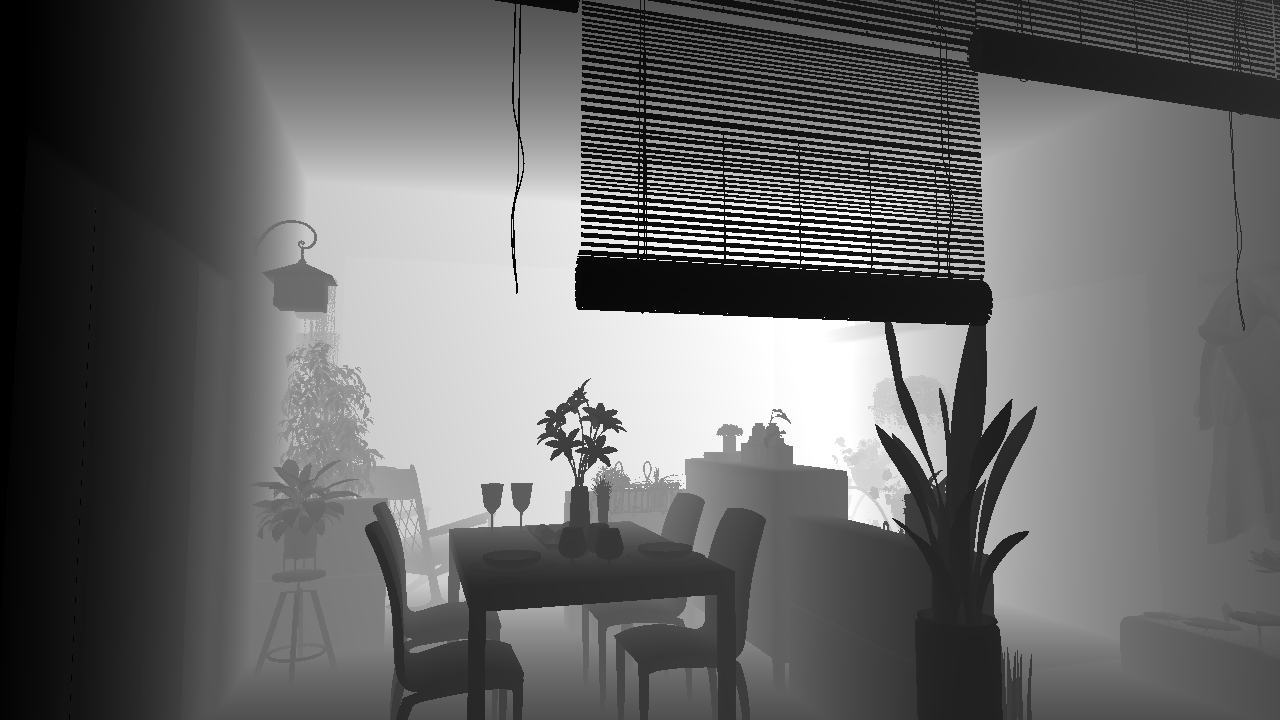}

\par\vspace{5pt}

\textcolor{green!45!black}{\textbf{Question:}}
If the red point lies at a depth of 4.1 meters, how deep is the
center of the blue point? Calculate or judge based on the 3D center
points of these objects. The unit is meter. Input a single number to
complete your answer.

\par\vspace{4pt}

\textcolor{green!45!black}{\textbf{Answer:}}
4.6

\par\vspace{6pt}

\textcolor{blue!60!black}{\textbf{Caption:}}
The dining table and the surrounding chairs are positioned in the
foreground, forming the closest major surfaces in the scene. The
cabinet and the plant in the lower right are located at a similar
depth, slightly farther back than the table but still in the immediate
foreground. The large window in the background is clearly the farthest
element, creating a significant depth separation from the furniture.
The coat hanging on the right side appears at a depth comparable to the
cabinet and plant, suggesting it is attached to the same wall plane.
The brick-patterned wall on the left and the sliding door are at a depth
similar to the wall behind the window, indicating they are part of the
same distant background plane. The hanging light is suspended above the
table, with its depth falling between the table and the ceiling, and its
position is confirmed by the depth discontinuity at the top of the
table.
\end{trainingcase}

\begin{trainingcase}{Depth closeness comparison.}{fig:train_depth_close}
\depthinput
    {}
    {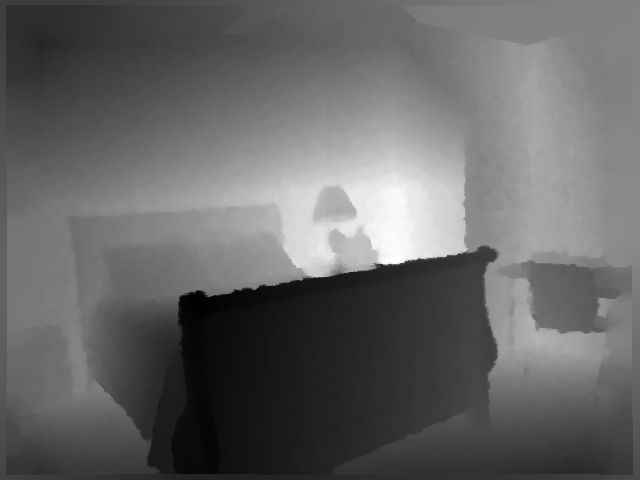}

\par\vspace{5pt}

\textcolor{green!45!black}{\textbf{Question:}}
Is the bed closer to the camera than the desk?

\par\vspace{4pt}

\textcolor{green!45!black}{\textbf{Answer:}}
Yes

\par\vspace{6pt}

\textcolor{blue!60!black}{\textbf{Caption:}}
The bed and the stack of books in the lower left are at similar depths,
forming a foreground cluster. The headboard is positioned directly
behind the bed, creating a continuous depth surface with it. The lamp
on the bedside table is slightly farther back than the bed's headboard,
indicating it is on a separate surface. The desk and chair in the lower
right are located at a depth comparable to the lamp, suggesting they are
on the same side wall plane. The window and the picture above it are on
the far wall, appearing at a greater depth than the bed, desk, and chair.
The ceiling fan is the farthest object from the viewer, positioned high
above the other items. The depth transitions from the foreground bed
area to the background wall are gradual, with clear depth discontinuities
at the edges of the furniture.
\end{trainingcase}
\end{casespage}

\begin{casespage}
\noindent\textbf{Point Cloud.}
Each point cloud is rendered from two viewpoints for visualization only; the model receives the underlying 3D point cloud as input. Each example also shows its instruction--response pair and scene-level caption used for sensory representation alignment.

\begin{trainingcase}{Point-cloud object localization.}{fig:train_pc_loc}
\pointcloudinput
    {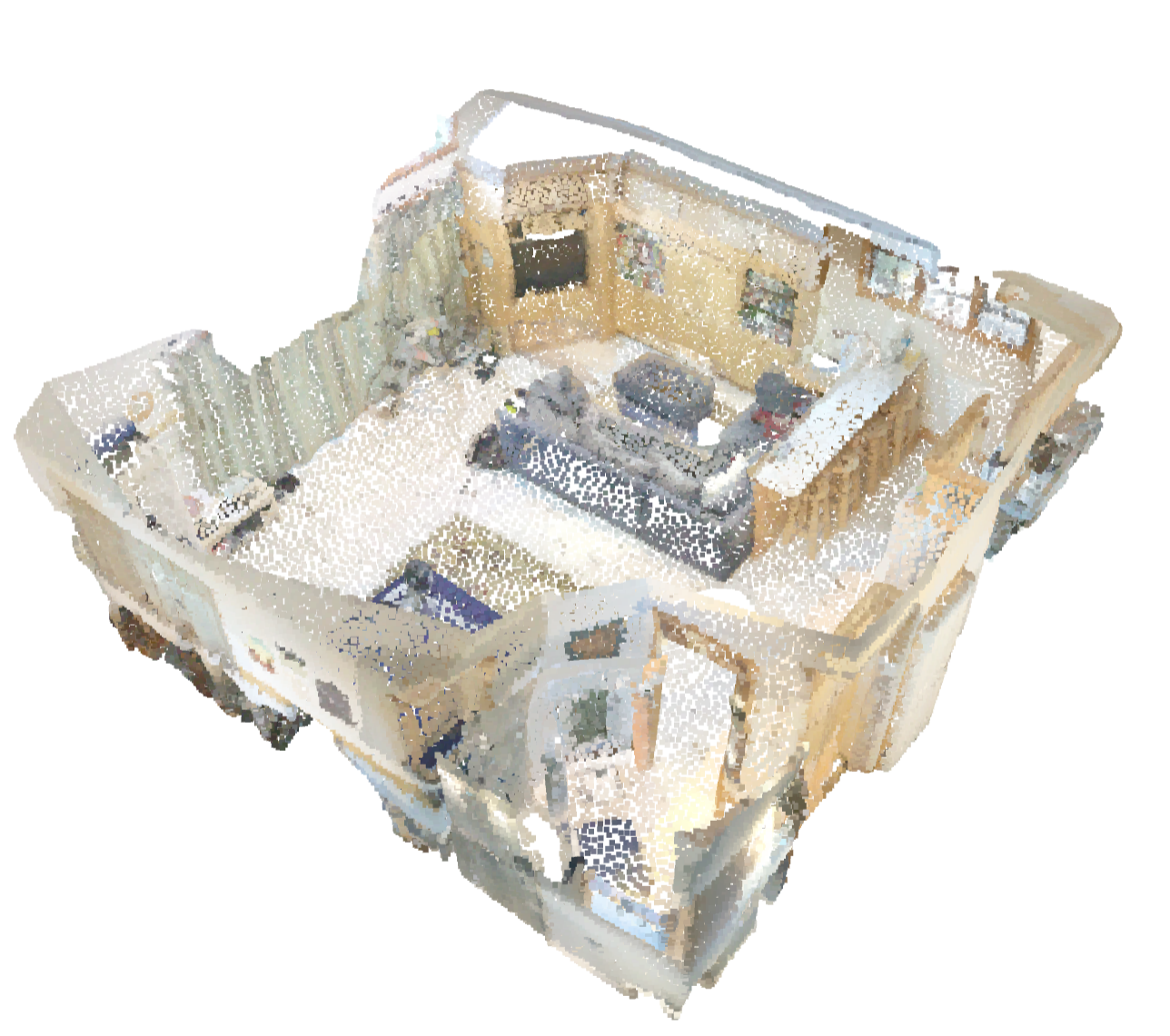}
    {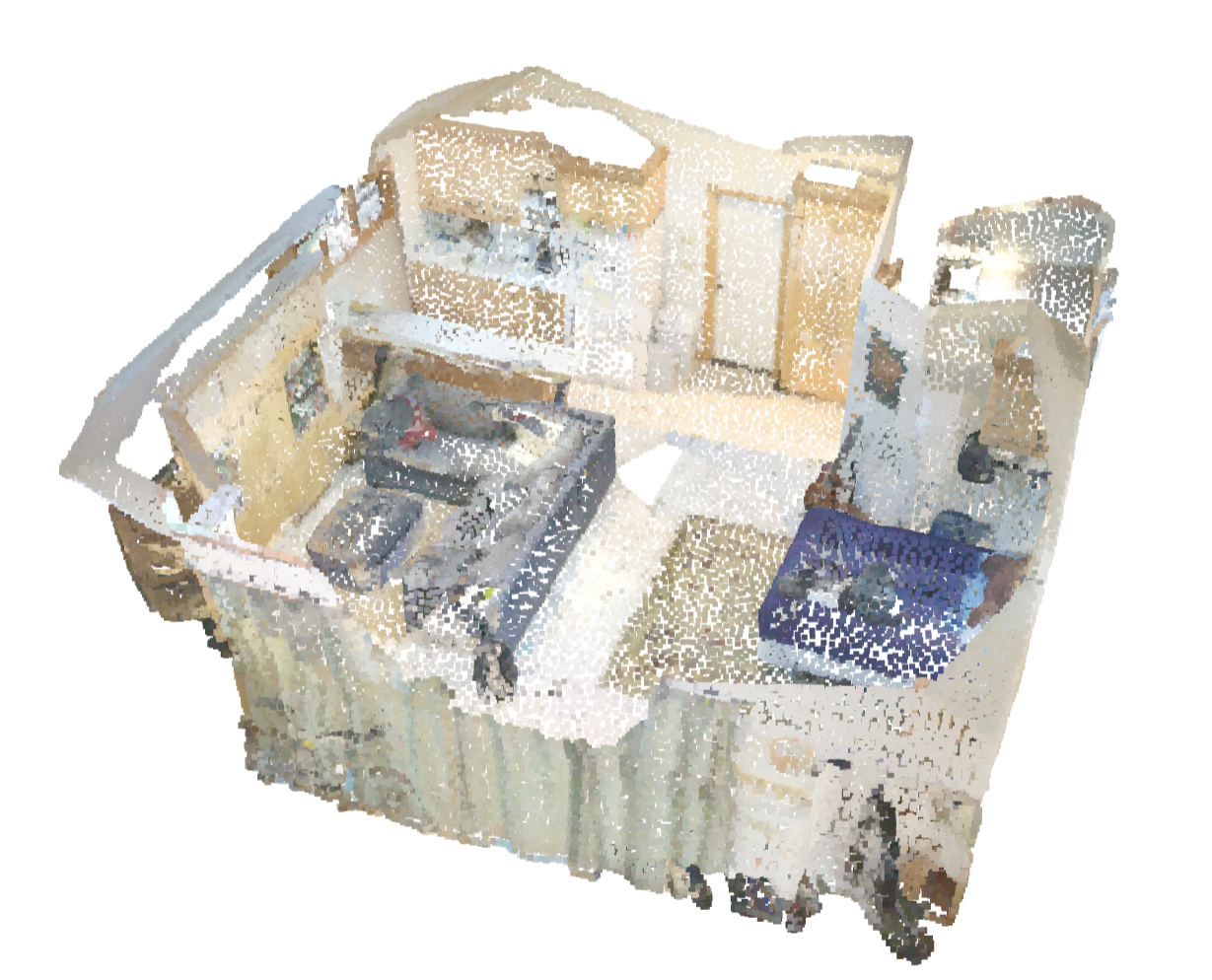}

\par\vspace{5pt}

\textcolor{green!45!black}{\textbf{Question:}}
Where is the large wooden bookshelf placed?

\par\vspace{4pt}

\textcolor{green!45!black}{\textbf{Answer:}}
To the right of the brown wooden cabinet and television, against the
wall near the windows.

\par\vspace{6pt}

\textcolor{blue!60!black}{\textbf{Caption:}}
In this apartment, there are 5 cabinets, 1 bed, 3 trash cans,
1 microwave, and 1 TV. The cabinets are positioned in front of the
trash cans, while the bed is in front of the cabinet. The trash cans
are also behind the cabinet and to the left of the bed. The TV is
inside one of the cabinets. The bed is positioned behind the cabinet
and to the right of the trash cans. This apartment seems to be
well-equipped with storage options and has a comfortable sleeping area.
\end{trainingcase}

\begin{trainingcase}{Point-cloud indoor layout QA.}{fig:train_pc_layout}
\pointcloudinput
    {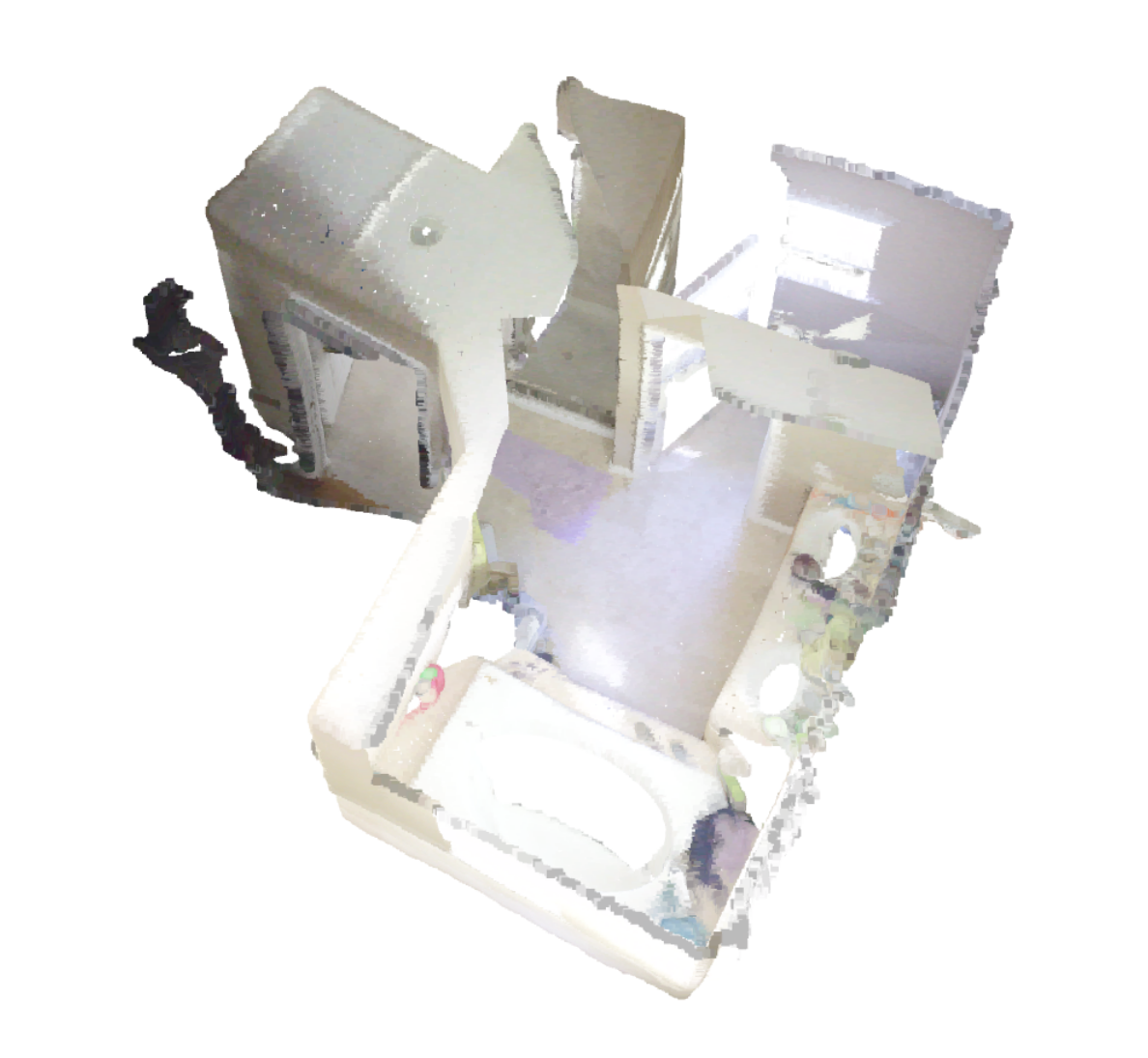}
    {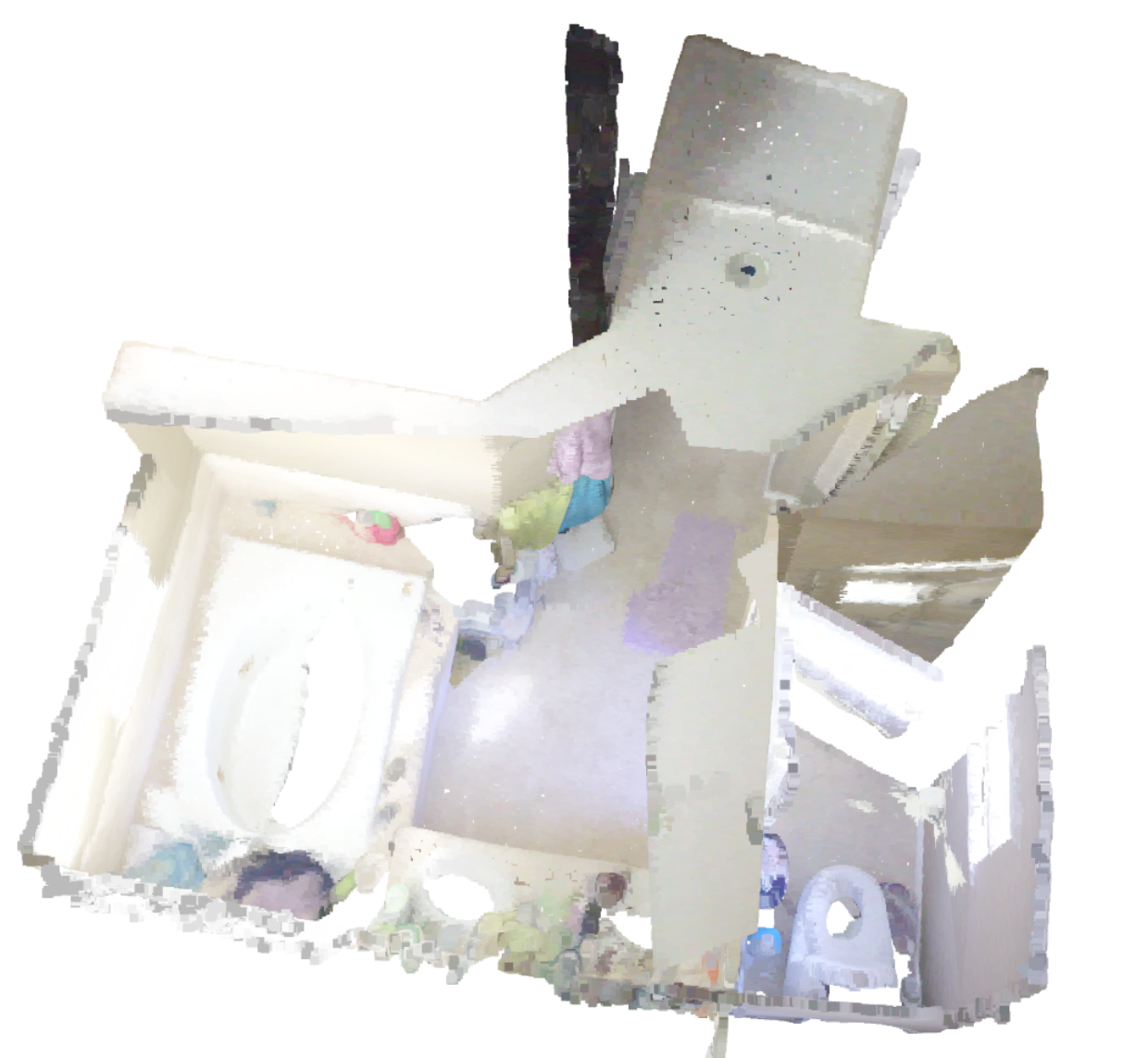}

\par\vspace{5pt}

\textcolor{green!45!black}{\textbf{Question:}}
Based on the 3D scene and the provided view, answer the following
question briefly. Where is the toilet room located when you enter the room?

\par\vspace{4pt}

\textcolor{green!45!black}{\textbf{Answer:}}
Next to the bathroom vanity.

\par\vspace{6pt}

\textcolor{blue!60!black}{\textbf{Caption:}}
In this bathroom scene, there are three windows, one mirror, one cup,
and one toilet. The room exudes a sense of cleanliness and
functionality. The windows allow natural light to filter in, creating
a bright and airy atmosphere. The mirror serves as a reflection of
one's appearance, while the cup is likely used for drinking or
rinsing. The toilet, a necessary fixture, provides a place for personal
hygiene. The objects are positioned in a way that suggests convenience
and accessibility.
\end{trainingcase}
\end{casespage}

\begin{casespage}
\noindent\textbf{IMU.}
Each IMU sequence is visualized as accelerometer and gyroscope curves for presentation only; the model receives the underlying six-channel time series. Each example also shows its instruction--response pair and sensor caption used for sensory representation alignment.

\begin{trainingcase}{IMU activity classification.}{fig:train_imu_cls}
\imuinput{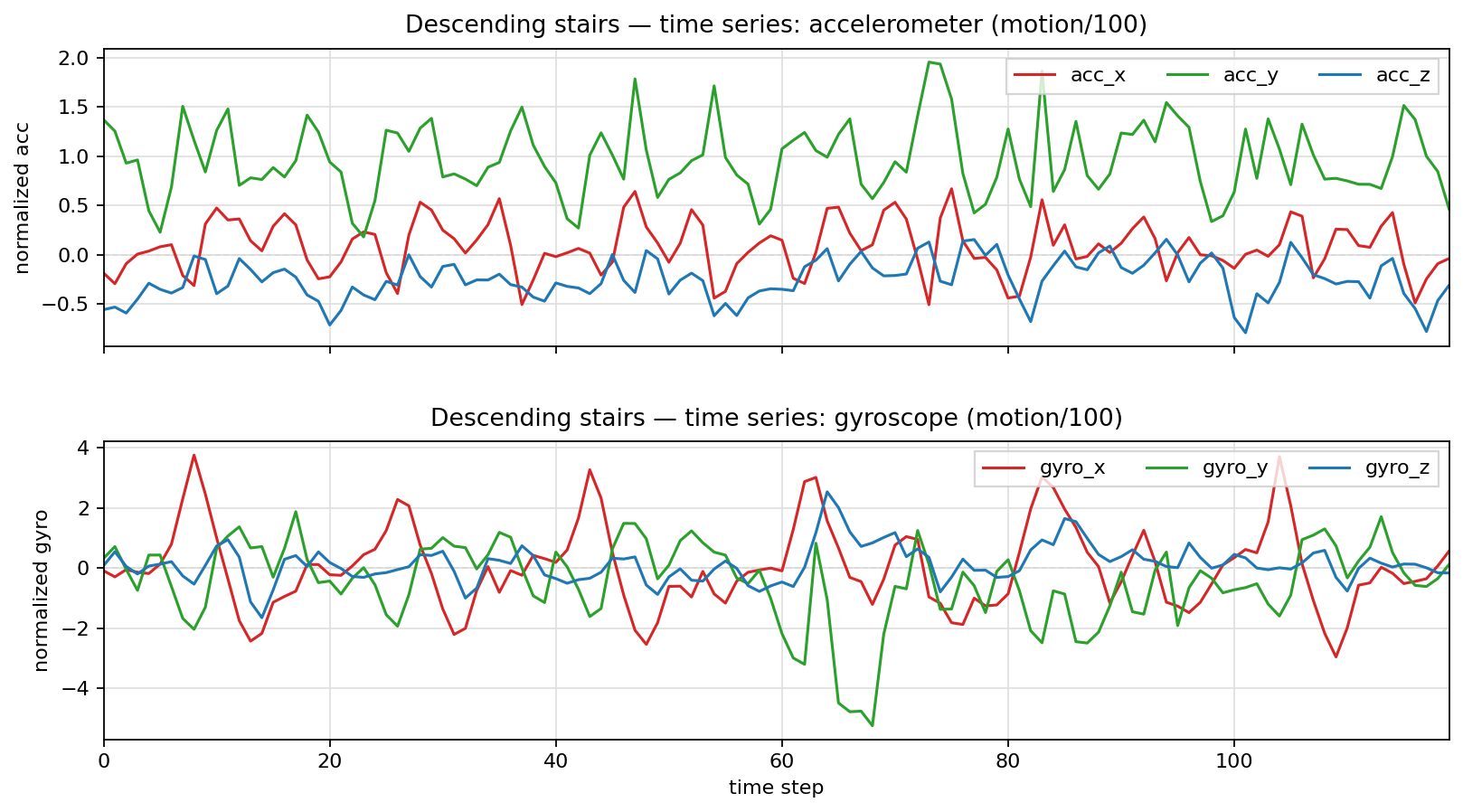}

\par\vspace{5pt}

\textcolor{green!45!black}{\textbf{Question:}}
The given IMU sensor data can be associated with one of the following
classes: climbing stairs, descending stairs, jogging, sitting, standing,
or walking. Write only the name of the identified class.

\par\vspace{4pt}

\textcolor{green!45!black}{\textbf{Answer:}}
Descending stairs

\par\vspace{6pt}

\textcolor{blue!60!black}{\textbf{Caption:}}

\textbf{Features:}
Staircase descent, varied acceleration and rotation.

\smallskip
\textbf{Narration:}
The gyroscope data shows inconsistent rotational movements as values
fluctuate. Accelerometer readings display a mix of vertical peaks and
troughs indicative of stair descending. Initially, as the user begins
to descend, the accelerometer shows a spike in vertical motion as the
user takes a step downward. The gyroscope values indicate slight
adjustments in rotation to maintain balance on the stairs. There are
moments where both the gyroscope and accelerometer data show quick
changes, possibly when adjusting for stair height or landing. Towards
the end of the activity, both sensors show a decrease in motion as the
user reaches the bottom of the stairs.
\end{trainingcase}

\begin{trainingcase}{IMU walking-motion analysis.}{fig:train_imu_walk}
\imuinput{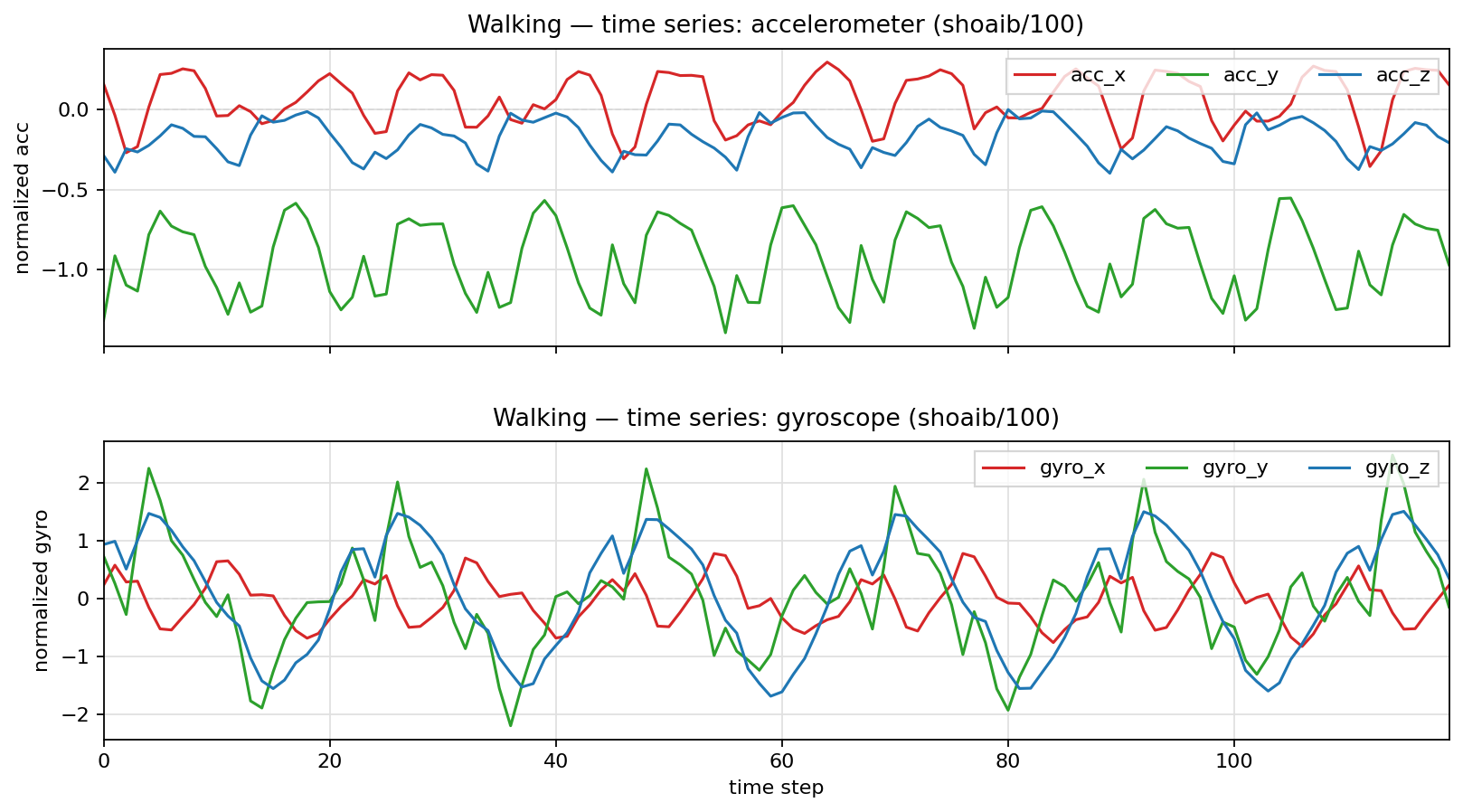}

\par\vspace{5pt}

\textcolor{green!45!black}{\textbf{Question:}}
How can we differentiate between gyroscope readings indicating arm
swinging during walking and those indicating rotational motion of the
torso?

\par\vspace{4pt}

\textcolor{green!45!black}{\textbf{Answer:}}
Gyroscope readings for arm swinging during walking typically fluctuate
along the axis corresponding to the swinging motion, whereas torso
rotation produces consistent patterns across multiple axes.

\par\vspace{6pt}

\textcolor{blue!60!black}{\textbf{Caption:}}

\textbf{Features:}
Walking pattern analysis, vibrational characteristics.

\smallskip
\textbf{Narration:}
The initial gyroscope readings show slight movements along all axes.
Accelerometer data indicates variations in acceleration as the user takes steps.
Gyroscope data displays consistent rotation patterns, suggesting periodic movement.
Accelerometer measurements change in accordance with the steps taken.
The gyroscope values indicate rotational motion, possibly caused by arm swinging.
Accelerometer data exhibits fluctuations aligned with individual footsteps.
The gyroscope readings show smooth oscillations, indicating a rhythmic gait.
Accelerometer data exhibits continuous variations typical of walking.
The gyroscope reflects stable angular velocities characteristic of a steady pace.
Accelerometer values change periodically with the user's walking rhythm.
\end{trainingcase}
\end{casespage}

\begin{casespage}
\noindent\textbf{Thermal.}
Each example shows a thermal image, its instruction--response pair, and a temperature-aware caption used for sensory representation alignment.

\begin{trainingcase}{Thermal hotspot identification.}{fig:train_thermal_hot}
\thermalinput{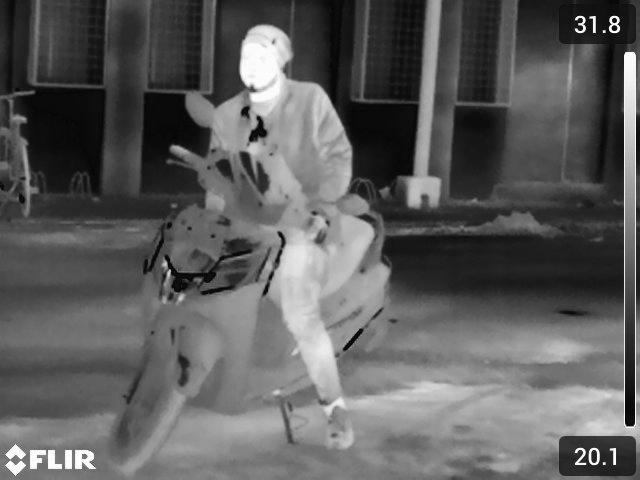}

\par\vspace{5pt}

\textcolor{green!45!black}{\textbf{Question:}}
What part of the image shows the highest temperature?

\par\vspace{3pt}

\begin{tabular}{@{}p{0.48\linewidth}p{0.48\linewidth}@{}}
(A) The ground.
&
(B) The background wall. \\[2pt]
(C) The scooter's handles.
&
(D) The person's head.
\end{tabular}

\par\vspace{4pt}

\textcolor{green!45!black}{\textbf{Answer:}}
D

\par\vspace{6pt}

\textcolor{blue!60!black}{\textbf{Caption:}}
From left to right, the image shows the front of a scooter, a rider in
the center, and the rear part of the scooter. A building and road
appear in the background. Whiter regions indicate higher temperatures;
the rider's face, neck, and exposed leg are warmer than the clothing
and scooter, while the road and building appear cooler.
\end{trainingcase}

\begin{trainingcase}{Thermal heat-source identification.}{fig:train_thermal_src}
\thermalinput{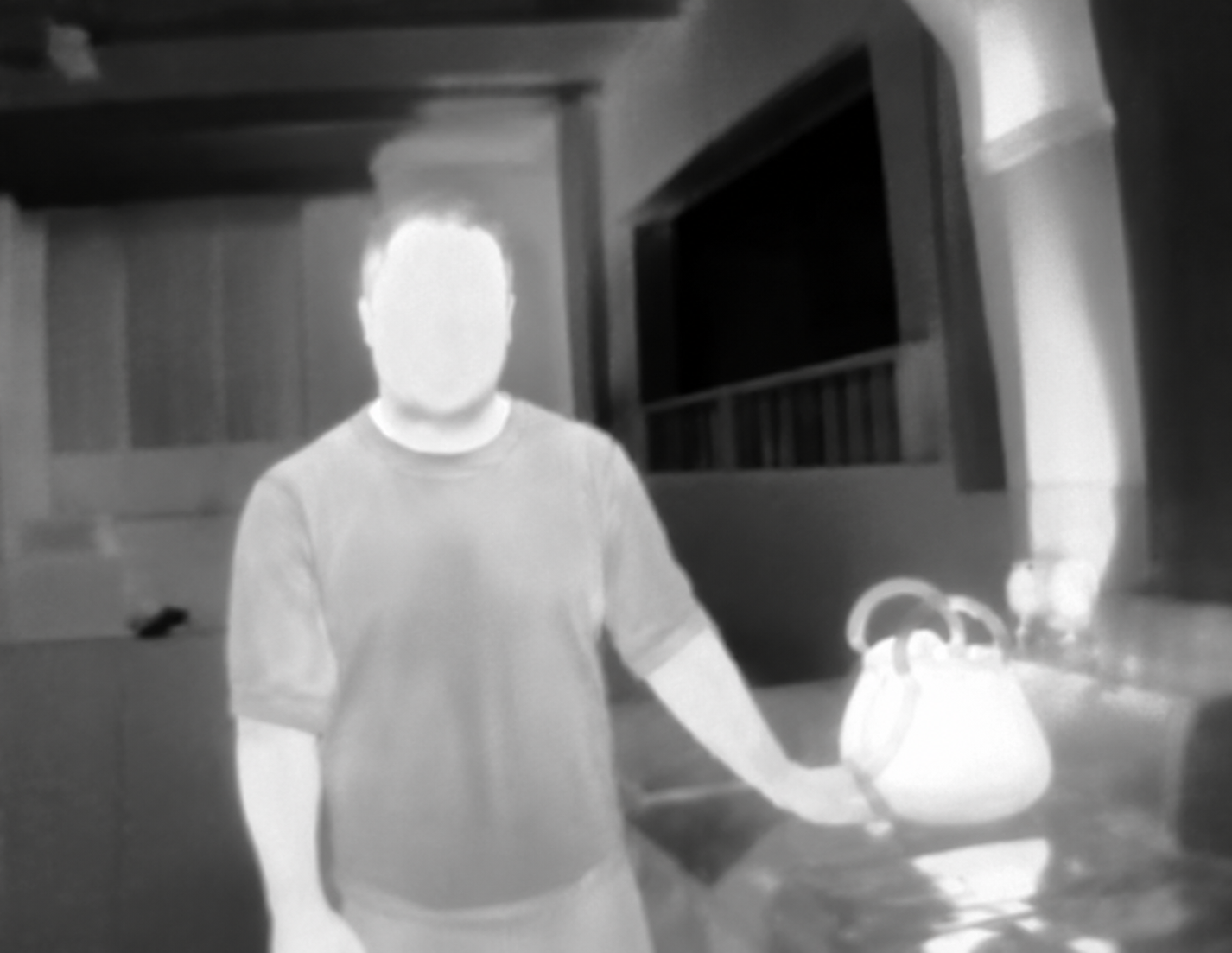}

\par\vspace{5pt}

\textcolor{green!45!black}{\textbf{Question:}}
What object is emitting the most heat in the image?

\par\vspace{3pt}

\begin{tabular}{@{}p{0.48\linewidth}p{0.48\linewidth}@{}}
(A) The curtains in the background.
&
(B) The kettle on the table. \\[2pt]
(C) The ceiling above.
&
(D) The person standing.
\end{tabular}

\par\vspace{4pt}

\textcolor{green!45!black}{\textbf{Answer:}}
B

\par\vspace{6pt}

\textcolor{blue!60!black}{\textbf{Caption:}}
From left to right, the image shows a doorway, a person standing in the
center, and a kettle on a countertop. The walls, window, and columns
are visible in the background. Whiter regions indicate higher
temperatures; the person's face and exposed arms are warmer than the
clothing, while the kettle and the area near its opening also appear hot.
\end{trainingcase}
\end{casespage}

\begin{casespage}
\noindent\textbf{Tactile.}
Each tactile observation is represented by three sampled frames.
Single-object examples contain one frame triplet, whereas multi-object
examples contain a separate triplet for each object. The corresponding
instruction--response pair and tactile property caption are shown below.

\begin{trainingcase}{Tactile surface description.}{fig:train_tactile_surf}
\tactiletriplet
    {Tactile Input}
    {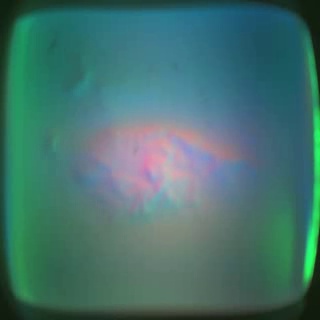}
    {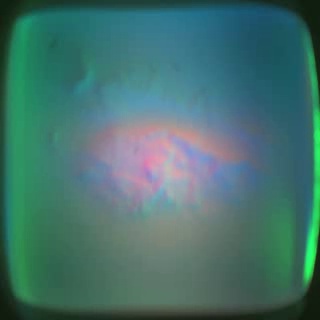}
    {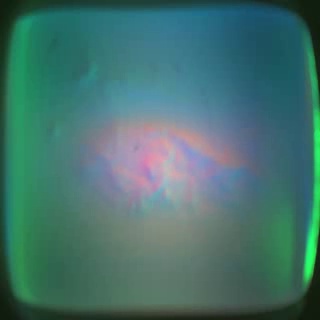}

\par\vspace{5pt}

\textcolor{green!45!black}{\textbf{Question:}}
Can you describe the surface characteristics shown in this tactile
video?

\par\vspace{4pt}

\textcolor{green!45!black}{\textbf{Answer:}}
The surface is extremely hard, with strong protrusions, no elasticity,
and strong friction.

\par\vspace{6pt}

\textcolor{blue!60!black}{\textbf{Caption:}}
The surface is extremely hard, with strong protrusions, no elasticity,
and strong friction.
\end{trainingcase}

\begin{trainingcase}{Tactile two-object comparison.}{fig:train_tactile_cmp}
\tactiletriplet
    {First Object}
    {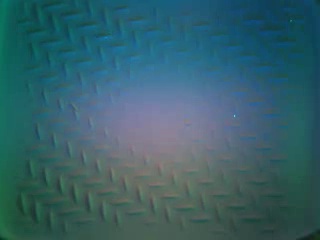}
    {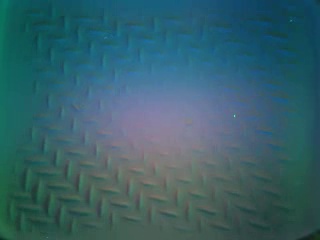}
    {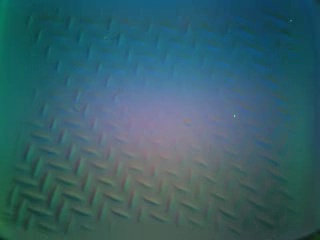}

\vspace{3pt}

\tactiletriplet
    {Second Object}
    {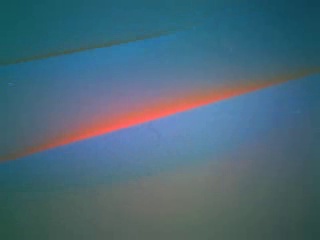}
    {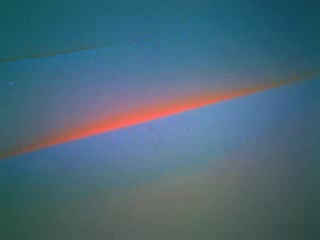}
    {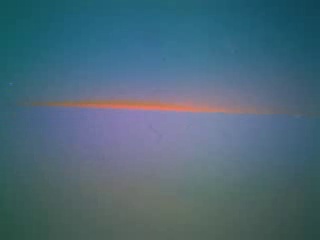}

\par\vspace{5pt}

\textcolor{green!45!black}{\textbf{Question:}}
Is the first object covered with smaller bumps than the second object?
Describe both objects before answering.

\par\vspace{4pt}

\textcolor{green!45!black}{\textbf{Answer:}}

\textbf{First object:}
The surface feels somewhat firm and notably rough, with small bumps
and raised areas. Overall, it is moderately hard and rough with small
bumps.

\smallskip
\textbf{Second object:}
The surface feels hard and rigid but notably smooth, with no prominent
bumps or ridges. Overall, it is hard and smooth with no bumps.

\smallskip
\textbf{Conclusion:}
No. The first object has small bumps, whereas the second object has no
prominent bumps.

\par\vspace{6pt}

\textcolor{blue!60!black}{\textbf{Caption:}}

\textbf{First object:}
The surface feels somewhat firm and notably rough, with small bumps
and raised areas. Overall, it is moderately hard and rough with small
bumps.

\smallskip
\textbf{Second object:}
The surface feels hard and rigid but notably smooth, with no prominent
bumps or ridges. Overall, it is hard and smooth with no bumps.
\end{trainingcase}
\end{casespage}

\begin{casespage}
\noindent\textbf{Radar.}
Each radar sequence is visualized using sampled point-cloud frames and signal heatmaps for presentation only; the model receives the underlying temporal mmWave point clouds. Each example also shows its instruction--response pair and action caption used for sensory representation alignment.

\begin{trainingcase}{Radar action recognition.}{fig:train_radar_act}
\radarinput{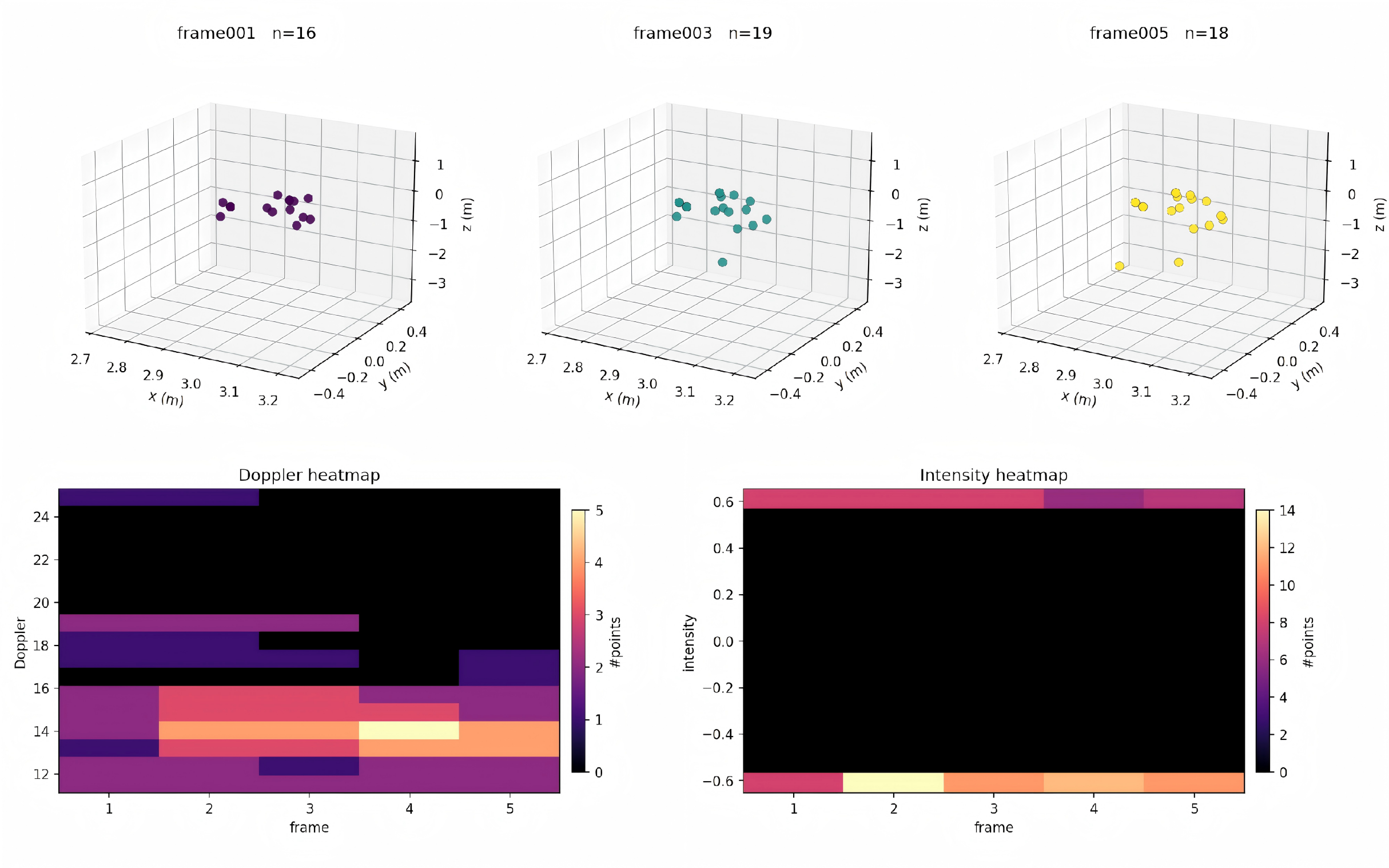}

\par\vspace{5pt}

\textcolor{green!45!black}{\textbf{Question:}}
Identify a suitable phrase from the action list that describes the
human action represented by the input data.

\par\vspace{3pt}

\textbf{Action List:}
stretching and relaxing; horizontal chest expansion; vertical chest
expansion; left twist; right twist; mark time; left limb extension;
right limb extension; lunge toward left-front; lunge toward
right-front; both limb extension; squat; raising left hand; raising
right hand; lunge toward left side; lunge toward right side; waving
left hand; waving right hand; picking up things; throwing toward left
side; throwing toward right side; kicking toward left side; kicking
toward right side; left body extension; right body extension; jumping
up; bowing.

\par\vspace{4pt}

\textcolor{green!45!black}{\textbf{Answer:}}
stretching and relaxing

\par\vspace{6pt}

\textcolor{blue!60!black}{\textbf{Caption:}}
The radar sequence shows a person in a room performing gentle
full-body stretches to warm up and relax their muscles.
\end{trainingcase}

\begin{trainingcase}{Radar twist-direction QA.}{fig:train_radar_twist}
\radarinput{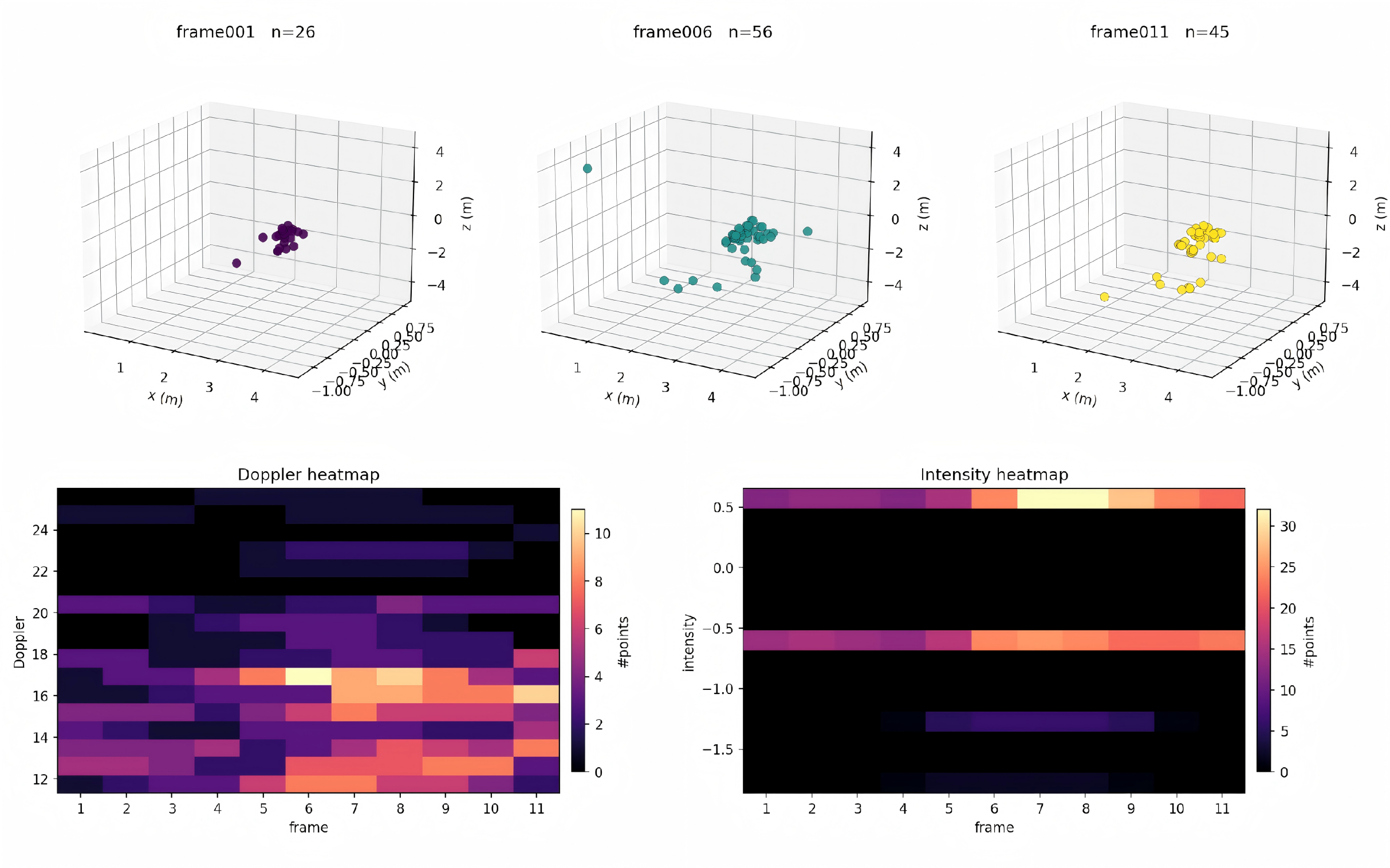}

\par\vspace{5pt}

\textcolor{green!45!black}{\textbf{Question:}}
Which twist direction is shown in this interval? Answer with only one
word: left or right.

\par\vspace{4pt}

\textcolor{green!45!black}{\textbf{Answer:}}
left

\par\vspace{6pt}

\textcolor{blue!60!black}{\textbf{Caption:}}
The person in the radar sequence performs a left twist, rotating their
torso to the left in a focused movement centered around the waist.
\end{trainingcase}
\end{casespage}

\begin{casespage}
\noindent\textbf{Motion.}
Each skeleton-motion sequence is visualized using sampled poses and joint trajectories for presentation only; the model receives the underlying sequence of 3D joint coordinates. Each example also shows its instruction--response pair and temporal caption used for sensory representation alignment.

\begin{trainingcase}{Motion temporal comparison.}{fig:train_motion_cmp}
\motioninput{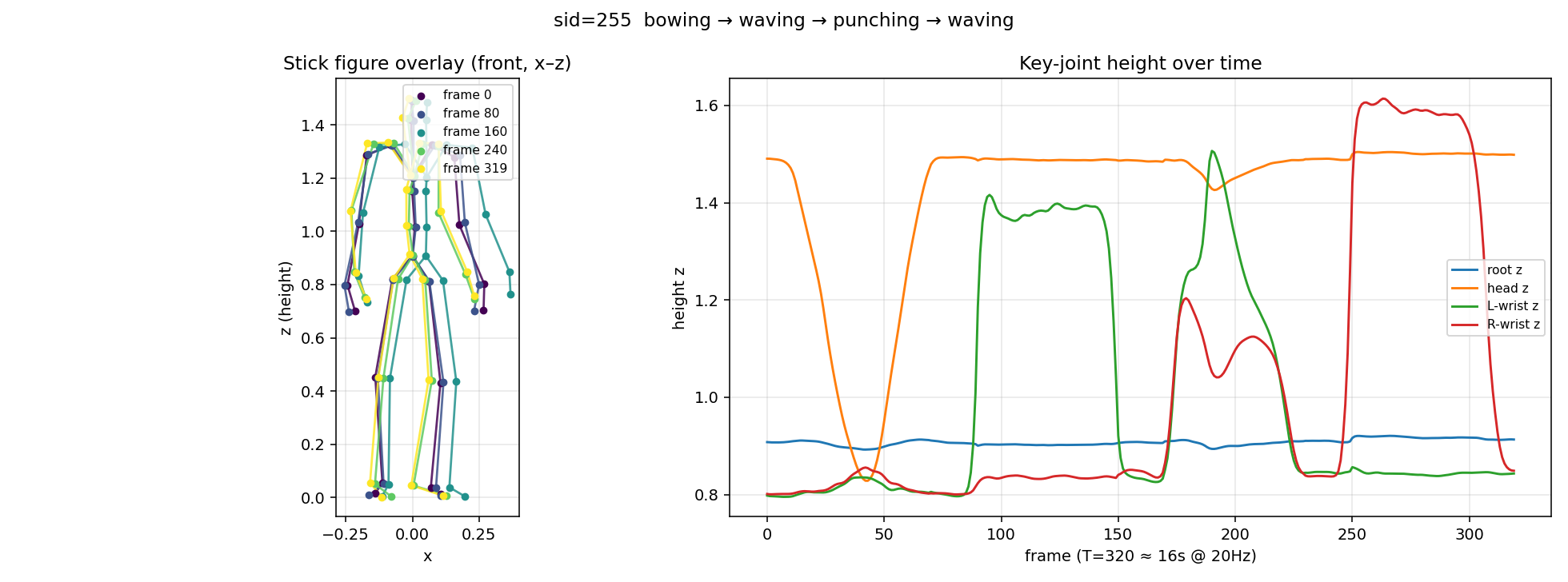}

\par\vspace{5pt}

\textcolor{green!45!black}{\textbf{Question:}}
Are the actions at 4.199 seconds and 5.884 seconds essentially the same?

\par\vspace{4pt}

\textcolor{green!45!black}{\textbf{Answer:}}
Yes

\par\vspace{6pt}

\textcolor{blue!60!black}{\textbf{Caption:}}
The first action is bowing, which lasts from 0.0 to 4.0 seconds.
The person then waves from 4.0 to 8.0 seconds, punches from 8.0 to
12.0 seconds, and waves again from 12.0 to 16.0 seconds.
\end{trainingcase}

\begin{trainingcase}{Motion action sequencing.}{fig:train_motion_seq}
\motioninput{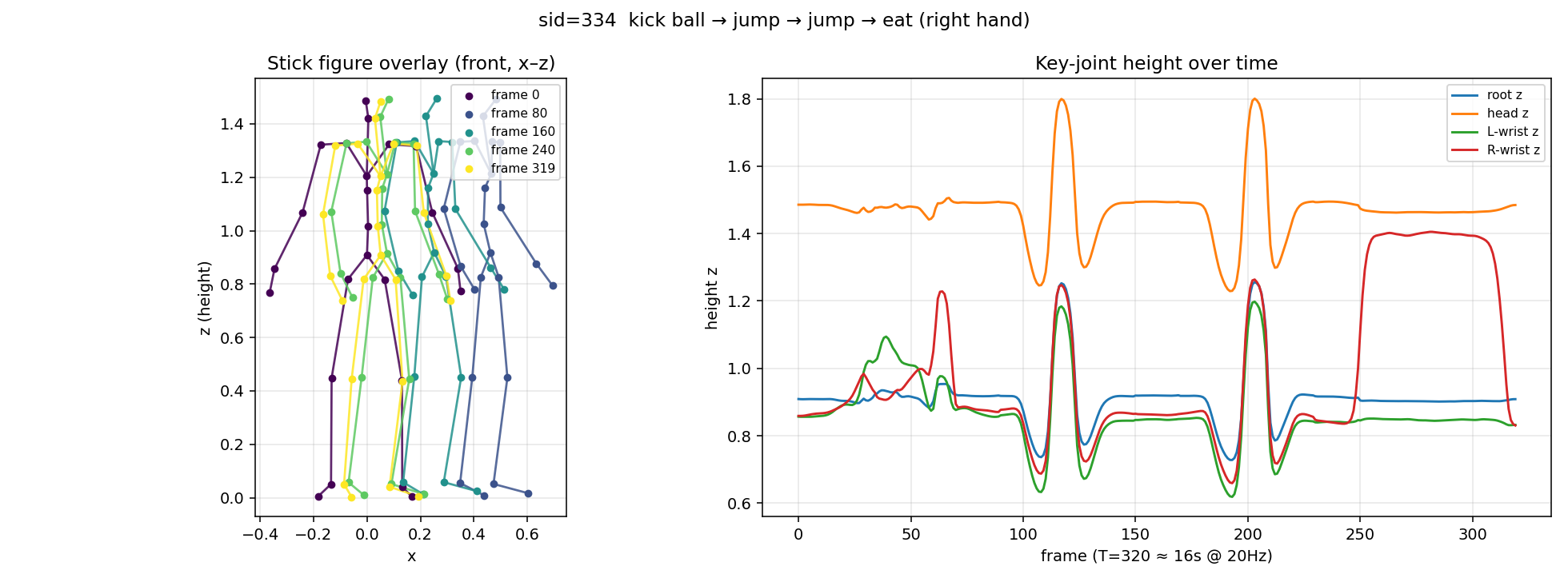}

\par\vspace{5pt}

\textcolor{green!45!black}{\textbf{Question:}}
What action took place immediately after the person kicked a ball?

\par\vspace{4pt}

\textcolor{green!45!black}{\textbf{Answer:}}
The person jumped once.

\par\vspace{6pt}

\textcolor{blue!60!black}{\textbf{Caption:}}
The person begins by kicking a ball from 0.0 to 4.0 seconds, followed
by one jump from 4.0 to 8.0 seconds and another from 8.0 to
12.0 seconds. The final action is eating with the right hand from
12.0 to 16.0 seconds.
\end{trainingcase}
\end{casespage}

\begin{casespage}
\noindent\textbf{Smell.}
Each smell sensor sequence is visualized as six-channel response curves for presentation only; the model receives the underlying numerical measurements. Each example also shows its instruction--response pair and sensor caption used for sensory representation alignment.

\begin{trainingcase}{Smell substance identification.}{fig:train_smell_id}
\smellinput{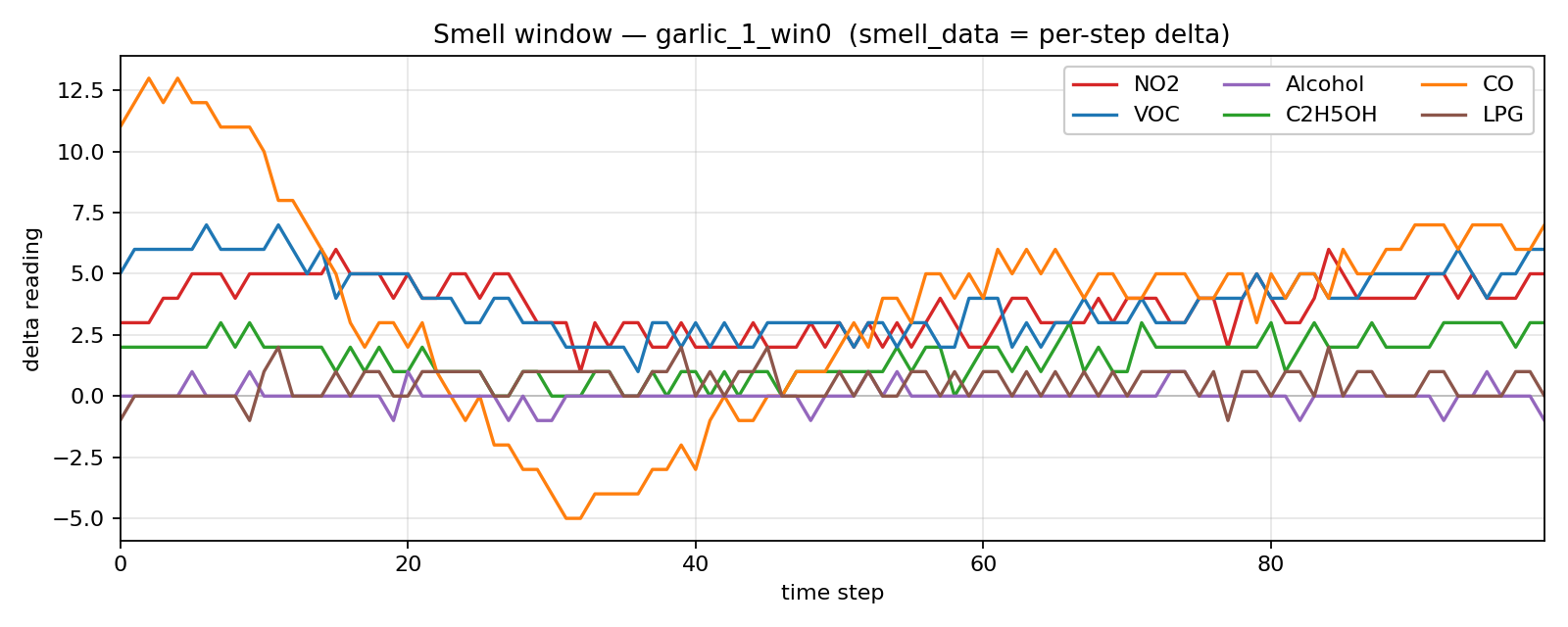}

\par\vspace{5pt}

\textcolor{green!45!black}{\textbf{Question:}}
The given smell sensor reading can be associated with one of the
following substances: allspice, almond, angelica, apple, asparagus,
avocado, banana, brazil\_nut, broccoli, brussel\_sprouts, cabbage,
cashew, cauliflower, chamomile, chervil, chestnuts, chives, cinnamon,
cloves, coriander, cumin, dill, garlic, ginger, hazelnut, kiwi, lemon,
mandarin\_orange, mango, mint, mugwort, mustard, nutmeg, oregano,
peach, peanuts, pear, pecans, pili\_nut, pineapple, pistachios, potato,
radish, saffron, star\_anise, strawberry, sweet\_potato, tomato,
turnip, and walnuts. Write only the name of the identified substance.

\par\vspace{4pt}

\textcolor{green!45!black}{\textbf{Answer:}}
garlic

\par\vspace{6pt}

\textcolor{blue!60!black}{\textbf{Caption:}}
\textbf{Features:}
CO and VOC peak in Q1 and then fall to troughs by Q3--Q4 before
rebounding. NO$_2$ declines until Q4 and then rises, while
C$_2$H$_5$OH reaches its minimum at Q3 before increasing through Q8.
LPG exhibits little variation.
\end{trainingcase}

\begin{trainingcase}{Smell multiple-choice recognition.}{fig:train_smell_mc}
\smellinput{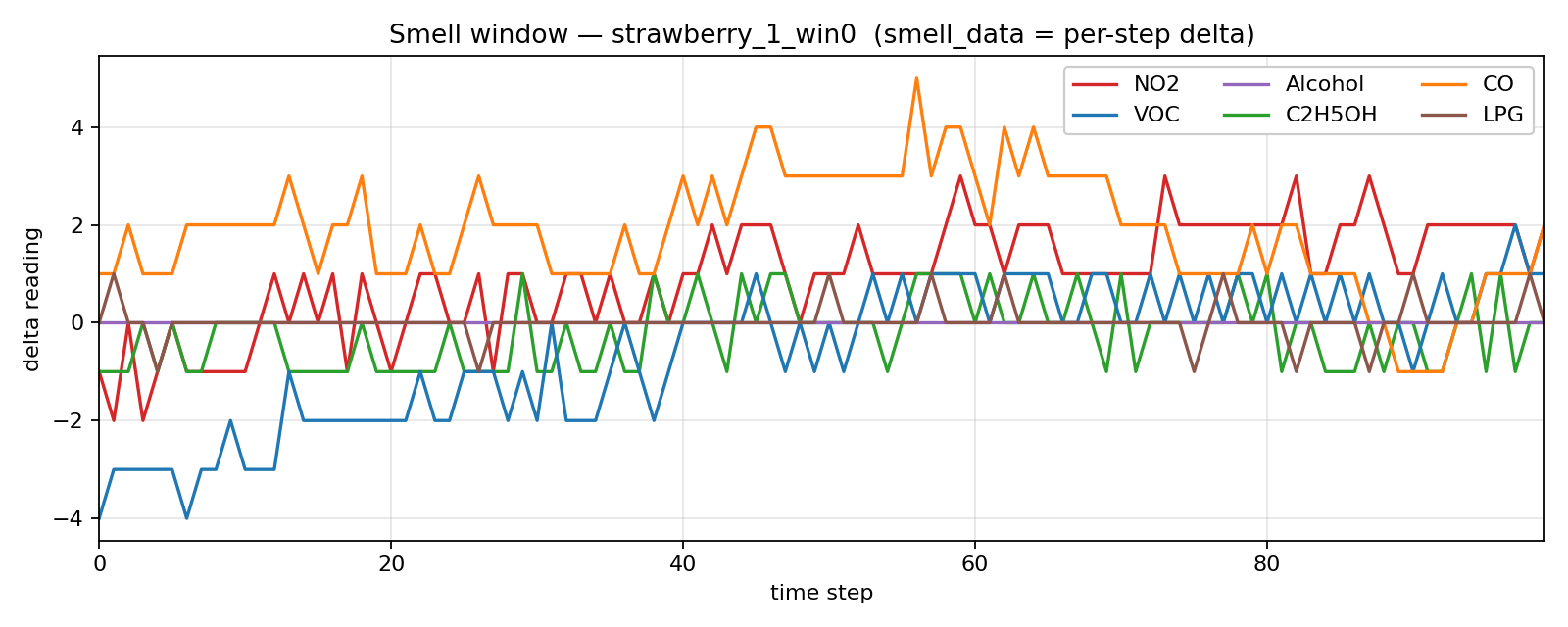}

\par\vspace{5pt}

\textcolor{green!45!black}{\textbf{Question:}}
Which substance best matches this gas sensor recording?

\par\vspace{3pt}

\begin{tabular}{@{}p{0.48\linewidth}p{0.48\linewidth}@{}}
(A) strawberry
&
(B) cinnamon \\[2pt]
(C) pecans
&
(D) mango
\end{tabular}
\par\vspace{4pt}

\textcolor{green!45!black}{\textbf{Answer:}}
strawberry

\par\vspace{6pt}

\textcolor{blue!60!black}{\textbf{Caption:}}
Features: NO2 dips at Q1 then climbs steadily to a peak in Q7, easing in Q8; CO rises to a Q5 peak then drops to trough at Q8; VOC rebounds from Q1 trough to Q6 peak then softens; C2H5OH rises to Q5 then falls; Alcohol and LPG stay close to unchanged.
\end{trainingcase}
\end{casespage}

\end{document}